\documentclass[11pt, letterpaper, shortlabels]{archer}

\usepackage{amsmath,amsfonts,bm}
\usepackage{amsthm}

\theoremstyle{plain}

\theoremstyle{definition}

\theoremstyle{remark}

\def\eqref#1{equation~\ref{#1}}

\usepackage{microtype}
\usepackage{graphicx}
\usepackage{pgf}
\usepackage{booktabs}
\usepackage{subcaption}
\usepackage{color}
\usepackage{algorithm}
\usepackage{algorithmicx}
\usepackage{algpseudocode}
\usepackage{multirow}
\usepackage{resizegather}
\usepackage{hyperref}
\usepackage{color-edits}
\usepackage{siunitx}

\usepackage{comment}
\usepackage[utf8]{inputenc}
\usepackage[T1]{fontenc}
\usepackage{url}
\usepackage{amsfonts}
\usepackage{nicefrac}
\usepackage{tabulary}
\usepackage{tabularx}
\usepackage{array}
\usepackage{bbm}
\usepackage{indentfirst}
\usepackage{cleveref}
\usepackage{empheq}
\usepackage{mdframed}
\usepackage{amsmath, amssymb, amsthm, bm}
\usepackage{mathtools}

\newlength\savewidth

\usepackage[textsize=tiny]{todonotes}
\usepackage{wrapfig}

\usepackage{enumitem}
\setlist[itemize]{leftmargin=11pt, itemsep=1pt, topsep=0pt}
\setlist[enumerate]{leftmargin=11pt, itemsep=1pt, topsep=0pt}

\Crefname{problem}{Problem}{Problems}

\usepackage[numbers]{natbib}

\hypersetup{
    colorlinks = true,
    citecolor = {magenta},
}

\usepackage{xspace}

\newcommand{\name}{\textit{Pillar-0}\xspace}

\makeatletter
\newtheorem*{rep@definition}{\rep@title}
\newcommand{\newrepdefinition}[2]{%
	\newenvironment{rep#1}[1]{%
		\def\rep@title{#2 \ref{##1}}%
		\begin{rep@definition}}%
		{\end{rep@definition}}}
\makeatother

\newrepdefinition{definition}{Definition}
\newrepdefinition{lemma}{Lemma}
\newrepdefinition{proposition}{Proposition}

\usepackage{booktabs}
\usepackage[labelfont=bf]{caption}
\usepackage{hhline}
\usepackage{array}
\usepackage{longtable}
\definecolor{rowgray}{gray}{0.97}
\definecolor{linegray}{gray}{0.9}
\renewcommand{\footnotesize}{\fontsize{7}{9}\selectfont}

\usepackage{tabularx}
\usepackage{pifont}

\usepackage{url}

\usepackage{color}
\usepackage{graphicx}
\usepackage{float}
\usepackage{multirow}

\usepackage{datetime}
\title{
    
        Pillar-0: 
        A New Frontier for Radiology Foundation Models
    
}

\author{
Kumar Krishna Agrawal$^{1,\ast}$     ,
Longchao Liu$^{2,\dagger}$,
Long Lian$^{1,\dagger}$,
Michael Nercessian$^{2,\dagger}$,
Natalia Harguindeguy$^{2,\dagger}$, \newline
Yufu Wu$^{3}$,
Peter Mikhael$^{4}$,
Gigin Lin$^{3,5,6}$,
Lecia V. Sequist$^{7}$,
Florian Fintelmann$^{8,9}$,
Trevor Darrell$^{1}$,
Yutong Bai$^{1}$,
Maggie Chung$^{10,\ddagger}$,
Adam Yala$^{2,\ddagger}$
}

\begin{document}

\maketitle

\noindent
$^{1}$ Department of Electrical Engineering and Computer Science, UC Berkeley, USA\\
$^{2}$ Computational Precision Health, UC Berkeley and UC San Francisco, USA\\
$^{3}$ Department of Medical Imaging and Intervention, Chang Gung Memorial Hospital at Linkou, Taiwan\\
$^{4}$ Department of Electrical Engineering and Computer Science, Massachusetts Institute of Technology, USA\\
$^{5}$ Department of Medical Imaging and Radiological Sciences, Chang Gung University, Taiwan\\
$^{6}$ Clinical Metabolomics Core and Imaging Core Laboratory, Institute for Radiological Research, 
Chang Gung Memorial Hospital at Linkou and Chang Gung University, Taiwan\\
$^{7}$ Mass General Brigham Cancer Institute, USA\\
$^{8}$ Massachusetts General Hospital, USA\\
$^{9}$ Harvard Medical School, USA\\
$^{10}$ Department of Radiology and Biomedical Imaging, UC San Francisco, USA

\bigskip

\noindent
$^\ast$ Project lead\\
$\dagger$ Core contributor \textit{(these authors contributed equally; order among core contributors was determined at random)}\\
$\ddagger$ Co-senior author

\abstract
\textbf{Abstract}

Radiology plays an integral role in modern medicine, yet rising imaging volumes have far outpaced workforce growth, contributing to burnout and challenges in care delivery. Foundation models offer a path toward assisting with the full spectrum of radiology tasks, but existing medical models remain limited: they process volumetric CT and MRI as low-fidelity 2D slices, discard critical grayscale contrast information, and lack evaluation frameworks that reflect real clinical practice.
Here, we introduce \textbf{\name}, a radiology foundation model pretrained on 42,990 abdomen-pelvis CTs, 86,411 chest CTs, 14,348 head CTs, and 11,543 breast MRIs from a large academic center, together with \textit{RATE}, a scalable framework that extracts structured labels for 366 radiologic findings with near-perfect accuracy using large language models. Across internal test sets of 14,230 abdomen-pelvis CTs, 10,646 chest CTs, 4,906 head CTs, and 1,585 breast MRIs, \name establishes a new performance frontier, achieving mean AUROCs of 86.4, 88.0, 90.1, and 82.9, outperforming MedGemma (Google), MedImageInsight (Microsoft), Lingshu (Alibaba), and Merlin (Stanford) by 7.8-15.8 AUROC points and ranking best in 87.2\% (319/366) tasks. 
\name similarly outperforms all baselines in an external validation on the Stanford abdomen-pelvis CT dataset, including Merlin (82.2 vs 80.6 AUROC), which uses the Stanford dataset for development. 
\name extends to tasks beyond its pretraining, such as long-horizon lung cancer risk prediction, where it improves upon the state-of-the-art Sybil by 3.0 C-index points on NLST, and generalizes with gains of 5.9 (MGH) and 1.9 (CGMH).
In brain hemorrhage detection, \name obtained a >95 AUROC when using only $\frac{1}{20}$ of the data of the next most sample efficient baseline. 
\name and RATE together provide an open, clinically rigorous foundation for building high-performance radiology systems, enabling applications that were previously infeasible due to computational, data, and evaluation constraints.

\endabstract

\setcounter{footnote}{0}

\newpage

\section{Main}
Radiology serves a key role in modern clinical practice, as it allows for the visualization of disease and guides patient management. Imaging utilization has continued to grow significantly year over year, with studies reporting annual growth rates ranging from 5 to 7\% \cite{fortunebusinessinsights, bccresearch}. This growth has far outpaced the expansion of the radiology workforce, resulting in radiologist burnout and challenges in traditional patient care delivery models \cite{radiologistshortage, rimmer2017radiologist, jing2025radiology, mirak2025}. Although numerous artificial intelligence (AI) tools have been proposed to improve the detection of pathology on imaging studies, including commercially available tools for the detection of lung nodules \cite{hendrix2023nodules} and intracranial hemorrhage \cite{fang2025hemorrhage}, their impact on overall radiology efficiency is limited. These tools assist with only a small fraction of radiologists' tasks. In practice, radiologists perform comprehensive image interpretation with a wide range of findings across all organ systems, modalities, and protocols \cite{dogra2025generalist, jones2021, incidental, bertrand2010incidental}. Assisting with this workload requires technology that can address the full spectrum of image findings.

Foundation models learn broad, transferable representations from diverse datasets and therefore hold promise for enabling comprehensive image interpretation \cite{evo2, wang2022omnivlonefoundationmodelimagelanguage}. An ideal radiology foundation model would 1) enhance performance across a wide range of downstream tasks, including classification, localization, prognosis, and report generation; 2) drastically reduce the amount of training data required for finetuning; and 3) serve as a de-facto platform for downstream model development. Despite extensive effort \cite{blankemeier2024merlin, wu2023generalistfoundationmodelradiology, sellergren2025medgemma, codella2024medimageinsight, xu2025lingshu, zhang2025biomedclipmultimodalbiomedicalfoundation}, these goals remain largely unrealized for computed tomography (CT) and magnetic resonance imaging (MRI) due to several challenges.

\textbf{Challenges in Modeling Volumetric Imaging.}
From a computational perspective, the primary challenge in modeling volumetric medical imaging is resolution. Medical volumes are immense in both spatial scale and bit-depth. For example, CT scans of the abdomen-pelvis often contain $512 \times 512 \times 768$ pixels and are over 4,000 times larger than normal $224 \times 224$ ImageNet images \cite{imagenet}. The cost of traditional vision transformers scales quadratically with input size, making direct 3D modeling computationally prohibitive. Consequently, leading foundation models from Google (MedGemma)\cite{sellergren2025medgemma}, Microsoft (MedImageInsight)\cite{codella2024medimageinsight}, and Alibaba (Lingshu)\cite{xu2025lingshu} process CT exams as independent 2D slices, ignoring volumetric structure and losing essential contextual information. We hypothesize that leveraging the native 3D structure of CT exams has the potential to deliver significant performance improvements. Moreover, CT scans are acquired in 12-16 bit voxels, capturing up to 65,536 grayscale values, which are significantly richer than the 8-bit (256-value) pixels found in ImageNet \cite{imagenet}. Radiologists routinely apply specialized windowing strategies to dynamically view different tissue types (e.g., bone, soft-tissue, lung, etc.) with different intensity ranges \cite{bae2005windowing, pellakuru2025windowing}. However, leading industry models downsample these rich voxels to 8-bit, resulting in loss of subtle contrast \cite{codella2024medimageinsight}. We hypothesize that preserving the dynamic range of volumetric imaging would similarly improve the capability of radiology models.

\textbf{Limitations of Current Evaluation Frameworks.}
Progress in developing radiology foundation models rests on the rigor of their evaluation. However, the leading evaluation benchmarks exhibit limitations that parallel the shortcomings of the models they seek to assess.
Visual question-answering benchmarks like VQA-RAD \cite{lau2018dataset} and SLAKE \cite{liu2021slake} convert high-dimensional medical volumes into downsampled 2D slices stored as 8-bit JPEGs and pair them with simplified questions that often do not reflect real diagnostic tasks. PMC-VQA \cite{zhang2023pmc} compounds this problem by sourcing image-text pairs from scientific publications rather than clinical imaging, limiting their relevance to routine clinical practice. 
These benchmarks fail to capture the richness of inputs (i.e., full resolution DICOM volumes) or the diversity of clinically meaningful outputs (i.e., wide range of imaging findings) across modalities and indications \cite{hu2024omnimedvqa, yue2024mmmu, zuo2025medxpertqa}. As a result, researchers cannot answer the core clinical question: whether a model can detect hundreds of diverse imaging findings within high-resolution 3D volumes, as radiologists do in clinical practice. Consequently, a shortage of rigorous benchmarks hinders the development of robust radiology foundation models. We hypothesize that improved evaluation frameworks, designed to leverage the full complexity of radiology data, are critical for catalyzing meaningful progress in radiology foundation models. 

Given these challenges, there remains an unmet need for general-purpose radiology foundation models. State-of-the-art models for tasks ranging from disease detection \cite{hirsch2025bmr} to risk prediction \cite{mikhael2023sybil} continue to rely on natural-image pretraining or manual feature engineering (e.g., radiomics). To address this gap, we introduce \name, a radiology foundation model that advances the broad frontier of CT and MRI understanding, improving performance across hundreds of radiological tasks.

\begin{figure}[p]
\centering
\captionsetup{justification=justified, skip=6pt}
\includegraphics[width=\linewidth]{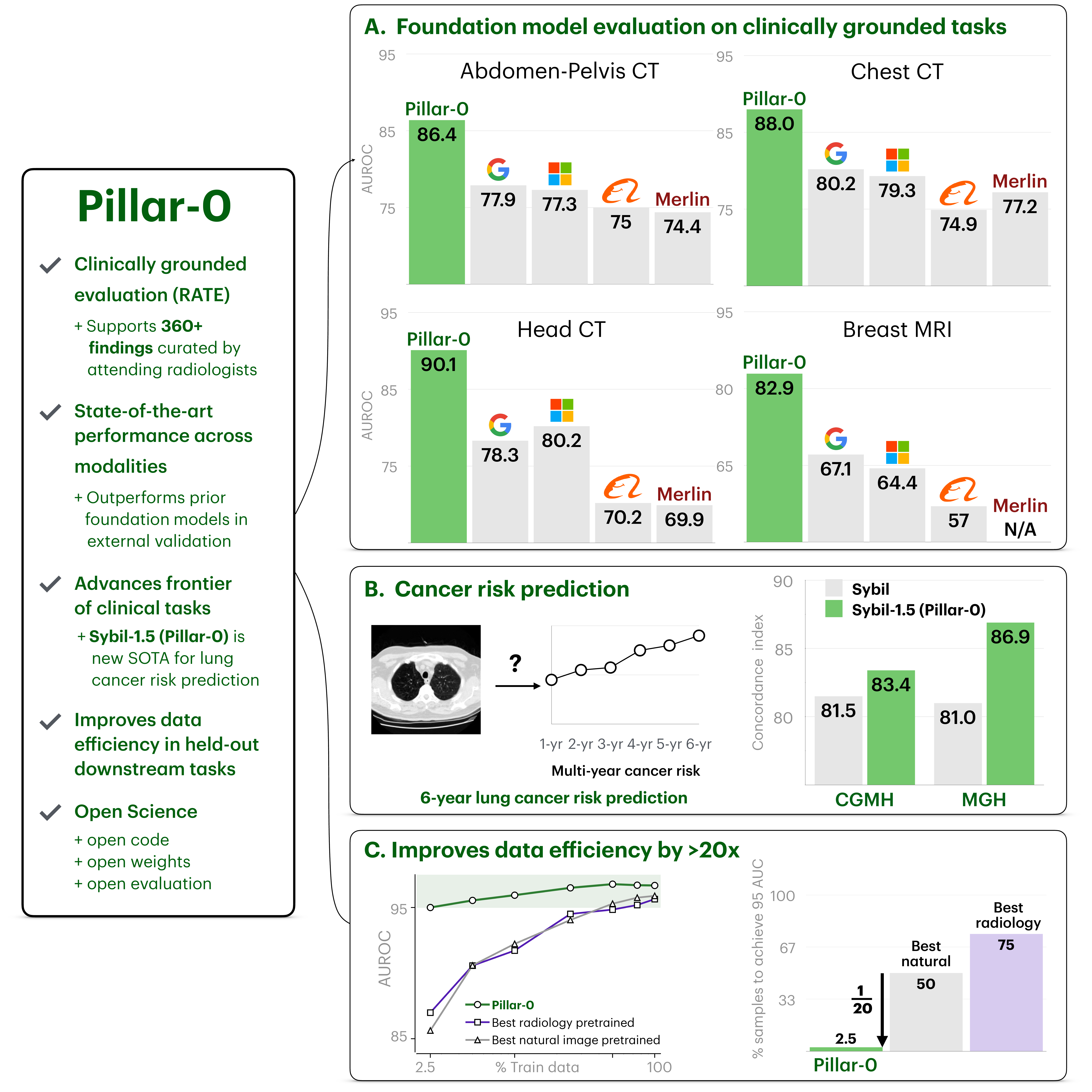} 
\caption{Overview of \name and key results across modalities and tasks. (A) We evaluate \name on abdomen-pelvis CT, chest CT, head CT, and breast MRI using RATE, a clinically grounded evaluation framework designed to overcome the limitations of existing radiology benchmarks. \name substantially outperforms competitive baselines across modalities. (B) \name's capabilities extend to real-world clinical prediction tasks outside the standard of care, setting a new state-of-the-art for future lung cancer risk prediction with Sybil-1.5 (\name finetuned). On rigorous multi-institution external validation, Sybil-1.5 outperforms Sybil \cite{mikhael2023sybil}, a strong specialist baseline, by a wide margin. (C) Finally, \name demonstrates superior data efficiency, reaching 95 AUROC on the RSNA Intracranial Hemorrhage detection benchmark \cite{rsna-intracranial-hemorrhage-detection} using {20-30$\times$} less training data relative to best-in-class natural image-pretrained (Swin3D-t \cite{yang2023swin3}) and radiology-pretrained (Merlin \cite{blankemeier2024merlin}) models. The entire \name system---open-code, open-weights, and open-evaluation---is released to the community.}
\label{fig:pillar0_overview}
\end{figure}

\newpage 

Our work provides the following contributions: 

\begin{enumerate}
    \item \textbf{Clinically Grounded Evaluation Framework.} We develop \textit{\textbf{Ra}diology \textbf{T}ext \textbf{E}ngine} (\textit{RATE}), a clinically grounded evaluation framework for medical volumes. A team of board-certified radiologists curated a broad list of 366 radiological findings abdomen-pelvis CT, chest CT, head CT, and breast MRI exams, and validated that open large language models (LLMs) such as Qwen3 \cite{qwen3embedding} could extract the findings from radiology reports with near perfect accuracy. RATE provides an efficient LLM interface to automatically extract large labeled datasets of imaging findings from unstructured radiology reports, establishing the foundation for reproducible model comparison.
    
    \item  \textbf{Dominant Performance on Internal Test Sets.} We release \name, a radiology foundation model pretrained across CT exams of the abdomen-pelvis (n=42,990), chest (n=86,411), and head (n=14,348), as well as 11,543 breast MRI exams. For internal validation, we evaluated \name on held-out imaging exams spanning head CT (n=4,906), chest CT (n=10,646), abdomen-pelvis CT (n=14,230), and breast MRI (n=1,585). \name obtained an average AUROC of 90.1\footnote{While AUROCs and concordance (C-)indices range from 0-1, we report all AUROCs and C-indices as 100x the value (i.e., 90.1 instead of 0.901) for better legibility.}, 88.0, 86.4, and 82.9 across these modalities, reflecting an absolute improvement of 7.8–15.8 over the next best model. Across 366 evaluated tasks, \name was the top-performing model in 87.2\% (319/366), outperforming MedGemma \cite{sellergren2025medgemma}, MedImageInsight \cite{codella2024medimageinsight}, Lingshu \cite{xu2025lingshu}, and Merlin \cite{blankemeier2024merlin}.

    \item \textbf{Outperforming Foundation Models in External Validation.} 
    We externally validated \name on a CT abdomen-pelvis dataset from Stanford (USA)\cite{blankemeier2024merlin}, previously used to develop the Merlin CT foundation model. 
    Here, \name outperformed MedGemma, MedImageInsight, Lingshu, and Merlin. Moreover, when we retrained \name using only Merlin's training data, \name still outperformed Merlin (82.2 vs 80.6), despite leveraging fewer forms of supervision.
    
    \item \textbf{Improving Ability to Predict Future Cancer from Asymptomatic Screening.}  
    Sybil \cite{mikhael2023sybil} previously demonstrated that low-dose CTs could be used to predict the risk of future lung cancer. This prediction task, which humans cannot perform, is outside the distribution of \name ’s pretraining. By finetuning \name, we significantly outperformed the original Sybil model in predicting lung cancer risk. This improvement generalized to the same external validation test sets used to validate the original Sybil, Massachusetts General Hospital (MGH, Boston, USA) and Chang Gung Memorial Hospital (CGMH, Taipei, Taiwan), with condordance index \cite{cindex} improvements of 5.9 (81.0 to 86.9) and 1.9 (81.5 to 83.4) respectively.
    
    \item \textbf{Reducing Downstream Training Data Needs by >20x.} We showed that \name significantly improves the sample efficiency of building radiology AI tools with a case study on brain hemorrhage detection using the RSNA Brain CT challenge dataset \cite{rsna-intracranial-hemorrhage-detection}. When finetuning foundation models for hemorrhage detection, we found that \name achieved dominant performance across all data regimes. Moreover, \name reached >95 AUROC when using only $\frac{1}{20}$ of the data of the next most sample-efficient baseline.
 
    \item \textbf{Open Science.} To enable the broader research community to benefit from our advances, we release a broad suite of open-source tools, including evaluation, volume preprocessing, model pretraining, finetuning and inference repositories. We also release all pretrained models.
\end{enumerate}

\section{Results}
\setcounter{subsection}{-1} 
\subsection{Overview of \name and Core Innovations}

\begin{figure}[!ht]
\centering
\captionsetup{singlelinecheck=false,skip=6pt}
\includegraphics[width=\linewidth]{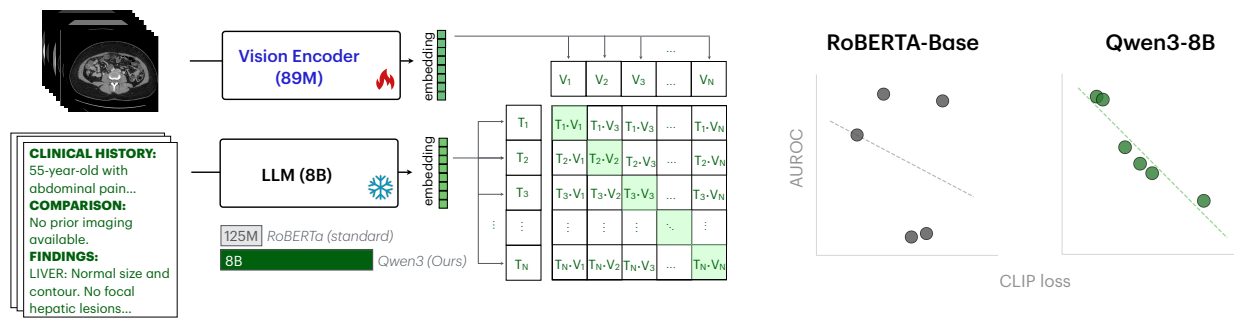}
\caption{\name key innovations across tokenization, architecture, and pretraining. (A) Modality-specific multi-windowing converts full-resolution CT and MRI volumes into multi-channel inputs that emulate radiologist workflow presets, preserving clinically relevant contrast. Training with multi-windowing leads to a 4.6 point gain in AUROC in abdomen-pelvis CT. (B) The Atlas vision backbone employs hierarchical Multi-Scale Attention to efficiently process long-context volumes \cite{agrawal2025atlasmsa}. As a result, \name is 175$\times$ faster than ViT-S, and achieves state-of-the-art performance with fewer parameters than other medical foundation models. (C) Asymmetric contrastive pretraining aligns Atlas volume embeddings with embeddings from a much larger frozen LLM text encoder. Using this powerful text encoder leads to a much stronger correlation between CLIP loss and downstream performance, providing a reliable signal for clinical utility to guide pretraining experiments.} 
\label{fig:pillar0_ablations}
\end{figure}

We introduce \name, a general-purpose foundation model for volumetric medical imaging, built on key innovations across tokenization, model architectures, and pretraining (Figure \ref{fig:pillar0_ablations}). To leverage the full bit-depth of CT and MRI scans, we propose custom tokenizers inspired by radiology workflows. Rather than using a single 8-bit range, we project volume patches into multiple intensity range windows to highlight different tissue properties across image channels (Figure \ref{fig:pillar0_ablations}a).
Next, to reason over large spatial context, we leverage multi-scale attention and the Atlas neural network architecture \cite{agrawal2025atlasmsa} (Figure \ref{fig:pillar0_ablations}b). Atlas processes abdomen-pelvis CT scans at 175$\times$ the speed of a comparable vision transformer.
Finally, to distill rich radiology report supervision during \name pretraining, we leverage asymmetric contrastive learning analogous to CLIP \cite{radford2021learningtransferablevisualmodels}, where we learn to align Atlas volume representations with Qwen-8B \cite{qwen3embedding} text representations (Figure \ref{fig:pillar0_ablations}c). Using large text encoders, such as modern LLMs, instead of smaller models (e.g., RoBERTa; <500M parameters) \cite{liu2019robertarobustlyoptimizedbert}, boosts radiology foundation model performance and yields a much stronger correlation between pretraining loss and downstream results, enabling more predictable scaling. Additional dataset and methodological context is provided in Sections \ref{sec:methods_dataset} and \ref{sec:training_methods}.

\subsection{RATE: Clinically Grounded Evaluation Framework }

\begin{figure}[t]
\centering
\captionsetup{singlelinecheck=false,skip=6pt}
\includegraphics[width=\linewidth]{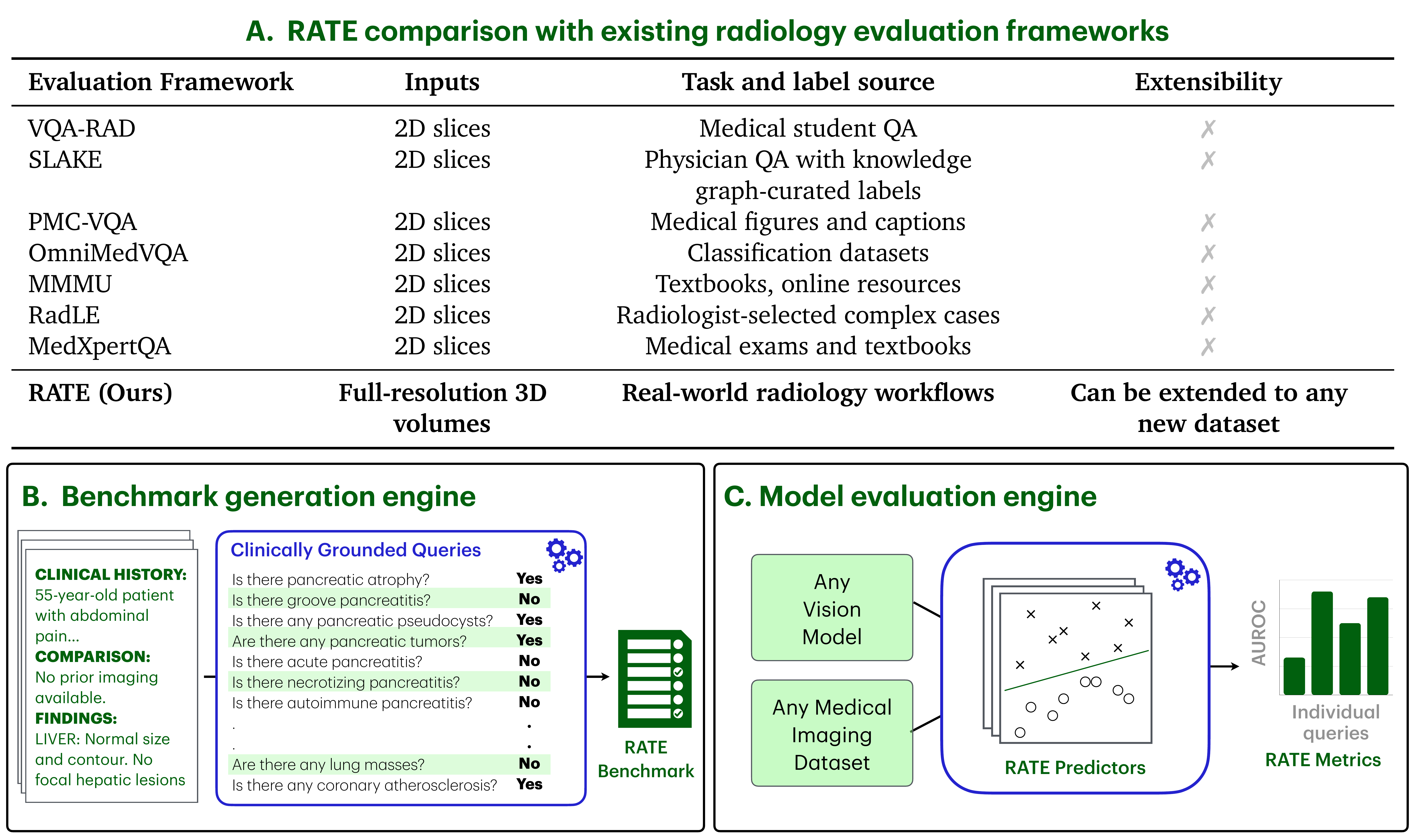}
\caption{RATE: a clinically grounded evaluation framework for volumetric radiology. (A) Comparison of RATE to existing benchmarks (VQA-RAD \cite{lau2018dataset}, SLAKE \cite{liu2021slake}, PMC-VQA \cite{zhang2023pmc}, OmniMedVQA \cite{hu2024omnimedvqa}, MMMU \cite{yue2024mmmu}, RadLE \cite{datta2025radiology}, MedXpertQA \cite{zuo2025medxpertqa}) along three axes: inputs, task and label source, and extensibility. RATE is the only framework that takes full-resolution volumes as input, uses clinically grounded tasks with labels derived from routine clinical practice, and can be extended to any radiology image-report dataset. (B) RATE’s benchmark generation engine applies a large language model to unstructured radiology reports to extract answers to a set of clinically grounded queries. (C) RATE-Evals, the model evaluation engine, provides a standardized evaluation protocol with per-task linear probing (RATE Predictors) for any vision model. The inputs to the engine are full-resolution medical volumes and RATE-extracted labels from corresponding reports. Together, RATE and RATE-Evals enable extensible, clinically aligned evaluation of radiology foundation models.}
\label{fig:pillar0_rate}
\end{figure}

We introduce RATE, a unified framework designed to evaluate any vision model on full-fidelity medical volumes, using authentic clinical tasks derived from real-world radiology practice (Figure \ref{fig:pillar0_rate}).

Existing radiology benchmarks fall short along three key dimensions (Figure \ref{fig:pillar0_rate}a). First, none use full-resolution volumetric data---instead, most use 2D slices in the format of JPEG images \cite{lau2018dataset, liu2021slake}, or snapshots from medical textbooks \cite{yue2024mmmu, zuo2025medxpertqa}. Second, the task definitions and accompanying labels are not sourced from routine clinical practice. 
Task definitions often do not reflect the typical responsibilities of radiologists, and instead rely on hand-crafted or automated prompts which can be as simple as identifying the imaging modality \cite{zhang2023pmc, lau2018dataset}. 
Labels are typically synthetic or manually annotated, rather than extracted from normal clinical workflows \cite{datta2025radiology}. Finally, they are not extensible, offering only a fixed dataset without a pathway to expand or adapt tasks \cite{hu2024omnimedvqa, datta2025radiology}.

RATE addresses all three key limitations. 
Board-certified radiologists 
curated 366 diverse radiologic findings reflecting real-world practice across our modalities. RATE then uses open large language models to extract binary labels for each task from radiology reports, enabling scalable and clinically grounded benchmarking (Figure \ref{fig:pillar0_rate}b). Full details are provided in Section \ref{sec:methods-RaTE}.

Building on this framework, we introduce RATE-Eval, a standardized protocol for assessing pretrained vision encoders on real-world medical imaging datasets. RATE-Evals employs the binary labels generated by RATE in a linear probe setup, freezing the encoder and training a single linear classifier per task over its embeddings (Figure \ref{fig:pillar0_rate}c). Performance on held-out exams measures the quality and transferability of the learned representations, analogous to CLIP-style linear evaluation \cite{radford2021learning}. Additional details appear in Section \ref{sec:methods-linear-probe}.

\subsection{\name Obtains Dominant Performance on Internal Test Sets}

\begin{figure}[htbp]
\centering
\captionsetup{singlelinecheck=false,skip=6pt}
\includegraphics[width=\linewidth]{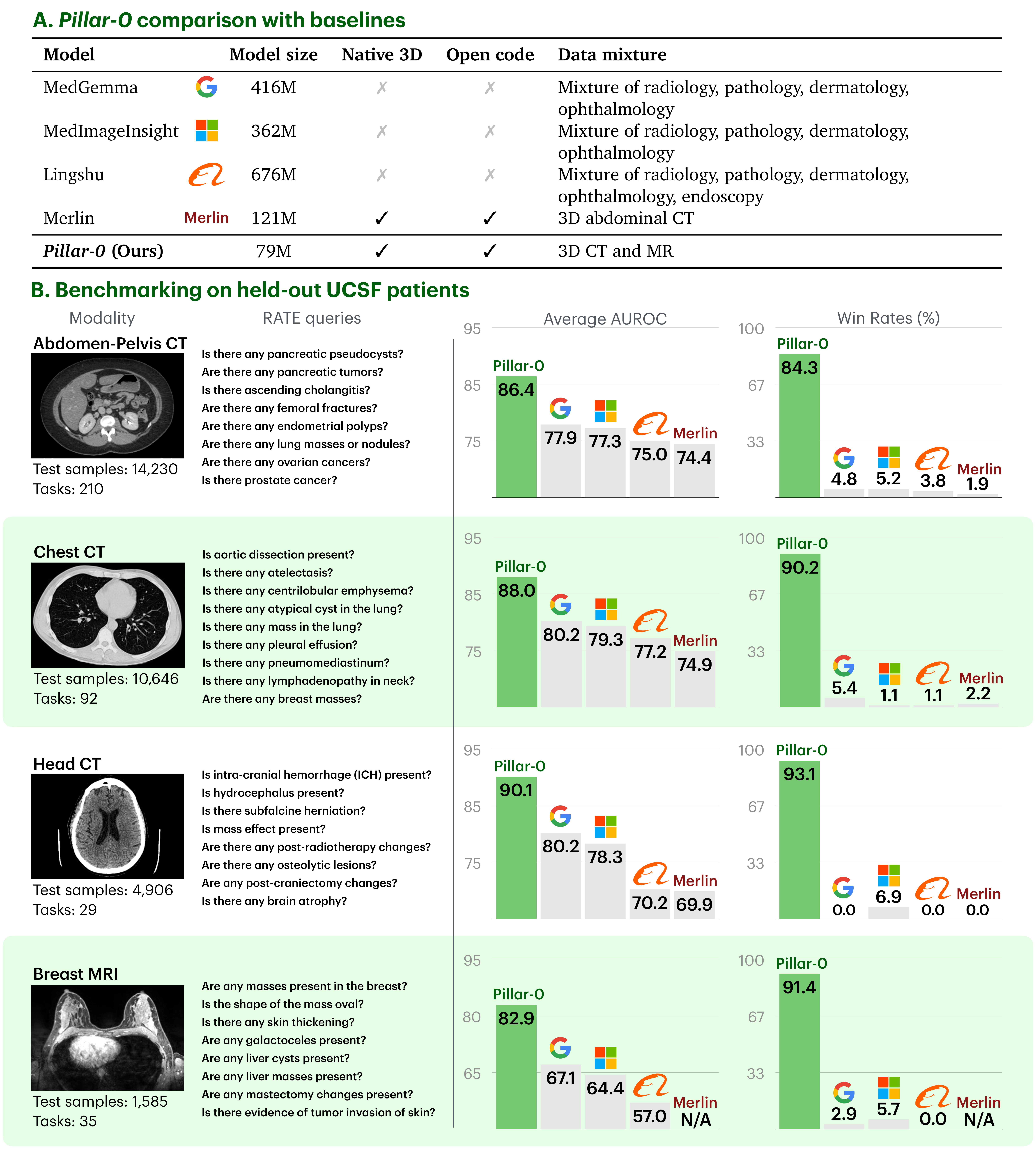}
\caption{\name achieves dominant performance over MedGemma, MedImageInsight, Lingshu, and Merlin on internal UCSF test sets across abdomen-pelvis CT, chest CT, head CT, and breast MRI. For each modality, \name attains the highest average AUROC, with modality-level AUROC improvements of 7.8-15.8 points over the closest baseline. Aggregated over modalities, \name wins on 319 of 336 findings (87.2\%), winning at least 84.3\% in every modality.}
\label{fig:pillar0_internal}
\end{figure}

\name achieves dominant overall performance across all evaluated modalities in UCSF held-out test sets. We compare \name against four competitive baselines representing the current state of medical imaging models (Figure \ref{fig:pillar0_internal}a). MedGemma, Lingshu, and MedImageInsight are large 2D medical models trained on diverse imaging mixtures, but do not natively process volumetric inputs. Merlin \cite{blankemeier2024merlin} is an open-source 3D model trained on an institutional dataset of abdomen-pelvis CTs. 
Additional information on the baselines is provided in Section~\ref{sec:methods-baselines}.

We assess \name and all baselines using binary labels for 366 radiologist-curated findings derived through RATE, and apply the RATE-Evals framework for model comparison. The UCSF held-out test sets include 14,230 abdomen-pelvis CT, 10,646 chest CT, 4,906 head CT, and 1,585 breast MRI exams, providing a diverse, clinically grounded benchmark. Full details of the evaluation protocol are provided in Section~\ref{sec:methods-internal-eval}.

As shown in Figure~\ref{fig:pillar0_internal}b, \name outperforms all baselines by a large margin in each modality. 
Overall, \name performs the best on 319/366 questions (87.2\%), far exceeding all baselines (MedGemma, 16/366, 4.3\%; MedImageInsight, 16/366, 4.3\%; Lingshu, 9/366, 2.5\%; Merlin, 6, 1.6\%). At the modality level, \name achieves an average AUROC computed across all questions in each modality of 86.4, 88.0, 90.1, and 82.9 for abdomen-pelvis CT, chest CT, head CT, and breast MRI respectively. In comparison, the second strongest model in each modality is MedGemma, which attains 77.9, 80.2, 80.2 and 67.1 average AUROC, meaning \name 
beats the closest baseline by 7.8-15.8 points. Compared to all models, \name's question-level win rates are 84.3\% for abdomen–pelvis CT, 90.2\% for chest CT, 93.1\% for head CT, and 91.4\% for breast MRI, consistently surpassing those of all baselines. No baseline exceeds 6.9\% in any modality.
Compared to MedGemma head-to-head, \name's question level win rates are 90.4\% for abdomen–pelvis CT, 92.4\% for chest CT, 96.6\% for head CT, and 94.3\% for breast MRI. 
A detailed question-level comparison between \name and MedGemma is provided in Section \ref{sec:medgemma-comparison}. Per-finding metrics can be found in Sections \ref{ref:abd-finding} (abdomen-pelvis CT), \ref{ref:chest-finding} (chest CT), \ref{ref:brain-finding} (head CT) and \ref{ref:breast-finding} (breast MRI).

\subsection{\name Outperforms Foundation Models on External Test Sets}

\name demonstrates strong external generalization, outperforming all baselines evaluated on the Stanford Merlin Abdominal CT Dataset \citep{blankemeier2024merlin}. This dataset, which was previously used to develop Merlin, contains 25,494 abdomen-pelvis CT–report pairs from 18,317 patients. Using RATE, we extracted 202 clinically relevant findings from this cohort, and evaluated all models with RATE-Evals (see Section~\ref{sec:methods-external-eval} for details). \name achieved an average AUROC of 82.2, outperforming Merlin (80.6), for which this dataset is internal, as well as MedGemma (72.6), MedImageInsight (74.9), and Lingshu (72.1) (Table \ref{tab:merlin}).



This dataset also provides an opportunity for a detailed head-to-head comparison with the Merlin model, which was trained using both radiology reports and electronic health record codes.
To isolate the effect of data source, we trained \textit{Pillar-0 (Stanford Only)} using the \name recipe with the Stanford Merlin Abdominal CT dataset. Despite leveraging only text supervision, \textit{Pillar-0 (Stanford Only)} still outperforms Merlin (82.2 vs 80.6), showing that the gains are attributable to improved methods. Moreover, \textit{Pillar-0 (Stanford Only)} performs similarly to \name, demonstrating strong generalization. 

Finally, we evaluated \name as an initialization for specialist foundation model training by constructing \textit{Pillar-0 (UCSF + Stanford)}, which initializes from \name, and is further pretrained on the  Stanford Merlin Abdominal CT data with the \name recipe. We find that \textit{Pillar-0 (UCSF + Stanford)} substantially outperforms Merlin (84.9 vs 80.6), indicating that models built on top of \name benefit from a superior initialization. Additional details are noted in Section \ref{sec:methods-external-eval}.

\begin{table}[h]
\centering
\caption{\name outperforms all baselines on external validation on the Stanford Merlin Abdominal CT Dataset. Notably, \name outperforms Merlin, which was developed using this dataset. \textit{Pillar-0 (Stanford Only)}, pretrained with the \name recipe on the Stanford data alone, also outperforms Merlin. \textit{Pillar-0 (UCSF + Stanford)}, which is initialized from \name and then finetuned on the Stanford dataset, pushes performance even further, establishing best average AUROC by a wide margin.}
\label{tab:merlin}
\begin{tabularx}{\linewidth}{l *{5}{>{\centering\arraybackslash}X}}
\toprule
\textbf{Model} &
\textbf{Dataset} &
\textbf{Average AUROC on Merlin RATE-Eval} \\
\midrule
MedGemma & Mixture of medical imaging  & 72.6 \\
MedImageInsight & Mixture of medical imaging  & 74.9 \\
Lingshu & Mixture of medical imaging & 72.1 \\
\midrule
Merlin (Stanford) & Merlin-Abd-CT & 80.6 \\ 
{\bf {\textit{Pillar-0 (Stanford Only)}}} & Merlin-Abd-CT & {\bf 82.2} \\
\midrule 

{\bf {\textit{Pillar-0}}} & UCSF-Abd-CT & {\bf 82.2} \\
{\bf {\textit{Pillar-0 (UCSF + Stanford)}}}  & UCSF-Abd-CT + Merlin-Abd-CT & {\bf 84.9} \\
\bottomrule
\end{tabularx}
\end{table}

\subsection{\name Improves the Ability to Predict Future Cancer from Asymptomatic Screening}

In addition to replicating tasks captured in routine radiology reports, \name can significantly improve performance in tasks beyond the current standard of care. Here, we study this capability in the context of lung cancer screening, where we train models to predict future cancer risk from a single low-dose CT (LDCT). 

Lung cancer screening with LDCT reduces lung cancer-specific mortality by 20\% among patients with a history of tobacco smoking \cite{nlst}. More accurate prediction of future lung cancer risk could enhance screening efficiency by personalizing follow-up intervals, maintaining long-term engagement, and identifying high-risk individuals who may not meet traditional smoking-based eligibility criteria. Sybil \cite{mikhael2023sybil} was recently developed to predict lung cancer risk from a single LDCT and has been clinically validated across multiple healthcare settings, including in patients who have never smoked \cite{sybilexternal}.

\begin{figure}[h]
\centering
\captionsetup{singlelinecheck=false,skip=6pt}
\includegraphics[width=\linewidth]{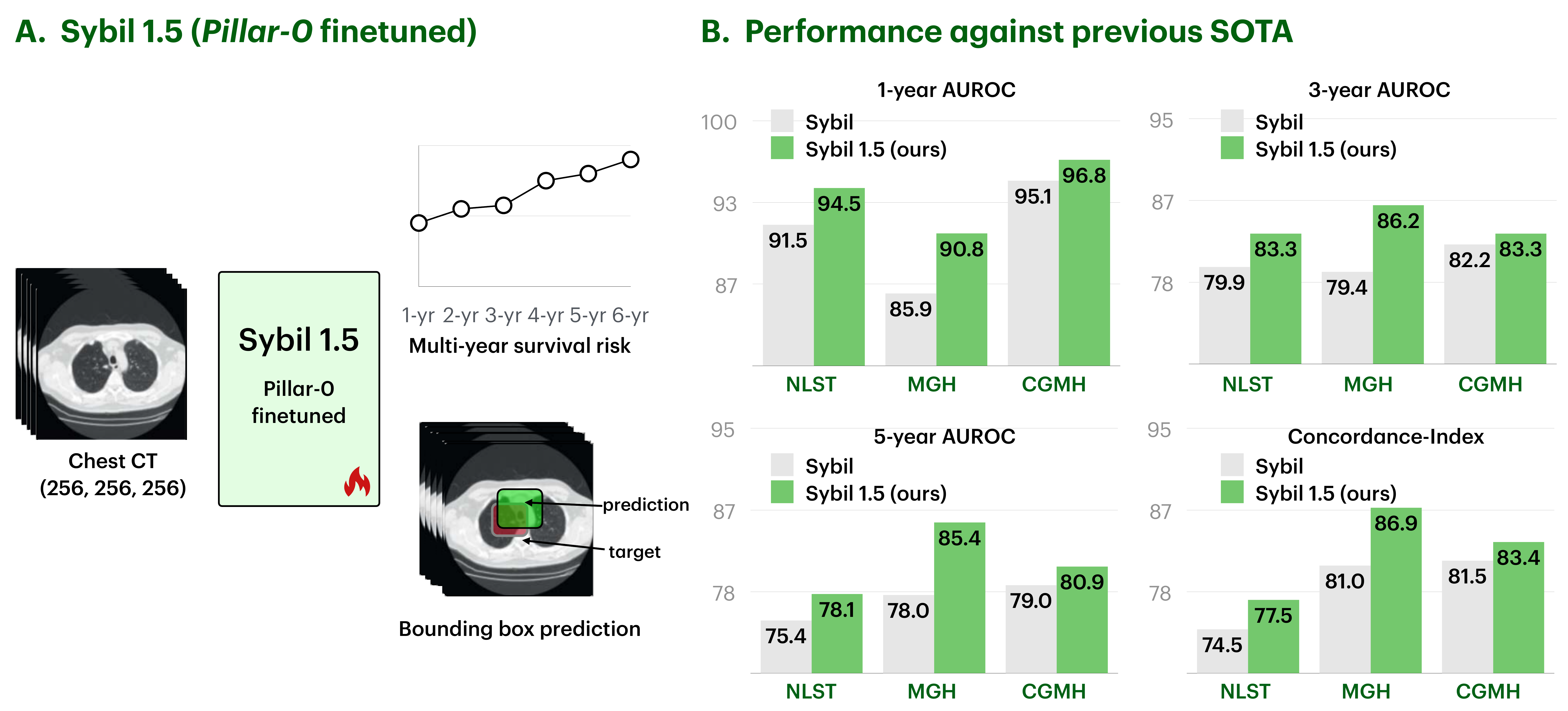}
\caption{Finetuning \name sets a new state-of-the-art for future lung cancer risk prediction. (A) Illustration of Sybil-1.5 (\name finetuned), trained on chest CTs and annotations from NLST to predict multi-year cancer risk and bounding boxes of suspicious regions. (B) Performance of Sybil and Sybil-1.5 on NLST, MGH, and CGMH cohorts, reported as 1-, 3-, and 5-year AUROC and 6-year overall concordance index. Across all datasets and time horizons, Sybil-1.5 improves risk stratification over Sybil.}
\label{fig:sybil}
\end{figure}

We finetuned \name on LDCTs from the National Lung Screening Trial (NLST) to develop Sybil-1.5 (\name finetuned), a new state-of-the-art model for lung cancer risk prediction. Sybil-1.5 predicts the location of suspicious lesions and outputs six annual risk scores corresponding to lung cancer diagnoses 1–6 years after screening (Figure \ref{fig:sybil}a). We assessed performance using Uno’s concordance index \cite{cindex} and AUROC across each year, and validated the model on the same NLST held-out test set used to evaluate Sybil (N=6,282), as well as two external cohorts: MGH (8,821 LDCTs, Protocol 2020P002652) and CGMH (12,280 LDCTs, IRB202301073B0). Additional training and evaluation details are provided in Section \ref{sec:methods-sybil}.

Across all test sets and evaluation metrics, Sybil-1.5 (\name finetuned) consistently outperforms Sybil (Figure \ref{fig:sybil}b). On the NLST held-out test set, Sybil-1.5 improves 1-year AUROC compared to Sybil from 91.5 (95\% CI 87.8 to 95.2) to 94.5 (95\% CI 92.0 to 96.9) (p=0.04). Sybil-1.5 also yields better generalization to external validation sets. For MGH and CGMH, the 1 year AUROC improves from 85.9 (95\% CI 82.6 to 89.2) to 90.8 (95\% CI 88.5 to 93.1) (p<0.001) and from 95.1 (95 \% CI 91.2 to 99.0) to 96.8 (95 \% CI 91.2 to 99.0) (p=0.03). We find that Sybil-1.5 performs well across race, sex, age, and smoking status, shown in detailed subgroup analyses in the supplemental materials (Table \ref{tab:group_auc}).

\subsection{\name Reduces Downstream Training Data Needs in Brain Hemorrhage Detection by >20$\times$}

\begin{figure}[!htbp]
\centering
\captionsetup{singlelinecheck=false,skip=6pt}
\includegraphics[width=\linewidth]{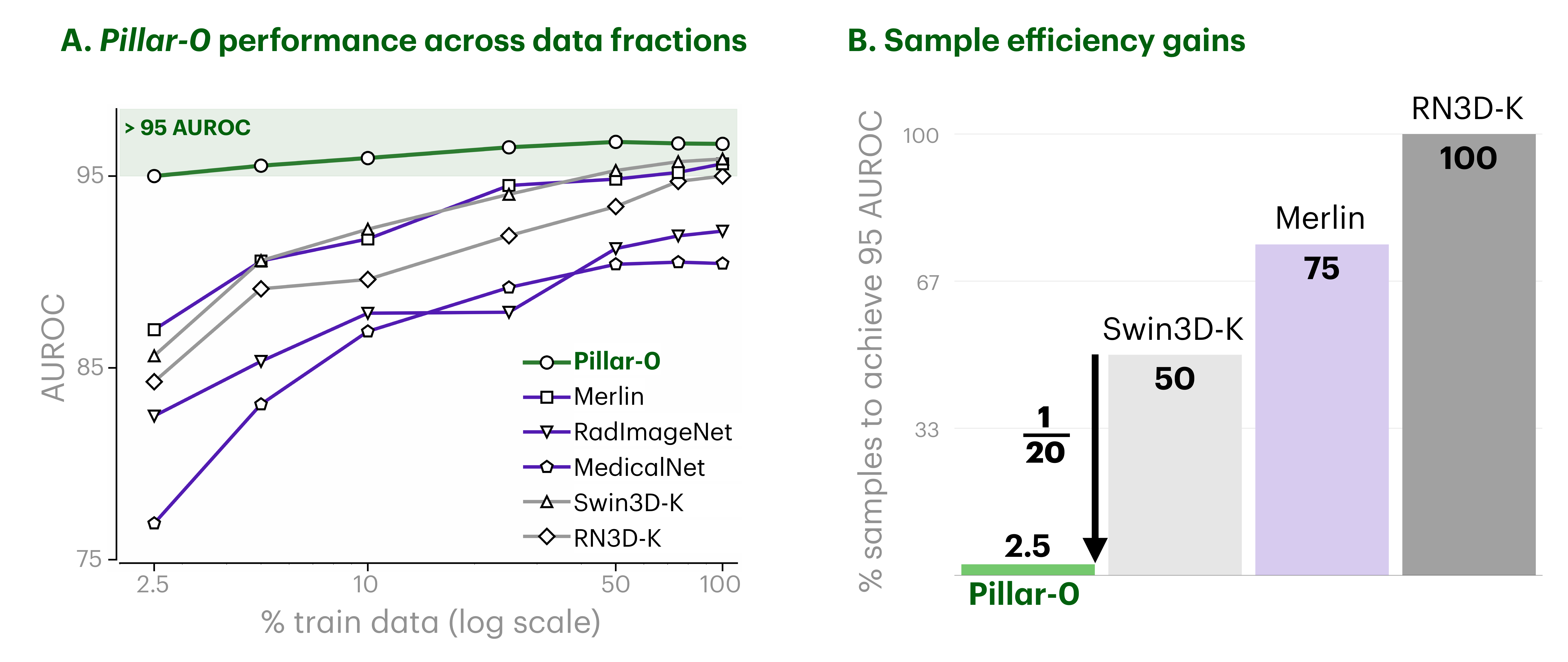}
\caption{\name dramatically improves data efficiency for brain hemorrhage detection on RSNA-2019. (A) AUROC for detecting any intracranial hemorrhage as a function of the fraction of training exams (2.5-100\%). Across all data fractions \name outperforms radiology-pretrained models (Merlin, RadImageNet \cite{radimagenet}, MedicalNet \cite{chen2019med3d}; purple) and natural image-pretrained models (Swin3D-K, RN3D-K; gray). With only a small fraction of the training data, \name matches or exceeds the best baselines trained on the full dataset. (B) Fraction of training data required for each model to reach an AUROC of 95.0; \name reaches this threshold with 20-40$\times$ fewer samples than baselines. MedicalNet and RadImageNet do not attain this performance even with the full dataset.}
\label{fig:sample_eff}
\end{figure}

\name achieves substantial gains in sample efficiency, outperforming competitive baselines using only a small fraction of the labeled data (Figure~\ref{fig:sample_eff}). We evaluated \name and baselines on the 2019 RSNA Brain Hemorrhage Detection Challenge dataset (RSNA-2019) \cite{rsna-intracranial-hemorrhage-detection}, following the protocol described in Section~\ref{sec:methods-sample-eff}. RSNA-2019 comprises 21,744 unique head CT exams from three institutions (Stanford University, Universidade Federal de São Paulo, and Thomas Jefferson University Hospital), with neuroradiologist annotations for the presence of five intracranial hemorrhage (ICH) subtypes.

In this case study, we finetuned \name along with two general-purpose 3D backbones (Swin3D-t and 3D ResNet-18; Kinetics-400 pretraining) and three radiology-specific models (MedicalNet 3D ResNet-18 \cite{chen2019med3d}, RadImageNet ResNet-50 \cite{radimagenet}, and Merlin \cite{blankemeier2024merlin}) to predict the presence or absence of each of the ICH subtypes, and the presence or absence of any ICH overall.  
We trained models using 2.5-100\% of the available training data, and report AUROC on the "any ICH" task as the primary metric.

Across all data fractions, \name outperforms every baseline (Figure \ref{fig:sample_eff}a). With just 2.5\% of the training dataset, \name achieves an AUROC of 95.0, outperforming the other baselines at this data fraction by 8.0-18.1 AUROC points. Swin3D-t, the best-performing baseline, requires 50\% of the samples to achieve this performance, while Merlin and ResNet3D require 75\% and 100\%, respectively, representing a 20- to 40-fold improvement in sample efficiency (Figure \ref{fig:sample_eff}b). 

Using 10\% of the training dataset, \name matches or outperforms the best baselines using the full dataset, and exceeds the other baselines by 3.4–10.5 AUROC points at this data fraction. With just 2.5\% of the training dataset, \name comes within 1 AUROC point of the best baseline model using the full dataset (Swin3D-t). Full results across all data fractions are shown in Table \ref{tab:sample-eff}.


\subsection{Ablation Study}
\label{sec:ablation}

To identify the core components driving the performance of \name, we conduct ablations along three principal axes: tokenization, model architecture, and pretraining objective. We use the Stanford Merlin CT dataset as a testbed to make our ablations fully reproducible. Across these experiments, we find that \name’s gains arise not from any single design choice, but from the combined effect of modality-aware tokenization, an efficient multi-scale 3D architecture, and a strong asymmetric contrastive objective.

{\bf Tokenization: Multi-Windowing.}
We evaluate our modality-native multi-window tokenization against a standard min–max normalization baseline (Figure \ref{fig:pillar0_overview}a). Multi-windowing substantially improves downstream performance, achieving an AUROC of 82.2 on Merlin RATE-Evals compared to 77.6 with min–max scaling. This highlights the importance of preserving subtle contrast information for CT understanding.


\textbf{Architecture Ablation: Multi-Scale Attention for 3D Efficiency.} To assess architectural efficiency for high-resolution volumetric inputs, we compare our multi-scale Atlas backbone to a standard ViT-S \cite{dosovitskiy2020image} model using identical input volumes ($384 \times 384 \times 384$) and patch sizes ($6 \times 6 \times 6$). Under this matched configuration, ViT-S requires approximately 38.8 seconds per sample at inference, whereas Atlas requires only 0.2 seconds, delivering a 175$\times$ speedup with similar parameter counts (79M for \name vs. 123M for ViT-S; Figure \ref{fig:pillar0_ablations}b). This demonstrates that multi-scale attention enables practical 3D modeling. Compared to MedGemma, MedImageInsight, and Lingshu, which process slices independently, \name is 32, 15 and 55$\times$ faster, respectively.


\textbf{Pretraining Ablation: Text Encoder.}
We study the impact of encoder asymmetry in contrastive learning by pairing the vision encoder with either a standard text encoder (RoBERTa-Base \cite{liu2019robertarobustlyoptimizedbert}, 125M parameters) or a large-scale LLM-based encoder (Qwen3 \cite{qwen3embedding}, 8B). Using Qwen3 yields a substantially stronger correlation between contrastive training loss and downstream RATE performance (Pearson r = $-$0.256 vs. $-$0.947; Figure \ref{fig:pillar0_ablations}c), enabling the loss to serve as a reliable proxy for clinical utility. The larger text encoder also improves overall performance, increasing downstream AUROC from 76.6 to 82.2 on Merlin Rate-Evals. Together, these results highlight the importance of a high-capacity text encoder in producing clinically aligned representations.



\section{Discussion}
Foundation models have transformed diverse areas of AI, including natural language processing \cite{devlin2019bert}, computer vision \cite{lecun2015dl}, and biology \cite{evo2}, by offering general-purpose backbones for downstream developers. To advance the field, these tools should improve performance on a broad spectrum of tasks, including those outside of their pretraining distribution. Moreover, they should reduce the required data to reach a target performance level. The development of these technologies rests on rigorous evaluation benchmarks that test models on realistic inputs and tasks that closely mirror real-world use. Despite extensive recent work in radiology foundation models, leading systems neither model the rich 3D structure of volumetric medical imaging nor evaluate their performance on clinically relevant benchmarks.

In this work, we present \name, a radiology foundation model designed to advance comprehensive understanding of CT and MRI imaging. To support rigorous evaluation, we introduce RATE, a clinically grounded framework that leverages large language models to extract structured labels for hundreds of radiologic findings from unstructured reports. RATE enables reproducible, scalable benchmarking across modalities by aligning model evaluation with radiologist-prioritized findings and diagnoses. Using RATE, we show that \name establishes a new performance frontier across abdomen-pelvis CT, chest CT, head CT, and breast MRI. We show that \name outperforms leading 2D and 3D medical models, and that high-fidelity volumetric representations yield substantially more transferable features. \name generalizes strongly across institutions and provides a superior initialization for downstream development, outperforming Merlin on its own in-domain dataset, and exceeding its performance even when pretrained on the same data. In addition to radiologist-performed tasks, \name enables risk modeling tasks beyond human perceptual limits. When finetuned on NLST data, \name sets a new state of the art for long-horizon lung cancer risk prediction and generalizes across external cohorts. \name also offers marked gains in sample efficiency. In brain hemorrhage detection, it matches or exceeds the best baselines trained on the full RSNA-2019 dataset using only 10\% of the training set, reducing data requirements by more than an order of magnitude. Ablations show that these gains arise from the combination of multi-window tokenization, an efficient multi-scale 3D backbone, and asymmetric contrastive pretraining with a strong text encoder, which together produce predictable, clinically aligned scaling behavior.

The clinical relevance of these advances is considerable. 
\name’s innovations across tokenization, architecture, and pretraining allow it to leverage the rich 3D spatial context and intensity patterns encoded in volumetric CT and MRI.
This yields higher performance on both core diagnostic tasks and tasks that exceed human perceptual capabilities, such as long-horizon cancer risk prediction. By reducing downstream training data requirements by more than an order of magnitude, \name addresses one of the most persistent barriers in clinical AI: the scarcity of large, high-quality labeled datasets in many radiologic subspecialties. This sample efficiency makes it feasible to build accurate specialist models even in domains with limited, heterogeneous, or costly annotations. Combined with its strong external generalization and superior utility as an initialization for downstream model development, \name is suited to serve as a robust foundation for radiology AI and enable applications that were previously infeasible. By releasing all models, code, and the RATE framework, we substantially lower the computational and data barriers required to build volumetric AI systems. This democratization enables even resource-constrained groups to build high-quality models with far less data, accelerating progress across the field.

While \name substantially advances the frontier of volumetric radiology foundation models, there remain obvious opportunities for further improvement. First, our pretraining data was from a single large tertiary academic medical center, using only $218,217$ CTs and MRIs in total, and we leveraged a small (89M parameter) vision encoder. We expect that significantly scaling data and model capacity will unlock further model improvements. Moreover, the scanner vendors, acquisition protocols, patient demographics, and disease prevalence may not fully capture the diversity of imaging practices and patient populations seen across other healthcare systems. Larger and more diverse datasets will be important for further improving model generalizability. Second, RATE extracted binary clinical labels from radiology reports. This approach inherits the inherent limitations of report-based supervision: radiologists may omit normal or incidental findings, and textual descriptions do not always reflect the full spectrum of imaging appearance \cite{incidental, bertrand2010incidental}. Additionally, treating missing mentions as negative labels, while consistent with radiology reporting behavior, introduces a potential source of label noise. RATE does not directly assess localization, segmentation, and temporal evolution, which are critical for many clinical applications. There remain opportunities to expand RATE to capture other clinically grounded tasks, offering rich feedback for the development of radiology foundation models. Finally, \name solely relies on contrastive pretraining, omitting many additional sources of supervision, including full report generation and additional clinical context. 

\enlargethispage{2\baselineskip}
\newpage
\appendix
\section{Methods}

\setcounter{subsection}{-1} 
\subsection{\name}
\label{sec:methods_pillar_0}

\subsubsection{Development Dataset}
\label{sec:methods_dataset}

We identified 71,510 abdomen-pelvis CT, 107,923 chest CT, 24,042 head CT, and 14,742 breast MRI exams performed between 2001 and 2025 in adult patients at UCSF. Exams were retrieved using the institutional radiology database (mPower Clinical Analytics; Nuance Communications) and the Automated Image Retrieval platform. 
For the breast MRI dataset, we employed regular expressions on the series descriptions to retain T1 fat saturation, T2 fat saturation, and peak contrast-enhanced series, disambiguating between post-contrast series using acquisition time. For the CT datasets, we first excluded biopsies and non-reportable exams. To eliminate redundant series, we grouped series by acquisition time and selected the axial series with the lowest slice thickness. For exams where multiple series remain, we select a series at random.
The dataset was  divided patient-wise into training, validation, and test splits with complete exam counts in Table \ref{tab:dataset-splits}. Tables \ref{tab:brain-demographics}, \ref{tab:ab-demographics}, \ref{tab:chest-demographics}, and \ref{tab:breast-demographics} contain detailed breakdowns of patient demographics and scanner models for each modality.

\begin{table}[htbp]
\centering
\caption{Summary of pretraining datasets. The count of exams per split is depicted for each modality.}
\label{tab:dataset-splits}
\begin{tabular}{@{}lrrrrr@{}}
\toprule
\textbf{Split} & \textbf{Abdomen-Pelvis CT} & \textbf{Chest CT} & \textbf{Head CT} & \textbf{Breast MRI} & \textbf{All Datasets} \\
\midrule
{Train} & $42,990$ & $86,411$ & $14,348$ & $11,543$ & $155,292$ \\
{Validation} & $14,290$ & $10,866$ & $4,788$ & $1,614$ & $31,558$ \\
{Test} & $14,230$ & $10,646$ & $4,906$ & $1,585$ & $31,367$ \\
\midrule
{Total} & $71,510$ & $107,923$ & $24,042$ & $14,742$ & $218,217$ \\
\bottomrule
\end{tabular}
\end{table}

\begin{table}[htbp]
\centering
\caption{Dataset characteristics for head CT. The age value is provided as mean ± standard deviation. Gender is provided as percentages of the total patients (n=19,118). Manufacturer is provided as the percentage of total exams (n=24,042).}
\label{tab:brain-demographics}
\begin{tabular}{@{}lrr@{}}
\toprule
\textbf{\textit{Head CT}} & & \\
\midrule
\textbf{Demographics} & \textbf{Patients (n=19,118)} & \textbf{Value} \\
\midrule
\textbf{Age} & - & $61.2 \pm 19.4$ \\
\textbf{Gender} & &  \\
Female & $9,364$ & $48.98\%$ \\
Male & $9,717$ & $50.83\%$  \\
\midrule
\textbf{Manufacturer} & \textbf{Exams (n=24,042)} & \textbf{Value}\\
\midrule
GE Medical Systems & $23,716$ & $98.64\%$ \\
Siemens & $229$ & $0.95\%$ \\
Siemens Healthineers & $37$ & $0.15\%$ \\
Philips & $24$ & $0.10\%$ \\
Samsung & $21$ & $0.09\%$ \\
Agfa & $12$ & $0.05\%$ \\
TOSHIBA & $2$ & $0.01\%$ \\
NeuroLogica & $1$ & $0.00\%$ \\
\bottomrule
\end{tabular}
\end{table}

\begin{table}[htbp]
\centering
\caption{Dataset characteristics for abdomen-pelvis CT. The age value is provided as mean ± standard deviation. Gender is provided as percentages of the total patients (n=45,483). Manufacturer is provided as the percentage of total exams (n=71,510).}
\label{tab:ab-demographics}
\begin{tabular}{@{}lrr@{}}
\toprule
\textbf{\textit{Abdomen-Pelvis CT}} & & \\
\midrule
\textbf{Demographics} & \textbf{Patients (n=45,483)} & \textbf{Value} \\
\midrule
\textbf{Age} & - & $58.8 \pm 17.1$ \\
\textbf{Gender} & &  \\
Female & $22,876$ & $50.23\%$ \\
Male & $22,556$ & $49.59\%$  \\
\midrule
\textbf{Manufacturer} & \textbf{Exams (n=71,510)} & \textbf{Value} \\
\midrule
GE Medical Systems & $65,289$ & $91.30\%$ \\
Siemens & $3,442$ & $4.81\%$ \\
Philips & $2,433$ & $3.40\%$ \\
Siemens Healthineers & $273$ & $0.38\%$ \\
TOSHIBA & $70$ & $0.10\%$ \\
Agfa & $2$ & $0.00\%$ \\
AMICAS Inc. & $1$ & $0.00\%$ \\
\bottomrule
\end{tabular}
\end{table}

\begin{table}[htbp]
\centering
\caption{Dataset characteristics for chest CT. The age value is provided as mean ± standard deviation. Gender is provided as percentages of the total patients (n=49,775). Manufacturer is provided as the percentage of total exams (n=107,923).}
\label{tab:chest-demographics}
\begin{tabular}{@{}lrr@{}}
\toprule
\textbf{\textit{Chest CT}} && \\
\midrule
\textbf{Demographics} & \textbf{Patients (n=49,775)} & \textbf{Value} \\
\midrule
\textbf{Age} & - & $60.6 \pm 17.1$\\
\textbf{Gender} & &  \\
Female & $23,661$ & $47.54\%$ \\
Male & $26,066$ & $52.37\%$  \\
\midrule
\textbf{Manufacturer} & \textbf{Exams (n=107,923)} & \textbf{Value} \\
\midrule
GE Medical Systems & $90,761$ & $84.10\%$\\
Siemens & $4,773$ & $4.42\%$ \\
AGFA & $4,374$ & $4.05\%$ \\
AGFA Healthcare & $1,467$ & $1.36\%$ \\
Visage PR & $488$ & $0.45\%$ \\
Siemens Healthineers & $405$ & $0.38\%$ \\
Philips & $159$ & $0.15\%$ \\
AGFA Healthcare Informatics & $54$ & $0.05\%$ \\
Imbio & $37$ & $0.03\%$ \\
Bunkerhill & $31$ & $0.03\%$ \\
TERARECON & 3 & $0.00\%$ \\
Philips Medical Systems & 2 & $0.00\%$ \\
Unknown & $5,368$ & $4.97\%$ \\
\bottomrule
\end{tabular}
\end{table}

\begin{table}[htbp]
\centering
\caption{Dataset characteristics for breast MRI. The age value is provided as mean ± standard deviation. Gender is provided as percentages of the total patients (n=6,444). Manufacturer is provided as the percentage of total exams (n=14,742).}
\label{tab:breast-demographics}
\begin{tabular}{@{}lrr@{}}
\toprule
\textbf{\textit{Breast MRI}} & & \\
\midrule
\textbf{Demographics} & \textbf{Patients (n=6,444)} & \textbf{Value} \\ 
\midrule
\textbf{Age} & - & $50.4 \pm 12.9$\\
\textbf{Gender} && \\
Female & $6,418$ & $99.60\%$ \\
Male & $18$ & $0.28\%$ \\
\midrule

\textbf{Manufacturer} & \textbf{Exams (n=14,742)} & \textbf{Value}\\
\midrule
GE Medical Systems & $9,551$ & $64.79\%$ \\
Siemens & $4,502$ & $30.54\%$ \\
Invivo & $137$ & $0.93\%$ \\
AGFA & $119$ & $0.81\%$ \\
Visage PR & $22$ & $0.15\%$ \\
Siemens Healthineers & $17$ & $0.12\%$ \\
Siemens & $9$ & $0.06\%$ \\
Visage Imaging & $6$ & $0.04\%$ \\
AGFA Healthcare & $2$ & $0.01\%$ \\
Unknown & $377$ & $2.56\%$ \\
\bottomrule
\end{tabular}
\end{table}

\newpage 
\subsubsection{\name Training Recipe}
\label{sec:training_methods}
\paragraph{Architecture.} 
The backbone of \name is the Atlas\cite{agrawal2025atlasmsa} vision encoder, which leverages Multi-Scale Attention to enable efficient long-context image modeling. \name is trained on full-resolution medical volumes and leverages fine-scale patch sizes (Table \ref{tab:modality_patch_size}). This configuration ensures anatomical coverage while maintaining sensitivity to small findings: $6^3$ patches preserve fine lesions (on the order of a few millimeters), while our input resolution is large enough to capture the entire field of view. As shown in Section \ref{sec:ablation}, training at this resolution is untenable for other transformer-based architectures. The Atlas backbone replaces $O(N^2)$ self-attention with an $O(N \log N)$ multi-scale mechanism, making high-resolution full-volume 3D training computationally feasible. We use a 3-stage Atlas-S configuration, with 2 blocks in each of the first 2 stages, and 8 blocks in the last stage.

\begin{table}[htbp]
\centering
\caption{Patch size and resolution by imaging modality. Our inputs range from 32k-256k tokens per volume after patchification, necessitating the use of an efficient architecture for long-context modeling (Atlas).}
\begin{tabular}{lccc}
\toprule
Modality         & Patch size & Resolution (HWD) & Tokens per volume \\
\midrule
Abdomen-pelvis CT &    $6 \times 6 \times 6$    &  $384 \times 384 \times 384$    &    256k       \\
Chest CT          &    $8 \times 8 \times 4$  &  $256 \times 256 \times 256$   & 64k       \\
Head CT           &   $8 \times 8 \times 4$    &  $256 \times 256 \times 128$ & 32k   \\
Breast MRI        &   $12 \times 12 \times 6$   &  $384 \times 384 \times 192$  & 32k \\
\bottomrule
\end{tabular}
\label{tab:modality_patch_size}
\end{table}

\paragraph{Radiology-specific tokenizer.}
CT volumes are calibrated in physically meaningful Hounsfield units (HU), spanning a wide dynamic range (roughly $-1000$ to $+3000$ HU). Critical structures occupy subspaces of this dynamic range, but standard display technologies make it impossible to render all anatomically relevant structures with optimal contrast at once. Rather than a single global min-max normalization or single window, we emulate radiologist practice via modality-specific \emph{multi-windowing}. For each CT modality, we define a small set of anatomically motivated window presets (e.g., lung, soft tissue, mediastinum, bone). Each preset is applied independently to the raw HU volume, clipped to the specified window width and level, linearly rescaled to $[0, 1]$, and treated as a separate channel. The model thus receives a multi-channel view in which different channels emphasize different anatomical structures, analogous to a radiologist scrolling through window presets at the workstation.

Breast MRI lacks a standardized physical unit and is typically acquired as multiple complementary series (e.g., T1-weighted, T2-weighted, diffusion-weighted). For MRI we therefore adopt an adaptive \emph{high-contrast windowing} strategy. For each series, we compute a foreground intensity histogram and set the window to span the 1st to 99th percentile of this distribution, followed by linear rescaling to $[0, 1]$. This yields a robust, data-driven normalization that maximizes contrast for relevant tissues across heterogeneous scanners and protocols. Section \ref{sec:ablation} shows that this modality-specific, multi-window representation substantially outperforms a simple min-max baseline. The code for this functionality is included in our vision engine, with details in Section \ref{sec:RaVE}.

\paragraph{Report preprocessing and text encoder.}
For the inputs to the text encoder, we use radiology reports processed by the RATE pipeline (Section \ref{sec:methods-RaTE}). We first remove the \emph{comparisons} section, which describes longitudinal changes relative to prior exams and is not directly observable from a single study. We then extract the \emph{findings} section verbatim, isolating the radiologist’s image-centric description while excluding clinical history, acquisition parameters, and administrative metadata. The resulting findings text serves as a clean, clinically grounded caption for vision-language pretraining. Each findings section is tokenized and passed through the frozen Qwen3-Embedding-8B model; the pooled text representation is then projected into the shared embedding space by a learned linear layer.

\paragraph{Pretraining pipeline.}
The first stage in our pretraining pipeline is supervised classification on ImageNet-1K upsampled to $1,024 \times 1,024$ resolution. Concretely, we optimize a cross-entropy loss with the AdamW optimizer \cite{loshchilov2017decoupled}, a global batch size of 2,048, and weight decay 0.24, using cosine learning rate decay and a linear warmup schedule. Training is performed for 320 epochs (30 epoch warmup) on  32$\times$H100 GPUs, reaching 80.1\% top-1 validation accuracy after approximately 33.5 hours. The resulting checkpoint serves as the initialization for all subsequent vision-language pretraining runs.

Starting from the ImageNet-pretrained weights, we perform single-modality vision-language pretraining separately for abdomen-pelvis CT, chest CT, head CT, and breast MRI. We find empirically that single-modality pretraining is easier to tune compared to cross-modality pretraining. For each modality, we instantiate a vision encoder and a small projection head. The text encoder is a frozen Qwen3-Embedding-8B large language model \cite{qwen3embedding}, which is substantially larger than the vision encoder (8B vs 79M). We refer to this training setup as \textit{asymmetric contrastive learning}, in which only the vision encoder and the projection layers are updated. 
We utilize a contrastive learning objective for vision-language pretraining, analogous to CLIP, which encourages matched volume–report pairs to have high similarity while treating all other combinations in the batch as negatives.

Contrastive learning benefits from large batches, but high-resolution 3D volumes quickly exhaust GPU memory. To obtain an effective global batch size of up to 256, far exceeding what fits on a single 8$\times$H100 node, we accumulate features from 8 mini-batches to compose the full similarity matrix, and use gradient accumulation across mini-batches, considering examples from all other mini-batches as negatives. In particular, we use a mini-batch size of 32 with AdamW optimizer and maximum learning rate of $2.5 \times 10^{-4}$, and cosine learning rate schedule with linear warm-up of 200 steps. 

\paragraph{Model release.}
We release all modality-specific vision-language pretrained checkpoints for \name via HuggingFace at \url{https://huggingface.co/collections/YalaLab/pillar-0}. In addition, we release our vision-language pretraining code at \url{https://github.com/YalaLab/pillar-pretrain}.

\subsection{RATE: Clinically Grounded Evaluation Framework}
\label{sec:methods-RaTE}
\subsubsection{RATE}

We developed Radiology Text Engine (RATE) to convert unstructured radiology reports into structured artifacts which can be used for evaluating vision encoders on medical volumes. RATE takes a radiology report as input and produces binary clinical labels corresponding to an expert-curated set of Yes/No questions capturing clinically relevant findings for each exam. A key feature of this system is its extensibility: new modalities and question sets can be incorporated by providing additional radiologist-specified queries. 

RATE operates through a pipeline built on a single large language model (Qwen3-30B-A3B-FP8) using specific prompts. Using this setup, the system identifies binary Yes/No answers to expert-curated questions directly from the report text, generating structured labels suitable for evaluating clinical finding identification. The framework also produces records that facilitate quality control, allowing researchers to review and validate the extracted labels.

Additionally, the system enables the extraction of clinically grounded captions for vision-language pretraining. To do this, it removes the "comparisons" section, which describes longitudinal changes, and then extracts the "findings" section verbatim.  

We release the full RATE implementation, including prompts, templates, and evaluation utilities at \url{https://github.com/YalaLab/rate}.

\subsubsection{Using RATE to evaluate any vision encoder}
\label{sec:methods-linear-probe}

RATE-Evals is a modular framework to systematically assess the performance of any vision encoder on medical imaging tasks. It is designed with extensibility in mind: we provide simple templates for datasets and models that can be easily modified to support additional use-cases. We release the evaluation code at \url{https://github.com/YalaLab/rate-evals}.

We evaluate models using linear probing on the embeddings to predict the binary clinical labels generated by RATE. This approach measures how well frozen vision encoders support curated clinical tasks, serving as a direct indicator of the quality and transferability of the extracted representations. For each model, image embeddings are extracted from the frozen encoder and used to train a lightweight linear classifier that predicts the binary clinical labels defined by RATE. This is formulated as a multi-label classification problem in which each question is treated as an independent binary target. 

Optimization is performed with Adam~\cite{kingma2017adammethodstochasticoptimization} optimizer with learning rate $10^{-3}$, batch size 8,192, and no weight decay for 1,000 epochs using a class-balanced binary cross-entropy loss to correct for label imbalance. Performance is summarized as per-question AUROC.

Because radiology reports often omit the mention of normal or absent findings, missing answers (i.e., answers to the queries are not mentioned in the reports) are treated as negative by default. Alternatively, the system supports a \textit{masking mode} that restricts training of the linear probes and analysis to explicitly labeled samples. 

RATE-Evals has a registry-based design that allows new datasets and models to be added as needed. It includes built-in support for the Merlin Abdominal CT Dataset \citep{blankemeier2024merlin}, as well as a lightweight synthetic dataset for rapid iteration. It also provides implementations for \name, and all external baselines evaluated in this work (Section \ref{sec:methods-baselines}). 

\subsubsection{Baselines}
\label{sec:methods-baselines}

\paragraph{MedGemma.} Multimodal 4B/27B models with a SigLIP-400M \cite{zhai2023sigmoidlosslanguageimage} vision encoder paired with an LLM; also a 27B text-only variant. Training data is comprised of medical imaging datasets including MIMIC-CXR (chest X-rays), proprietary de-identified CT and MRI exams (represented as sets of 2D slices), and additional data from histopathology, dermatology, and ophthalmology. Training follows a multi-stage procedure beginning with adaptation of the vision encoder on medical image–text pairs, followed by multimodal pre-training and post-training. In our implementation, each volume is first min-max normalized. Following MedGemma's documentation, we resize each slice to $896 \times 896$ and pass it through the vision encoder, treating the depth dimension as the batch dimension. We then extract features from the vision encoder, and apply average pooling to produce a single representation per volume.

\paragraph{MedImageInsight.} DaViT-based vision transformer~\cite{ding2022davit} trained on public and proprietary datasets including chest X-ray, CT, MRI, ultrasound, histopathology and additional domains. The model uses CLIP-style contrastive learning with the UniCL objective~\cite{yang2022unified} to align paired image and text embeddings. Following MedImageInsight's official implementation, we perform median pooling from the spatial features from the vision tower to
produce a single representation per volume.

\paragraph{Merlin.} A 3D ResNet image encoder coupled with a transformer-based text encoder. The training data consists of abdominal CT exams linked to EHR diagnosis labels and associated radiology reports. The training objective combines binary cross-entropy loss to predict diagnosis codes with an InfoNCE loss to align 3D CT volumes with the unstructured report text. Our implementation is based on Merlin's official implementation, and was checked with Merlin author Ashwin Kumar.

\paragraph{Lingshu.} Built on the Qwen2.5-VL backbone~\cite{bai2025qwen2} with a ViT-based vision encoder~\cite{dosovitskiy2020image}. The training data includes CT, MRI, ultrasound, histopathology, and other medical modalities. For volumetric modalities (e.g., CT/MRI), each 2D slice is processed independently. Training follows a four-stage pipeline consisting of: (1) medical shallow alignment, (2) deep alignment, (3) instruction tuning, and optionally (4) reinforcement learning. Our reproduction uses \href{https://huggingface.co/lingshu-medical-mllm/Lingshu-7B}{https://huggingface.co/lingshu-medical-mllm/Lingshu-7B} processes each slice by resizing to $896 \times 896$, applying min-max normalization, extracting encoder features, and average pooling to produce a single representation per volume.

The implementations of all our baseline models are in RATE-Evals.



\subsection{Internal Validation}
\label{sec:methods-internal-eval}

For internal evaluation, we used the UCSF held-out test set described in Section \ref{sec:methods_dataset} comprised of 14,230 abdomen-pelvis CT, 4,906 head CT, 10,646 chest CT, and 1,585 breast MRI. Using RATE (Section~\ref{sec:methods-RaTE}), 366 radiologist-curated clinical findings were extracted: 210 abdomen-pelvis CT, 29 head CT, 92 chest CT, and 35 breast MRI findings. We then applied the RATE-Evals (Section \ref{sec:methods-RaTE}) protocol to compute performance for \name and for all baseline models described in Section~\ref{sec:methods-baselines}.

To assess label quality, standardized quality control was performed on the binary clinical labels extracted from each modality. Board-certified radiologists manually reviewed RATE outputs for 20 randomly sampled reports per modality (80 total) and adjudicated the corresponding Yes/No findings. We observed 100\% agreement between RATE-derived labels and radiologist adjudications in this sample, supporting the use of RATE labels as high-fidelity evaluation data. Questions with no positive examples in the test set were excluded from performance estimates.



\subsection{External Validation}
\label{sec:methods-external-eval}

For external evaluation, we used the Merlin Abdominal CT Dataset \citep{blankemeier2024merlin}, which comprises 25,494 abdomen-pelvis CT–report pairs from 18,317 patients. The cohort consists of exams acquired in the Stanford Hospital Emergency Department between 2012–2018, identified using abdomen-pelvis CT CPT codes (72192–74178) via the STARR tool. For each exam, the DICOM series with the largest slice count was selected and converted to a compressed, de-identified NIfTI volume. As defined in the original dataset splits, the held-out test set used for evaluation contains 5,137 CT exams.

We applied RATE using the same abdomen-pelvis CT query set used in our internal evaluation (Section~\ref{sec:methods-internal-eval}). After excluding questions with no positive instances in the Merlin test split, this yielded 202 radiologist-curated clinical findings. 

For this analysis, we also pretrained two variants of \name. The first followed the complete \name pretraining recipe (Section~\ref{sec:training_methods}) but used only the Merlin training corpus to isolate the effect of data source. The second began from the pretrained \name checkpoint (Section~\ref{sec:training_methods}) and underwent an additional 14 epochs of contrastive pretraining on the Merlin dataset (early stopping in 50-epoch run) to assess how much performance can be gained by building on \name. We evaluated all baselines, \name, and both \name variants using  RATE-Evals (Section \ref{sec:methods-linear-probe}).

\subsection{Lung Cancer Case Study}
\label{sec:methods-sybil}
We developed Sybil-1.5 by finetuning \name on the same data used to develop Sybil\cite{mikhael2023sybil}. We applied for and were granted access to the radiologic and clinical data from a sample of 15,000 National Lung Screening Trial (NLST) \cite{nlst} participants in the LDCT arm, including all lung cancers in that arm. All NLST participants signed an institutional review board (IRB)–approved informed consent form.  Following Sybil's data and image suitability protocol, we used a dataset of 28,162 LDCTs in the training set, 6,839 LDCTs in the development set, and 6,282 LDCTs in the test set, with 1,444 (5.1\%), 337 (4.9\%), and 299 (4.8\%) positive LDCTs, corresponding to lung cancers diagnosed over the subsequent 6 years, respectively.  Following IRB-approved protocols, we also used 8,821 LDCTs from MGH (Protocol 2020P002652), including 169 (1.9\%) confirmed cancers. From CGMH (IRB202301073B0), we used 12,280 LDCTs from CGMH, including 101 (0.8\%) cancers.
CT volumes were cropped to $256 \times 256 \times 256$ and multi-windowing with 11 windows was used.

We built upon the \name vision encoder by adding two components: (i) a cumulative probability layer that outputs year-by-year cancer risk and (ii) a DETR-based head for bounding box prediction \cite{carion2020detr}. For bounding box prediction, all multi-scale vision features are interpolated to the spatial resolution of the finest feature map and concatenated along the channel dimension. For risk prediction, attention pooling is applied to the finest-scale features to produce a global representation. The entire model was then trained end-to-end using the original Sybil survival and attention regularization losses \cite{mikhael2023sybil} combined with DETR’s bounding-box regression and matching losses \cite{carion2020detr}. 

We release our complete code for finetuning \name for flexible downstream use-cases at \url{https://github.com/YalaLab/pillar-finetune}.

\subsection{Sample Efficiency Case Study}
\label{sec:methods-sample-eff}
We used 21,744 unique exams from the published RSNA-2019 training set, and further split patient-wise into 15,166, 3,261, and 3,317 exams for training, validation, and test, respectively. All models were trained up to 10,000 steps, or until performance plateaued (defined as no improvement in validation metric after 1,000 iterations), and the peak validation set performance was recorded. We observed sufficient convergence within this budget for all models. We used a fixed learning rate of $1 \times 10^{-5}$ with AdamW optimizer, batch size of 8, gradient accumulation over 4 batches, and applied random rotations to the training set for all experiments. 

All models used identical CT preprocessing. CTs were cropped or padded to $256 \times 256 \times 128$, and we used multi-windowing to generate an 11-channel volume from each CT. For the baseline models, we prefixed a linear projection to transform the input to the expected number of channels for that model, implicitly allowing each model to choose its windowing strategy. We used the official PyTorch implementations for Swin3D-t and 3D ResNet-18. For MedicalNet and RadImageNet, we mapped the publicly available weights to the PyTorch ResNet implementations. For Merlin, we used an adapted version of the official implementation. We applied attention pooling to the feature maps to get an output vector for classification for all models except Merlin, which outputs a pooled feature vector directly. For RadImageNet, which is a 2D model, we applied the encoder to each slice, concatenated the per-slice feature maps, and then applied attention pooling.

\subsection{RAVE: Unified, Efficient Radiology Data Processing}
\label{sec:RaVE}
We introduce Radiology Vision Engine (RAVE), a framework that unifies data compression with standardized preprocessing to make training on large-scale medical volume datasets feasible under hardware constraints. Our datasets comprise millions of DICOM slices (tens of TBs) which are prohibitively large to store on local NVMe storage. RAVE converts DICOM series and NIfTI volumes into compact formats using High Efficiency Video Coding (HEVC), which exploits slice-to-slice redundancy to achieve high compression ratios while preserving detail. This compression enables us to keep the entire training corpus in limited NVMe capacity, improving training efficiency.

In addition to compression, RAVE provides a preprocessing framework supporting isotropic resampling, spatial normalization to fixed dimensions, and multi-windowing, yielding standardized, GPU-ready tensors. Full details can be found in our open-source code at \url{https://github.com/YalaLab/rave}.

\begin{table}[h]
\centering
\caption{Suite of released open-source tools}
\begin{tabular}{ll}
\toprule
 Tool & Link  \\
\midrule
Vision-language pretraining checkpoints for \name & \url{https://huggingface.co/collections/YalaLab/pillar-0} \\
Finetuning \name code & \url{https://github.com/YalaLab/pillar-finetune} \\
Vision-language pretraining code & \url{https://github.com/YalaLab/pillar-pretrain} \\
RATE & \url{https://github.com/YalaLab/rate} \\
RATE-Evals & \url{https://github.com/YalaLab/rate-evals} \\
RAVE & \url{https://github.com/YalaLab/rave}\\
\bottomrule
\end{tabular}
\label{tab:open_source_tools}
\end{table}

\newpage
\bibliographystyle{naturemag}

\bibliography{references}

\clearpage
\appendix
\section{Sybil vs Sybil 1.5}

\begin{table}[h]
\caption{\textbf{Sybil-1.5} performance on held-out NLST test set for demographic and behavioral subgroups. AUROC at 1, 3, 5 years and C-index values are followed by 95\% confidence intervals, which are capped at 100.}
\label{tab:group_auc}
\begin{tabular}{lcccc}
\toprule
Group & 1-yr AUC & 3-yr AUC & 5-yr AUC & C-index \\
\midrule
\textbf{Age} &&&& \\ 
\hspace{2mm} 50-60 & 96.0 (93.3, 99.6) & 82.0 (73.0, 92.3) & 77.1 (68.8, 86.1) & 77.0 (69.1, 85.8) \\
\hspace{2mm} 60-70 & 95.9 (93.3, 99.3) & 84.8 (80.0, 89.8) & 80.7 (76.1, 85.5) & 79.9 (75.5, 84.9) \\
\textbf{Sex} &&&& \\ 
\hspace{2mm} Male & 96.2 (94.2, 99.2) & 82.1 (76.3, 88.5) & 75.6 (69.8, 81.7) & 75.0 (69.4, 81.0) \\
\hspace{2mm} Female & 91.6 (86.3, 98.4) & 85.7 (80.4, 91.6) & 82.6 (77.4, 88.0) & 82.2 (77.3, 87.4) \\
\textbf{Race} &&&& \\ 
\hspace{2mm} White & 94.2 (91.6, 97.4) & 82.9 (78.5, 88.0) & 77.6 (73.0, 82.7) & 77.0 (72.6, 81.6) \\
\hspace{2mm} Black & 99.5 (98.9, 100.0) & 98.6 (97.1, 100.0) & 96.4 (92.8, 100.0) & 96.0 (91.9, 100.0) \\
\hspace{2mm} Asian & 97.8 (95.7, 100.0) & 79.4 (58.8, 100.0) & 73.3 (54.4, 96.8) & 71.9 (55.0, 94.3) \\
\textbf{Smoking status} &&&& \\ 
\hspace{2mm} Current smoker & 94.2 (90.5, 99.5) & 83.0 (77.2, 88.9) & 75.9 (70.1, 81.9) & 75.0 (69.5, 80.8) \\
\hspace{2mm} Not current smoker & 95.0 (92.0, 98.8) & 83.4 (77.0, 90.9) & 79.5 (72.9, 87.0) & 79.2 (72.6, 86.3) \\
\textbf{Smoking duration} &&&& \\ 
\hspace{2mm} $\leq$ 40 pack-years & 96.4 (93.9, 100.0) & 82.4 (74.5, 91.2) & 79.2 (71.1, 87.7) & 79.2 (71.4, 87.9) \\
\hspace{2mm} > 40 pack-years & 92.6 (88.6, 97.8) & 82.1 (77.1, 87.4) & 74.5 (69.1, 80.0) & 73.6 (68.3, 79.0) \\
\bottomrule
\end{tabular}
\end{table}

\begin{table}[htbp]
\centering
\caption{Sample efficiency results on the RSNA-2019 dataset, in terms of validation set AUROC across training data fractions.}
\begin{tabular}{lccccccc}
\toprule
\multirow{2}{*}{Model} & \multicolumn{7}{c}{Training data (\%)} \\
\cmidrule(lr){2-8}
 & 2.5 & 5 & 10 & 25 & 50 & 75 & 100 \\
\midrule
\textbf{Pillar-0}   & \textbf{95.0} & \textbf{95.5} & \textbf{95.9} & \textbf{96.5} & \textbf{96.8} & \textbf{96.7} & \textbf{96.7} \\
Swin3D-t            & 85.6 & 90.6 & 92.2 & 94.1 & 95.3 & 95.7 & 95.9 \\
Merlin              & 87.0 & 90.6 & 91.7 & 94.5 & 94.8 & 95.2 & 95.6 \\
3D ResNet-18        & 84.3 & 89.1 & 89.6 & 91.9 & 93.4 & 94.7 & 95.1 \\
RadImageNet         & 82.5 & 85.3 & 87.9 & 87.9 & 91.2 & 91.9 & 92.1 \\
MedicalNet          & 76.9 & 83.1 & 86.9 & 89.2 & 90.4 & 90.5 & 90.4 \\
\bottomrule
\end{tabular}
\label{tab:sample-eff}
\end{table}
\newpage

\section{Comparing \name to MedGemma}
\label{sec:medgemma-comparison}

\begin{figure}[h]
\centering
\captionsetup{singlelinecheck=false}
\includegraphics[width=0.6\linewidth]{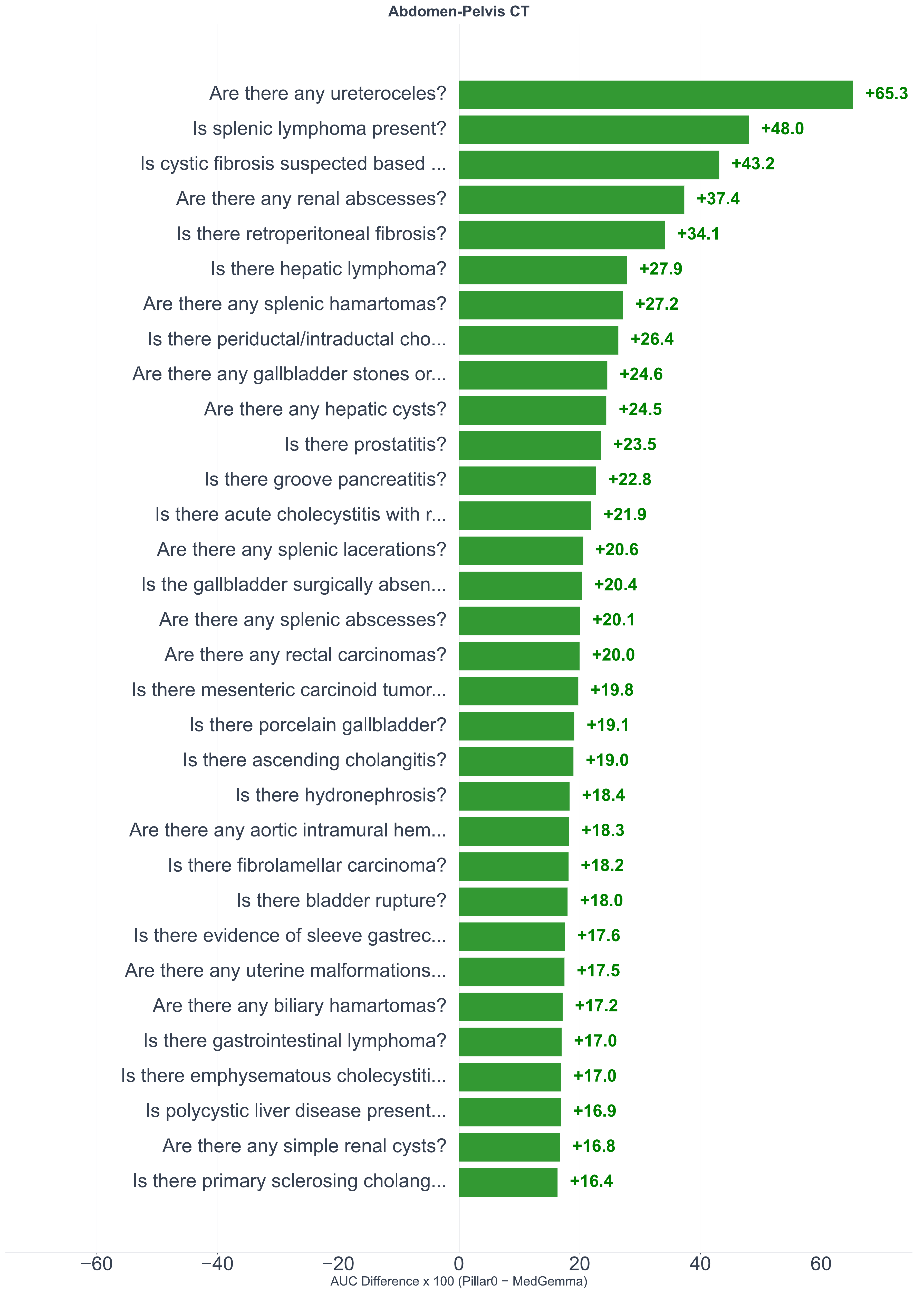}
\caption{\name vs MedGemma head-to-head on all UCSF Abdomen-Pelvis CT RATE-Evals tasks. \name wins on 190/210 (90.5\%, green bars); MedGemma wins on 20/210 (9.5\%, red bars). Part 1/7. }
\end{figure}

\newpage

\begin{figure}[t]
\centering
\captionsetup{singlelinecheck=false}
\includegraphics[width=0.7\linewidth]{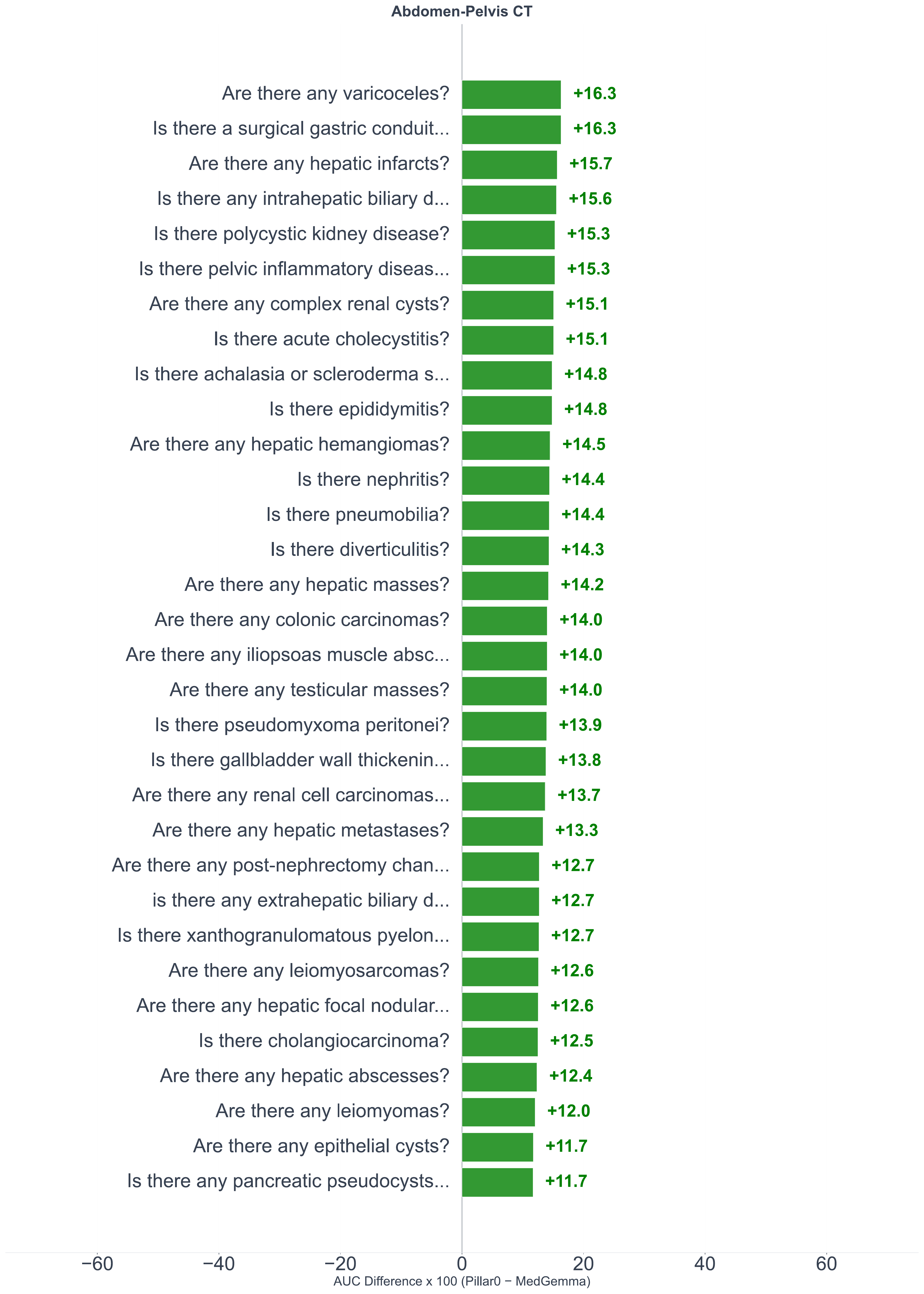}
\caption{\name vs MedGemma head-to-head on all UCSF Abdomen-Pelvis CT RATE-Evals tasks. \name wins on 190/210 (90.5\%, green bars); MedGemma wins on 20/210 (9.5\%, red bars). Part 2/7.}
\end{figure}

\begin{figure}[t]
\centering
\captionsetup{singlelinecheck=false}
\includegraphics[width=0.7\linewidth]{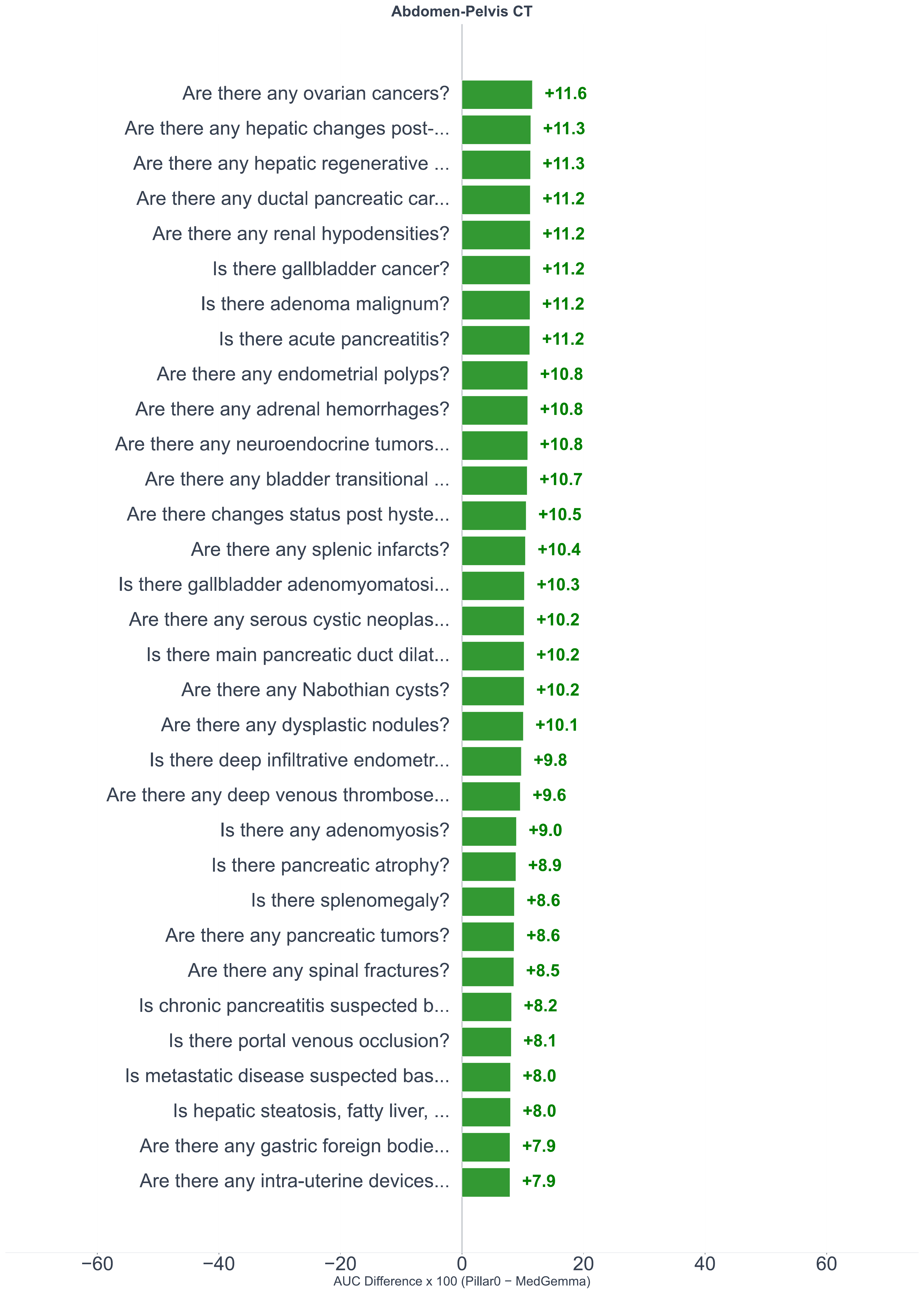}
\caption{\name vs MedGemma head-to-head on all UCSF Abdomen-Pelvis CT RATE-Evals tasks. \name wins on 190/210 (90.5\%, green bars); MedGemma wins on 20/210 (9.5\%, red bars). Part 3/7.}
\end{figure}

\begin{figure}[t]
\centering
\captionsetup{singlelinecheck=false}
\includegraphics[width=0.7\linewidth]{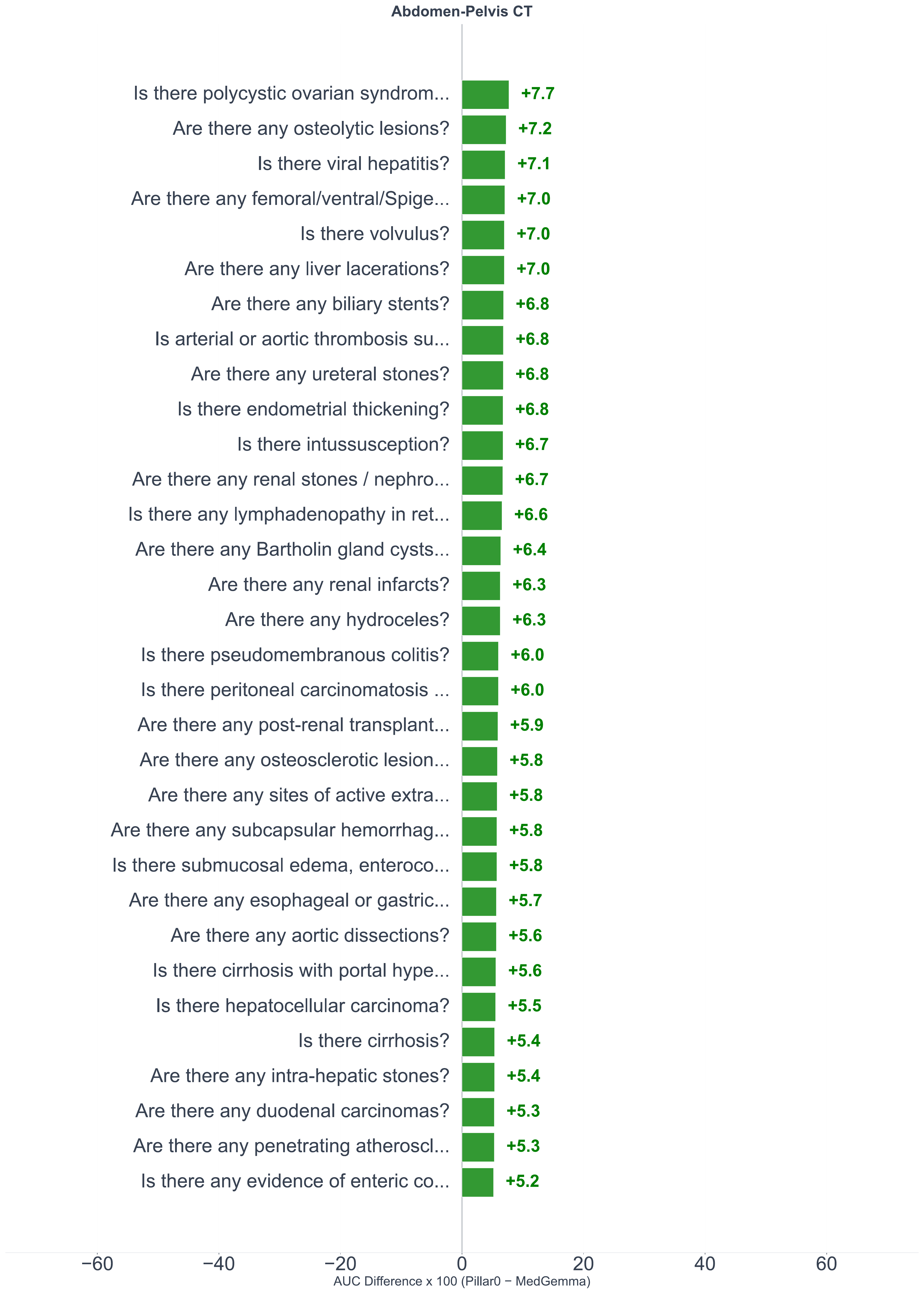}
\caption{\name vs MedGemma head-to-head on all UCSF Abdomen-Pelvis CT RATE-Evals tasks. \name wins on 190/210 (90.5\%, green bars); MedGemma wins on 20/210 (9.5\%, red bars). Part 4/7.}
\end{figure}

\begin{figure}[t]
\centering
\captionsetup{singlelinecheck=false}
\includegraphics[width=0.7\linewidth]{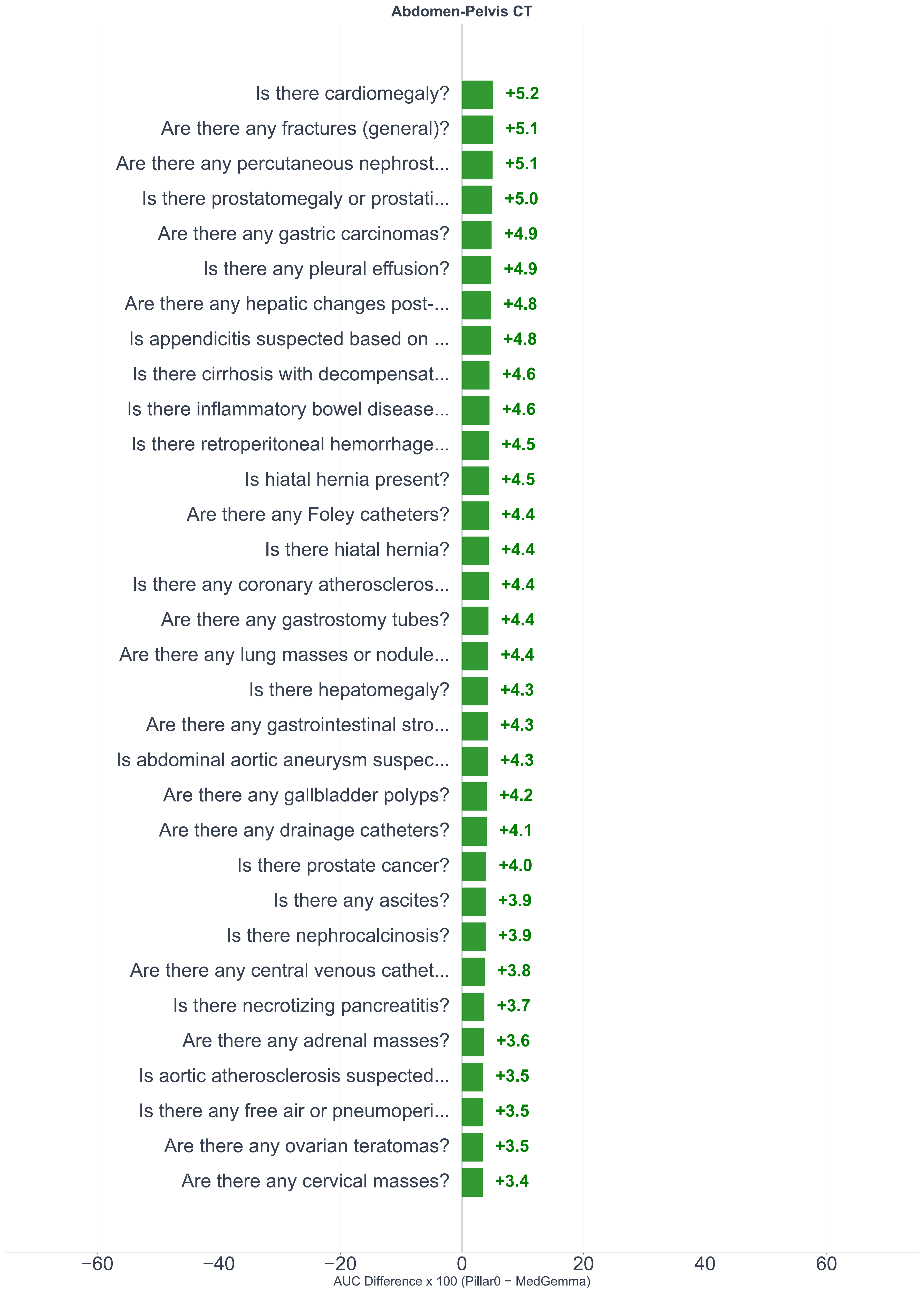}
\caption{\name vs MedGemma head-to-head on all UCSF Abdomen-Pelvis CT RATE-Evals tasks. \name wins on 190/210 (90.5\%, green bars); MedGemma wins on 20/210 (9.5\%, red bars). Part 5/7.}
\end{figure}

\begin{figure}[t]
\centering
\captionsetup{singlelinecheck=false}
\includegraphics[width=0.7\linewidth]{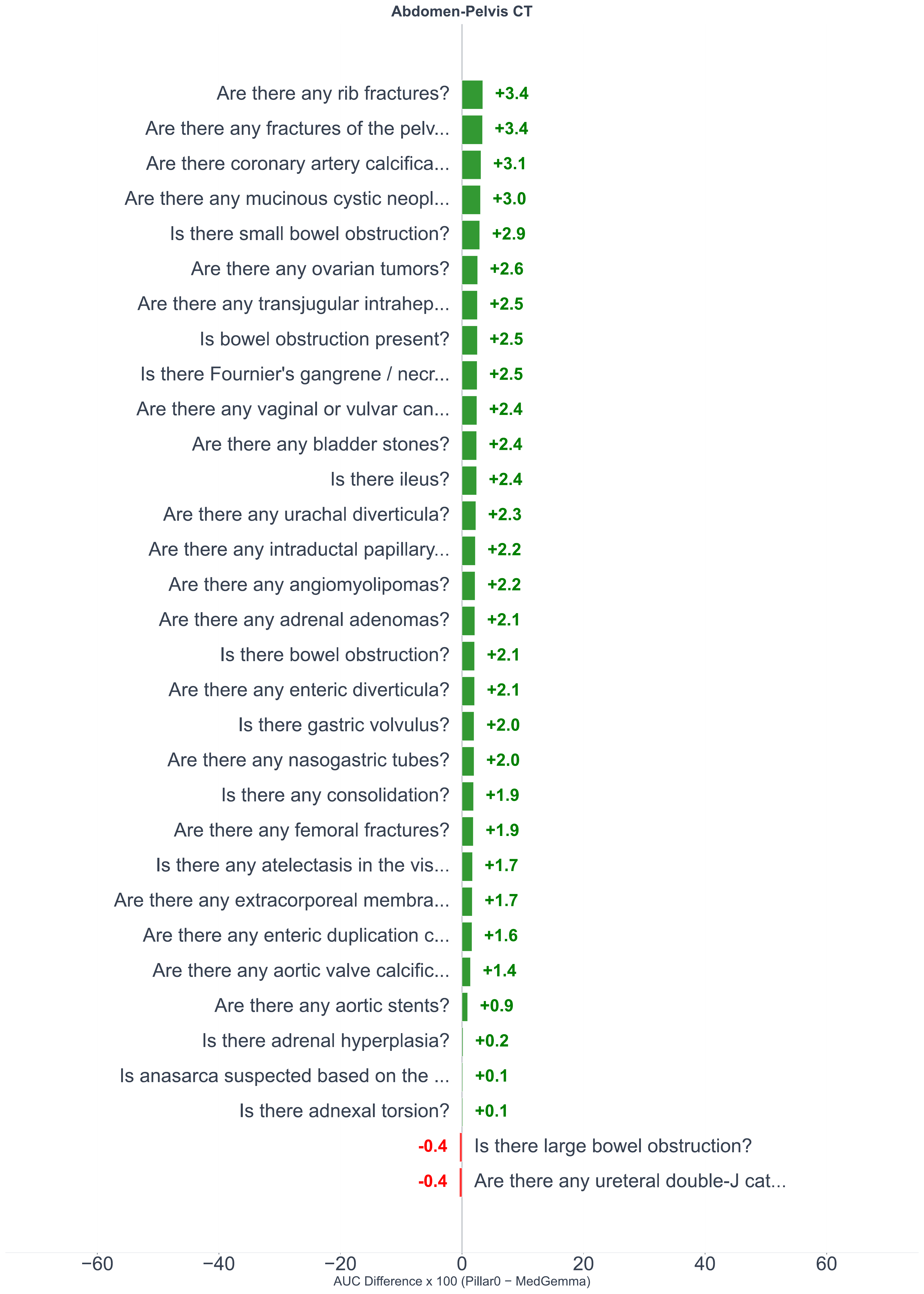}
\caption{\name vs MedGemma head-to-head on all UCSF Abdomen-Pelvis CT RATE-Evals tasks. \name wins on 190/210 (90.5\%, green bars); MedGemma wins on 20/210 (9.5\%, red bars). Part 6/7.}
\end{figure}

\begin{figure}[t]
\centering
\captionsetup{singlelinecheck=false}
\includegraphics[width=0.7\linewidth]{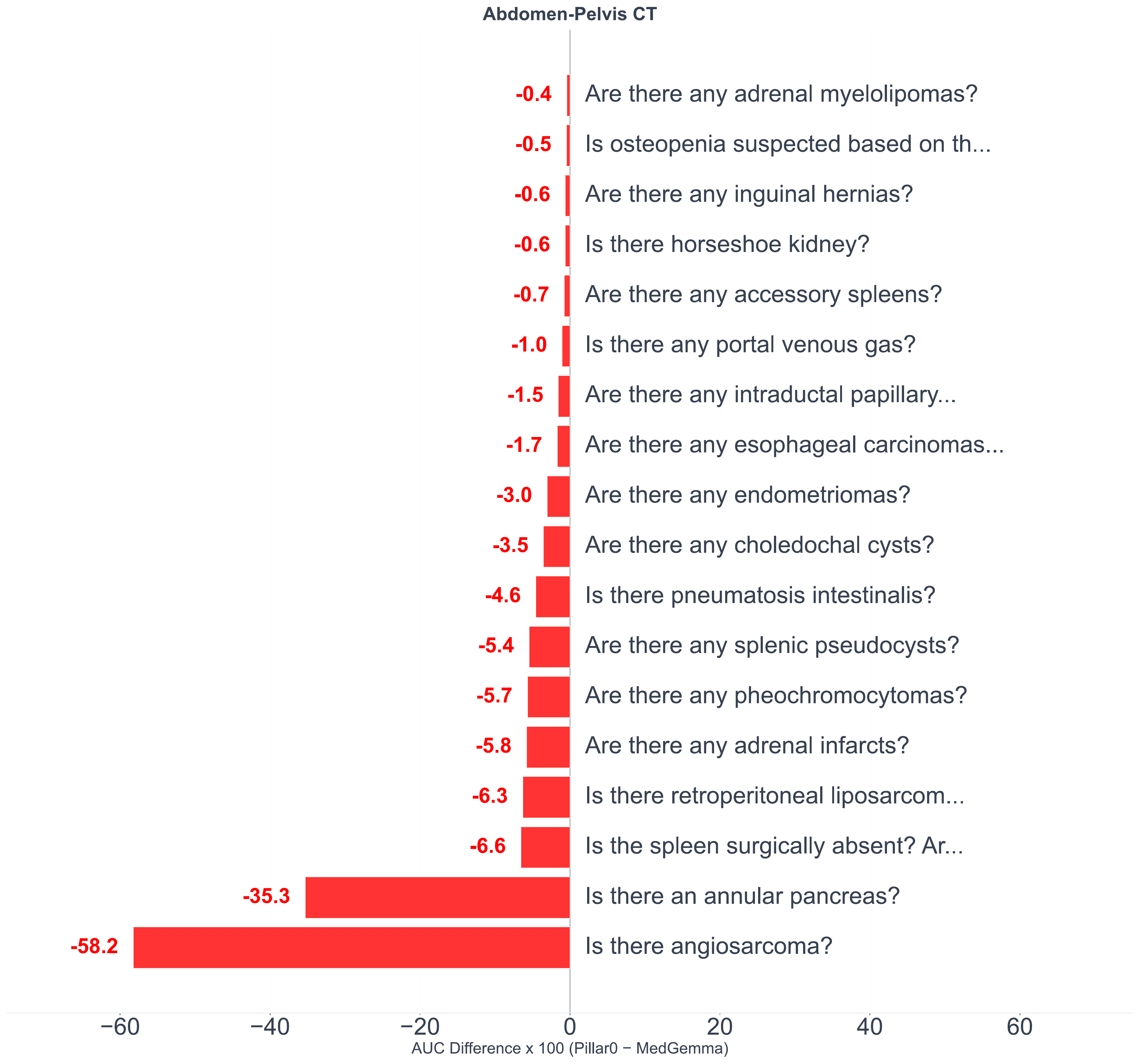}
\caption{\name vs MedGemma head-to-head on all UCSF Abdomen-Pelvis CT RATE-Evals tasks. \name wins on 190/210 (90.5\%, green bars); MedGemma wins on 20/210 (9.5\%, red bars). Part 7/7.}
\end{figure}

\newpage
\begin{figure}[t]
\centering
\captionsetup{singlelinecheck=false}
\includegraphics[width=0.7\linewidth]{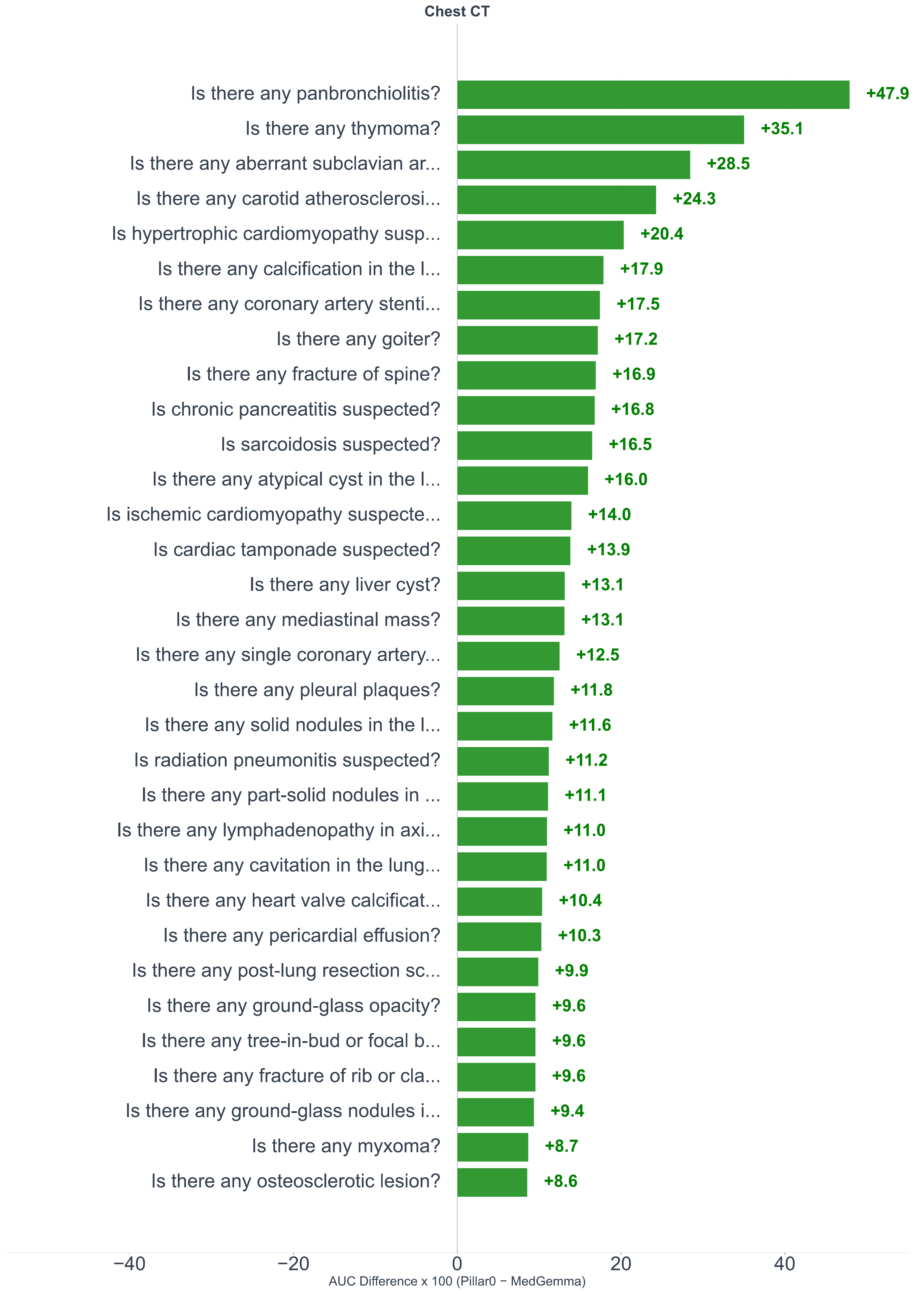}
\caption{\name vs MedGemma head-to-head on all UCSF Chest CT RATE-Evals tasks. \name wins on 85/92 (92.4\%, green bars); MedGemma wins on 7/92 (7.6\%, red bars). Part 1/3.}
\end{figure}

\begin{figure}[t]
\centering
\captionsetup{singlelinecheck=false}
\includegraphics[width=0.7\linewidth]{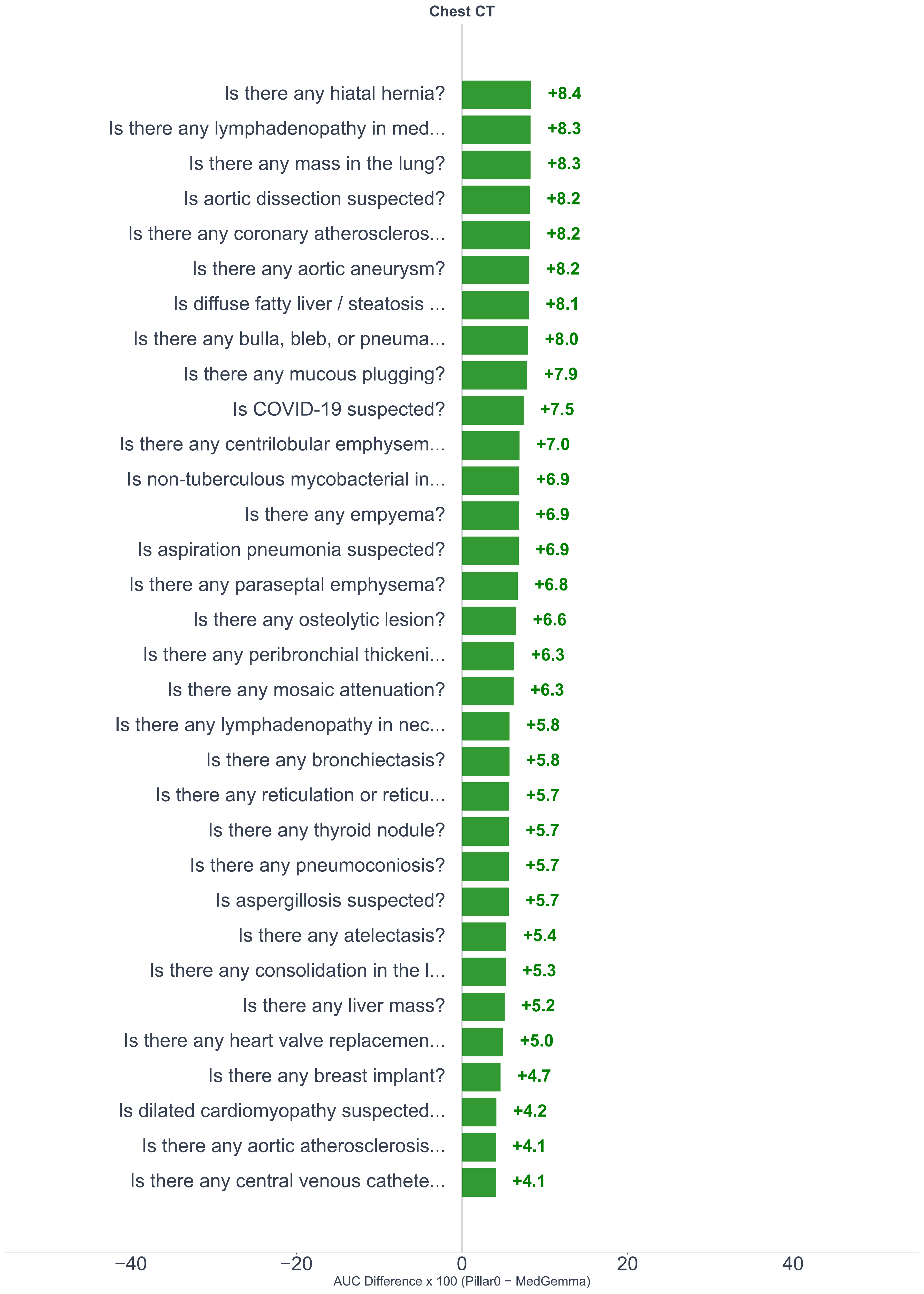}
\caption{\name vs MedGemma head-to-head on all UCSF Chest CT RATE-Evals tasks. \name wins on 85/92 (92.4\%, green bars); MedGemma wins on 7/92 (7.6\%, red bars). Part 2/3.}
\end{figure}

\begin{figure}[t]
\centering
\captionsetup{singlelinecheck=false}
\includegraphics[width=0.7\linewidth]{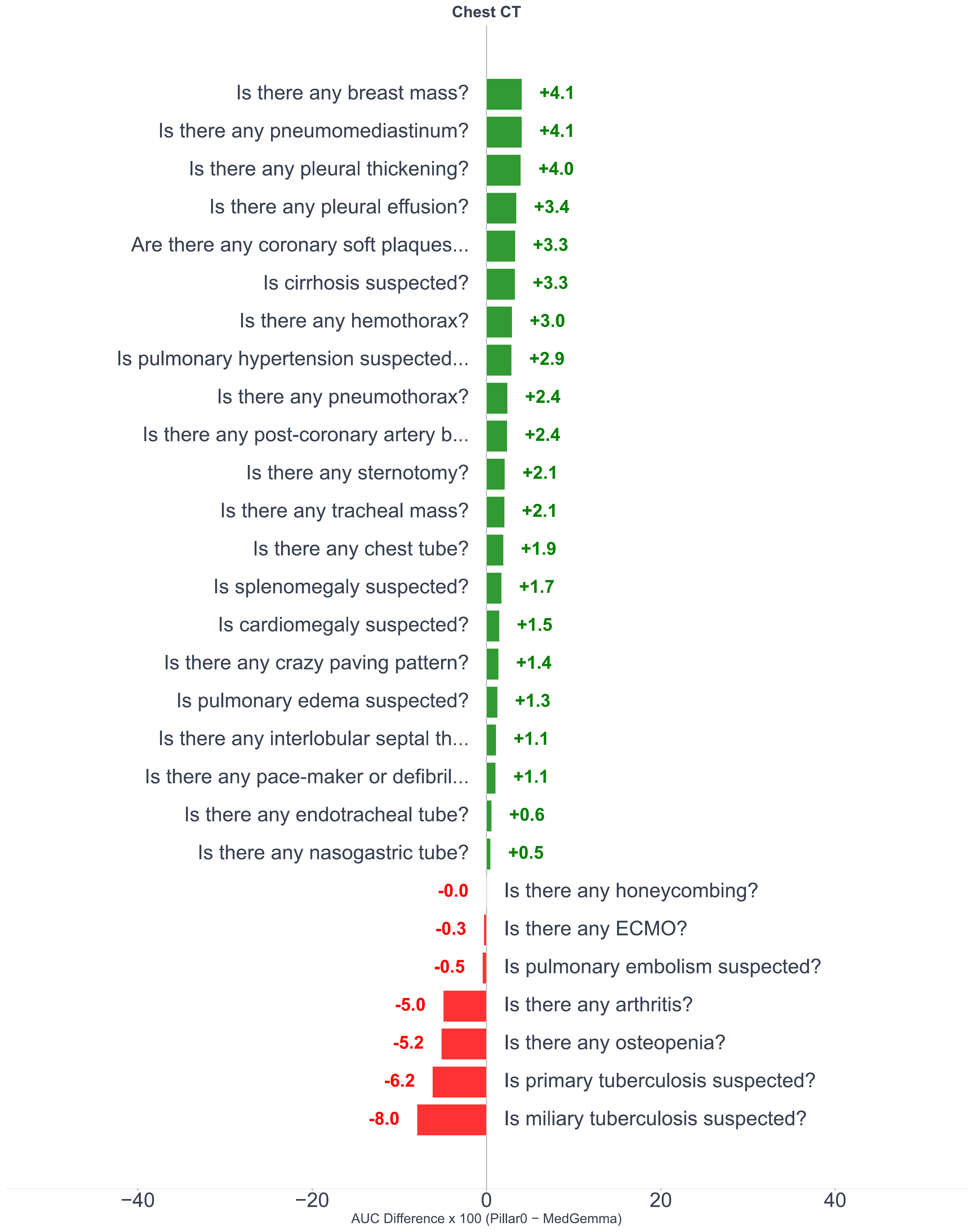}
\caption{\name vs MedGemma head-to-head on all UCSF Chest CT RATE-Evals tasks. \name wins on 85/92 (92.4\%, green bars); MedGemma wins on 7/92 (7.6\%, red bars). Part 3/3.}
\end{figure}

\newpage
\begin{figure}[t]
\centering
\captionsetup{singlelinecheck=false}
\includegraphics[width=0.7\linewidth]{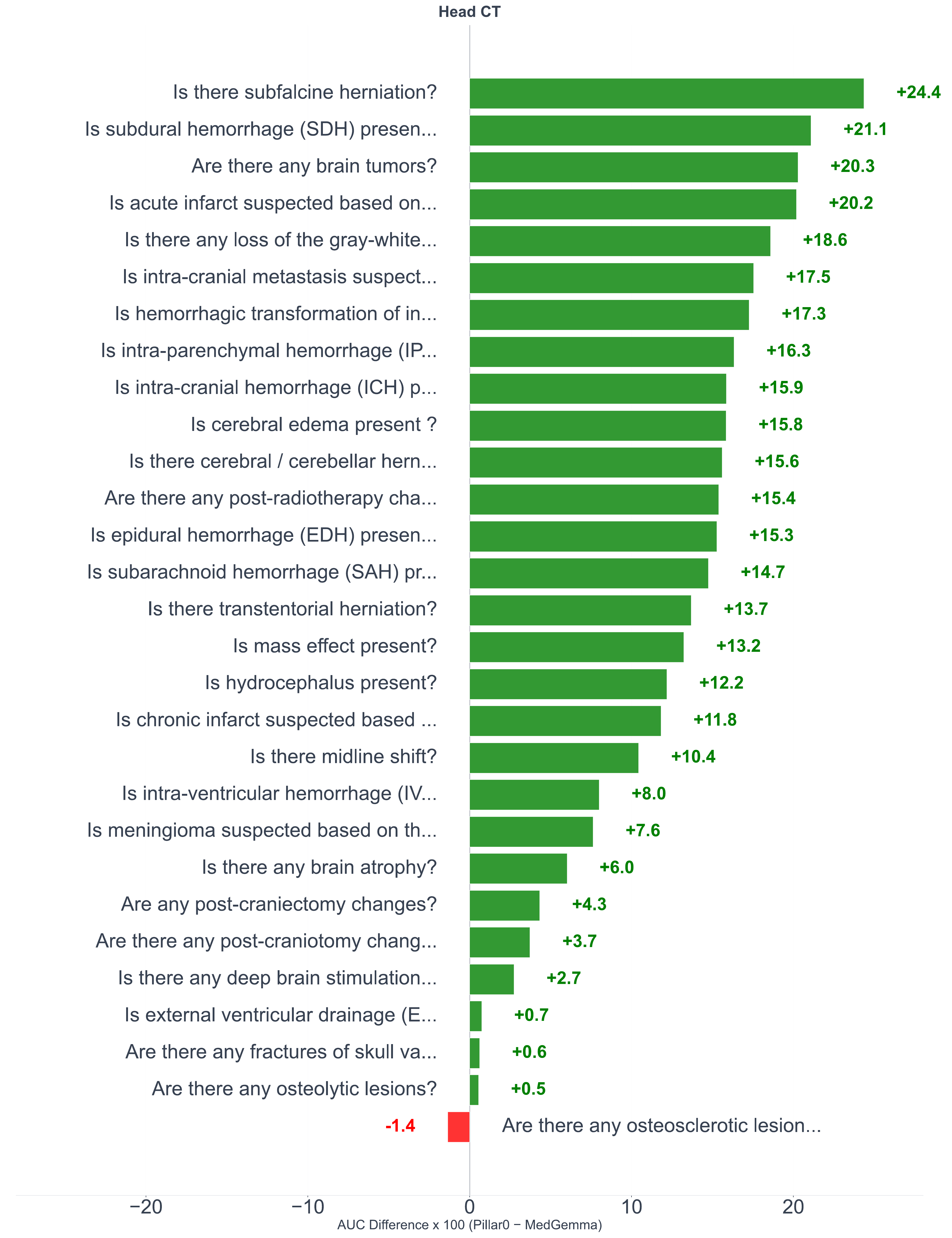}
\caption{\name vs MedGemma head-to-head on all UCSF Head CT RATE-Evals tasks. \name wins on 28/29 (96.6\%, green bars); MedGemma wins on 1/29 (3.4\%, red bars).}
\end{figure}

\newpage

\begin{figure}[t]
\centering
\captionsetup{singlelinecheck=false}
\includegraphics[width=0.7\linewidth]{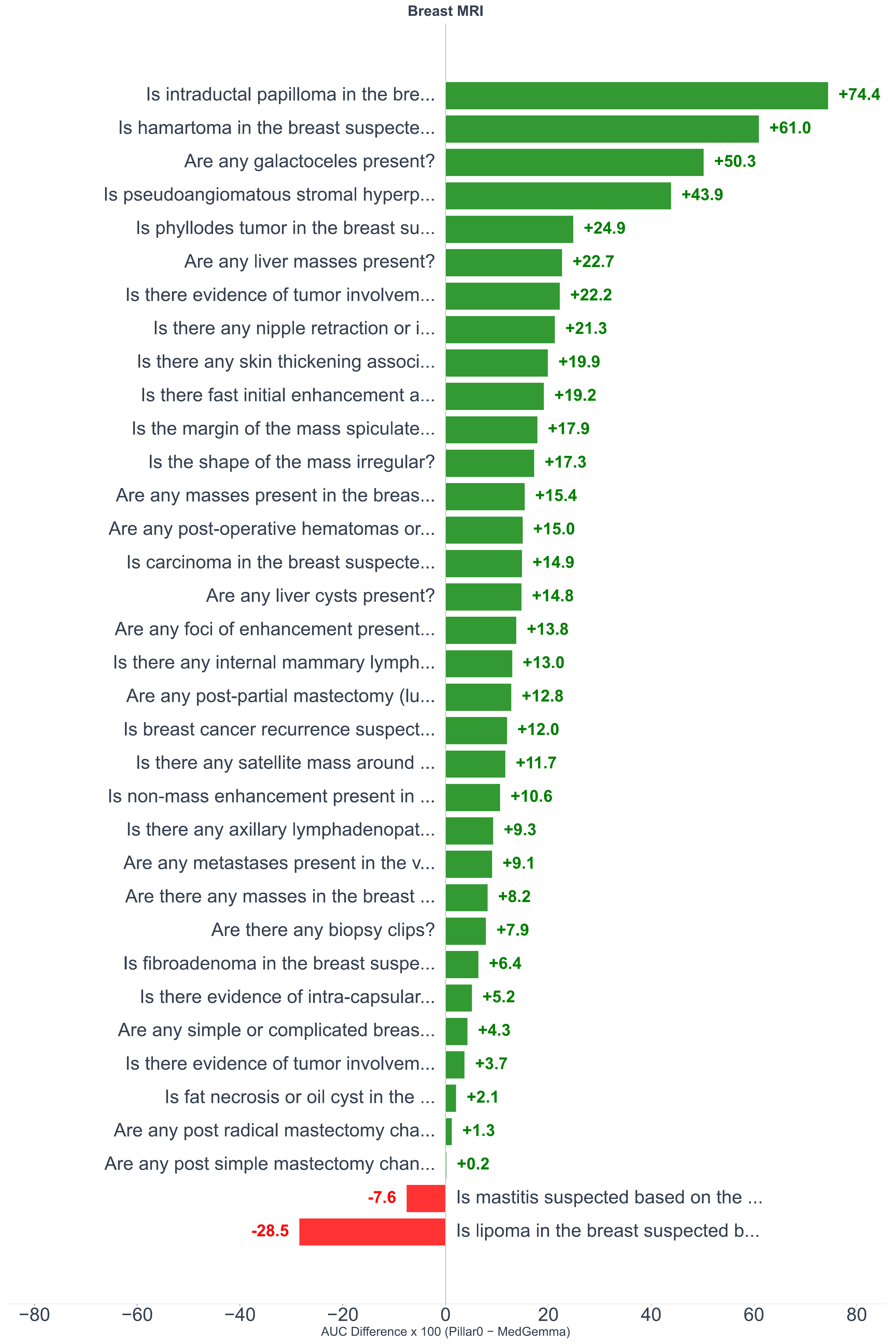}
\caption{\name vs MedGemma head-to-head on all UCSF Breast MRI RATE-Evals tasks. \name wins on 33/35 (94.3\%, green bars); MedGemma wins on 2/35 (5.7\%, red bars).}
\end{figure}

\newpage
\clearpage
\newpage
\section{Performance on full set of RATE-Evals tasks on UCSF Abdomen-Pelvis CT test set}
\label{ref:abd-finding}

\begin{center}
\footnotesize
\begin{longtable}{p{5cm} >{\centering\arraybackslash}p{2.0cm} >{\centering\arraybackslash}p{2.0cm} >{\centering\arraybackslash}p{2.0cm} >{\centering\arraybackslash}p{2.0cm} >{\centering\arraybackslash}p{2.0cm}} 
\caption{Model Performance (AUC x 100) for UCSF Abdomen-Pelvis CT} \label{tab:auc_ucsf_abd_ct} \\
\arrayrulecolor{black}
\toprule
& \multicolumn{5}{c}{\textbf{Model AUC (x 100)}} \\
\cmidrule(lr){2-6} 
\textbf{Finding} & \textbf{LingShu} & \textbf{MedGemma} & \textbf{MedImageInsight} & \textbf{Merlin} & \textbf{\name} \\
\arrayrulecolor{black}
\midrule
\endfirsthead

\multicolumn{6}{c}{%
\tablename\ \thetable{} -- Model Performance (AUC x 100) for UCSF Abd CT \textit{(continued)}} \\
\arrayrulecolor{black}
\toprule
& \multicolumn{5}{c}{\textbf{Model AUC (x 100)}} \\
\cmidrule(lr){2-6} 
\textbf{Finding} & \textbf{LingShu} & \textbf{MedGemma} & \textbf{MedImageInsight} & \textbf{Merlin} & \textbf{\name} \\
\arrayrulecolor{black}
\midrule
\endhead

\arrayrulecolor{black}
\midrule
\multicolumn{6}{r}{\textit{Continued on next page}} \\
\endfoot

\arrayrulecolor{black}
\bottomrule
\endlastfoot
\rowcolor{rowgray}
\textbf{Biliary findings} & & & & & \\
\rowcolor{rowgray}
\quad Are there any biliary cystadenomas? & NA & NA & NA & NA & NA \\
\arrayrulecolor{linegray}\hhline{------}
\rowcolor{rowgray}
\quad Are there any biliary hamartomas? & 71.2 & 57.2 & 60.4 & 52.7 & \textbf{74.4} \\
\arrayrulecolor{linegray}\hhline{------}
\rowcolor{rowgray}
\quad Are there any biliary stents? & 83.1 & 92.2 & 85.8 & 92.3 & \textbf{99.0} \\
\arrayrulecolor{linegray}\hhline{------}
\rowcolor{rowgray}
\quad Are there any choledochal cysts? & \textbf{29.8} & 29.2 & 9.2 & 22.6 & 25.6 \\
\arrayrulecolor{linegray}\hhline{------}
\rowcolor{rowgray}
\quad Are there any gallbladder polyps? & 67.2 & 68.0 & 66.9 & 62.9 & \textbf{72.1} \\
\arrayrulecolor{linegray}\hhline{------}
\rowcolor{rowgray}
\quad Are there any gallbladder stones or cholelithiasis? & 64.2 & 66.1 & 68.3 & 66.0 & \textbf{90.7} \\
\arrayrulecolor{linegray}\hhline{------}
\rowcolor{rowgray}
\quad Is the gallbladder surgically absent? & 71.7 & 77.0 & 85.0 & 78.4 & \textbf{97.4} \\
\arrayrulecolor{linegray}\hhline{------}
\rowcolor{rowgray}
\quad Is there gallbladder adenomyomatosis? & 68.8 & 68.1 & 69.9 & 64.9 & \textbf{78.4} \\
\arrayrulecolor{linegray}\hhline{------}
\rowcolor{rowgray}
\quad Is there gallbladder cancer? & 77.1 & 85.6 & 82.2 & 79.5 & \textbf{96.8} \\
\arrayrulecolor{linegray}\hhline{------}
\rowcolor{rowgray}
\quad Is there gallbladder wall thickening? & 69.9 & 73.6 & 75.5 & 72.9 & \textbf{87.4} \\
\arrayrulecolor{linegray}\hhline{------}
\rowcolor{rowgray}
\quad Is there porcelain gallbladder? & 65.3 & 67.1 & 54.2 & 61.7 & \textbf{86.3} \\
\arrayrulecolor{black}\midrule

\textbf{Bone/Fracture findings} & & & & & \\

\quad Are there any femoral fractures? & 84.8 & 86.7 & 84.0 & 85.3 & \textbf{88.6} \\
\arrayrulecolor{linegray}\hhline{------}

\quad Are there any fractures (general)? & 73.0 & 74.0 & 72.1 & 71.5 & \textbf{79.2} \\
\arrayrulecolor{linegray}\hhline{------}

\quad Are there any osteolytic lesions? & 71.5 & 76.0 & 72.4 & 70.6 & \textbf{83.3} \\
\arrayrulecolor{linegray}\hhline{------}

\quad Are there any rib fractures? & 72.9 & 74.4 & 73.4 & 72.6 & \textbf{77.8} \\
\arrayrulecolor{linegray}\hhline{------}

\quad Are there any spinal fractures? & 72.6 & 74.6 & 72.1 & 70.6 & \textbf{83.1} \\
\arrayrulecolor{black}\midrule
\rowcolor{rowgray}
\textbf{Cardiac findings} & & & & & \\
\rowcolor{rowgray}
\quad Is there any coronary atherosclerosis? & 83.5 & 86.0 & 84.8 & 84.3 & \textbf{90.4} \\
\arrayrulecolor{linegray}\hhline{------}
\rowcolor{rowgray}
\quad Is there cardiomegaly? & 81.7 & 86.6 & 83.7 & 85.4 & \textbf{91.8} \\
\arrayrulecolor{black}\midrule

\textbf{Chest/Lung findings} & & & & & \\

\quad Are there any lung masses or nodules? & 68.7 & 72.0 & 71.2 & 69.6 & \textbf{76.4} \\
\arrayrulecolor{linegray}\hhline{------}

\quad Is there any atelectasis in the visible lung? & 78.7 & 80.8 & 80.2 & 77.9 & \textbf{82.5} \\
\arrayrulecolor{linegray}\hhline{------}

\quad Is there any free air or pneumoperitoneum? & 85.7 & 89.4 & 90.7 & 87.2 & \textbf{92.9} \\
\arrayrulecolor{linegray}\hhline{------}

\quad Is there pneumobilia? & 76.8 & 84.3 & 83.0 & 84.6 & \textbf{98.7} \\
\arrayrulecolor{black}\midrule
\rowcolor{rowgray}
\textbf{Cyst findings} & & & & & \\
\rowcolor{rowgray}
\quad Are there any Bartholin gland cysts? & 65.1 & 77.3 & 77.0 & 70.6 & \textbf{83.6} \\
\arrayrulecolor{linegray}\hhline{------}
\rowcolor{rowgray}
\quad Are there any Gartner duct cysts? & NA & NA & NA & NA & NA \\
\arrayrulecolor{linegray}\hhline{------}
\rowcolor{rowgray}
\quad Are there any Nabothian cysts? & 74.7 & 71.6 & 70.5 & 69.7 & \textbf{81.8} \\
\arrayrulecolor{linegray}\hhline{------}
\rowcolor{rowgray}
\quad Are there any enteric duplication cysts? & \textbf{76.6} & 49.1 & 59.4 & 52.3 & 50.7 \\
\arrayrulecolor{linegray}\hhline{------}
\rowcolor{rowgray}
\quad Are there any epithelial cysts? & 60.8 & 63.5 & 63.1 & 60.0 & \textbf{75.3} \\
\arrayrulecolor{linegray}\hhline{------}
\rowcolor{rowgray}
\quad Are there any mucinous cystic neoplasms? & 71.3 & 71.9 & 72.8 & 64.2 & \textbf{75.0} \\
\arrayrulecolor{linegray}\hhline{------}
\rowcolor{rowgray}
\quad Are there any serous cystic neoplasms? & 74.8 & 77.9 & 66.7 & 57.6 & \textbf{88.1} \\
\arrayrulecolor{linegray}\hhline{------}
\rowcolor{rowgray}
\quad Is there acute cholecystitis with rupture? & 69.2 & 71.2 & 82.6 & 56.9 & \textbf{93.1} \\
\arrayrulecolor{linegray}\hhline{------}
\rowcolor{rowgray}
\quad Is there acute cholecystitis? & 77.0 & 78.2 & 86.0 & 74.9 & \textbf{93.3} \\
\arrayrulecolor{linegray}\hhline{------}
\rowcolor{rowgray}
\quad Is there emphysematous cholecystitis? & 70.0 & 80.8 & 75.6 & 84.4 & \textbf{97.8} \\
\arrayrulecolor{black}\midrule

\textbf{Device/Procedural findings} & & & & & \\

\quad Are there any Foley catheters? & 90.2 & 92.3 & 91.9 & 91.4 & \textbf{96.7} \\
\arrayrulecolor{linegray}\hhline{------}

\quad Are there any central venous catheters? & 84.7 & 86.6 & 84.3 & 84.3 & \textbf{90.4} \\
\arrayrulecolor{linegray}\hhline{------}

\quad Are there any drainage catheters? & 86.4 & 90.2 & 88.1 & 87.0 & \textbf{94.3} \\
\arrayrulecolor{linegray}\hhline{------}

\quad Are there any extracorporeal membrane oxygenation (ECMO) devices? & 96.5 & 97.6 & 98.2 & \textbf{99.4} & 99.3 \\
\arrayrulecolor{linegray}\hhline{------}

\quad Are there any gastrostomy tubes? & 89.2 & 92.2 & 90.5 & 88.0 & \textbf{96.6} \\
\arrayrulecolor{linegray}\hhline{------}

\quad Are there any post-nephrectomy changes? & 72.9 & 82.7 & 79.5 & 73.2 & \textbf{95.4} \\
\arrayrulecolor{black}\midrule
\rowcolor{rowgray}
\textbf{Gastrointestinal findings} & & & & & \\
\rowcolor{rowgray}
\quad Are there any colonic carcinomas? & 65.8 & 73.8 & 77.2 & 71.4 & \textbf{87.9} \\
\arrayrulecolor{linegray}\hhline{------}
\rowcolor{rowgray}
\quad Are there any duodenal carcinomas? & 63.0 & 91.9 & 88.9 & 83.3 & \textbf{97.2} \\
\arrayrulecolor{linegray}\hhline{------}
\rowcolor{rowgray}
\quad Are there any esophageal carcinomas? & 78.4 & 89.9 & 76.9 & \textbf{91.5} & 88.2 \\
\arrayrulecolor{linegray}\hhline{------}
\rowcolor{rowgray}
\quad Are there any esophageal or gastric varices? & 84.3 & 88.9 & 87.9 & 90.6 & \textbf{94.6} \\
\arrayrulecolor{linegray}\hhline{------}
\rowcolor{rowgray}
\quad Are there any gastric carcinomas? & 74.9 & 79.8 & 78.0 & 76.5 & \textbf{84.7} \\
\arrayrulecolor{linegray}\hhline{------}
\rowcolor{rowgray}
\quad Are there any gastric foreign bodies (not related to medical devices or post-surgical changes)? & 52.3 & 48.9 & 42.5 & 47.9 & \textbf{56.8} \\
\arrayrulecolor{linegray}\hhline{------}
\rowcolor{rowgray}
\quad Are there any gastrointestinal stromal tumors (GISTs)? & 68.5 & 63.9 & \textbf{75.0} & 64.9 & 68.2 \\
\arrayrulecolor{linegray}\hhline{------}
\rowcolor{rowgray}
\quad Are there any nasogastric tubes? & 92.3 & 94.4 & 91.2 & 93.2 & \textbf{96.3} \\
\arrayrulecolor{linegray}\hhline{------}
\rowcolor{rowgray}
\quad Is bowel obstruction present? & 88.0 & 92.5 & 93.2 & 90.2 & \textbf{95.0} \\
\arrayrulecolor{linegray}\hhline{------}
\rowcolor{rowgray}
\quad Is there a surgical gastric conduit? & 71.2 & 77.3 & 78.7 & 71.5 & \textbf{93.6} \\
\arrayrulecolor{linegray}\hhline{------}
\rowcolor{rowgray}
\quad Is there achalasia or scleroderma suspected based on gastrointestinal imaging findings? & 82.4 & 77.4 & 83.2 & 80.6 & \textbf{92.2} \\
\arrayrulecolor{linegray}\hhline{------}
\rowcolor{rowgray}
\quad Is there bowel obstruction? & 88.4 & 92.9 & 93.1 & 90.5 & \textbf{94.9} \\
\arrayrulecolor{linegray}\hhline{------}
\rowcolor{rowgray}
\quad Is there gastric volvulus? & 93.4 & 97.3 & 85.8 & 80.9 & \textbf{99.3} \\
\arrayrulecolor{linegray}\hhline{------}
\rowcolor{rowgray}
\quad Is there gastrointestinal lymphoma? & 81.9 & 70.5 & 79.6 & 74.4 & \textbf{87.6} \\
\arrayrulecolor{linegray}\hhline{------}
\rowcolor{rowgray}
\quad Is there inflammatory bowel disease? & 83.9 & 87.0 & 87.4 & 82.7 & \textbf{91.6} \\
\arrayrulecolor{linegray}\hhline{------}
\rowcolor{rowgray}
\quad Is there large bowel obstruction? & 89.4 & 93.7 & \textbf{94.7} & 89.4 & 93.3 \\
\arrayrulecolor{linegray}\hhline{------}
\rowcolor{rowgray}
\quad Is there pneumatosis intestinalis? & 87.7 & \textbf{91.4} & 89.0 & 78.2 & 86.9 \\
\arrayrulecolor{linegray}\hhline{------}
\rowcolor{rowgray}
\quad Is there small bowel obstruction? & 87.8 & 92.7 & 93.4 & 91.0 & \textbf{95.6} \\
\arrayrulecolor{black}\midrule

\textbf{Kidney-related findings} & & & & & \\

\quad Are there any adrenal adenomas? & 68.4 & 68.5 & \textbf{70.9} & 68.9 & 70.6 \\
\arrayrulecolor{linegray}\hhline{------}

\quad Are there any adrenal hemorrhages? & 60.7 & 49.4 & 64.3 & \textbf{66.3} & 60.2 \\
\arrayrulecolor{linegray}\hhline{------}

\quad Are there any adrenal infarcts? & 75.3 & \textbf{86.6} & 85.8 & 57.8 & 80.8 \\
\arrayrulecolor{linegray}\hhline{------}

\quad Are there any adrenal masses? & 63.2 & 64.8 & 64.8 & 64.0 & \textbf{68.5} \\
\arrayrulecolor{linegray}\hhline{------}

\quad Are there any adrenal myelolipomas? & 71.4 & 74.7 & \textbf{77.9} & 70.2 & 74.3 \\
\arrayrulecolor{linegray}\hhline{------}

\quad Are there any complex renal cysts? & 64.6 & 66.9 & 66.5 & 63.5 & \textbf{81.9} \\
\arrayrulecolor{linegray}\hhline{------}

\quad Are there any percutaneous nephrostomy tubes/catheters? & 88.7 & 93.1 & 87.0 & 85.5 & \textbf{98.2} \\
\arrayrulecolor{linegray}\hhline{------}

\quad Are there any post-renal transplantation changes? & 87.0 & 93.5 & 92.2 & 88.6 & \textbf{99.4} \\
\arrayrulecolor{linegray}\hhline{------}

\quad Are there any renal abscesses? & 61.4 & 52.0 & 53.8 & 58.1 & \textbf{89.4} \\
\arrayrulecolor{linegray}\hhline{------}

\quad Are there any renal cell carcinomas? & 70.9 & 77.3 & 76.9 & 71.6 & \textbf{91.0} \\
\arrayrulecolor{linegray}\hhline{------}

\quad Are there any renal hypodensities? & 62.9 & 64.7 & 64.4 & 62.7 & \textbf{76.0} \\
\arrayrulecolor{linegray}\hhline{------}

\quad Are there any renal infarcts? & 76.6 & 83.0 & 77.4 & 75.2 & \textbf{89.3} \\
\arrayrulecolor{linegray}\hhline{------}

\quad Are there any renal lacerations? & NA & NA & NA & NA & NA \\
\arrayrulecolor{linegray}\hhline{------}

\quad Are there any renal stones / nephrolithiasis? & 67.6 & 70.0 & 70.3 & 67.1 & \textbf{76.7} \\
\arrayrulecolor{linegray}\hhline{------}

\quad Are there any simple renal cysts? & 66.7 & 67.9 & 68.1 & 66.0 & \textbf{84.7} \\
\arrayrulecolor{linegray}\hhline{------}

\quad Are there any ureteral double-J catheters/stents? & 91.0 & \textbf{98.4} & 84.3 & 93.7 & 97.9 \\
\arrayrulecolor{linegray}\hhline{------}

\quad Are there any ureteral stones? & 85.0 & 87.4 & 86.6 & 82.5 & \textbf{94.2} \\
\arrayrulecolor{linegray}\hhline{------}

\quad Is there adrenal hyperplasia? & 57.6 & 59.3 & \textbf{62.2} & 48.9 & 59.4 \\
\arrayrulecolor{linegray}\hhline{------}

\quad Is there horseshoe kidney? & 67.7 & 73.9 & \textbf{78.6} & 68.1 & 73.3 \\
\arrayrulecolor{linegray}\hhline{------}

\quad Is there hydronephrosis? & 70.6 & 76.4 & 76.1 & 74.1 & \textbf{94.7} \\
\arrayrulecolor{linegray}\hhline{------}

\quad Is there nephrocalcinosis? & 75.9 & 77.0 & 77.2 & 59.1 & \textbf{80.9} \\
\arrayrulecolor{linegray}\hhline{------}

\quad Is there polycystic kidney disease? & 62.5 & 74.2 & 75.0 & 70.2 & \textbf{89.4} \\
\arrayrulecolor{linegray}\hhline{------}

\quad Is there renal lymphomas? & NA & NA & NA & NA & NA \\
\arrayrulecolor{black}\midrule
\rowcolor{rowgray}
\textbf{Liver-related findings} & & & & & \\
\rowcolor{rowgray}
\quad Are there any hepatic abscesses? & 77.1 & 81.7 & 77.0 & 77.1 & \textbf{94.1} \\
\arrayrulecolor{linegray}\hhline{------}
\rowcolor{rowgray}
\quad Are there any hepatic adenomas? & NA & NA & NA & NA & NA \\
\arrayrulecolor{linegray}\hhline{------}
\rowcolor{rowgray}
\quad Are there any hepatic changes post-resection? & 70.0 & 77.9 & 78.2 & 78.4 & \textbf{89.2} \\
\arrayrulecolor{linegray}\hhline{------}
\rowcolor{rowgray}
\quad Are there any hepatic changes post-transplantation? & 83.1 & 91.8 & 84.0 & 91.6 & \textbf{96.6} \\
\arrayrulecolor{linegray}\hhline{------}
\rowcolor{rowgray}
\quad Are there any hepatic cysts? & 63.0 & 65.6 & 65.7 & 64.1 & \textbf{90.0} \\
\arrayrulecolor{linegray}\hhline{------}
\rowcolor{rowgray}
\quad Are there any hepatic focal nodular hyperplasias? & 78.0 & 79.6 & 75.1 & 79.2 & \textbf{92.1} \\
\arrayrulecolor{linegray}\hhline{------}
\rowcolor{rowgray}
\quad Are there any hepatic hemangiomas? & 62.3 & 64.6 & 65.7 & 61.9 & \textbf{79.1} \\
\arrayrulecolor{linegray}\hhline{------}
\rowcolor{rowgray}
\quad Are there any hepatic infarcts? & 71.8 & 77.0 & 73.4 & 78.5 & \textbf{92.6} \\
\arrayrulecolor{linegray}\hhline{------}
\rowcolor{rowgray}
\quad Are there any hepatic masses? & 70.0 & 74.4 & 74.5 & 72.7 & \textbf{88.6} \\
\arrayrulecolor{linegray}\hhline{------}
\rowcolor{rowgray}
\quad Are there any hepatic metastases? & 74.4 & 81.2 & 81.1 & 79.9 & \textbf{94.5} \\
\arrayrulecolor{linegray}\hhline{------}
\rowcolor{rowgray}
\quad Are there any hepatic regenerative nodules? & 70.6 & 64.3 & 60.9 & 64.1 & \textbf{75.5} \\
\arrayrulecolor{linegray}\hhline{------}
\rowcolor{rowgray}
\quad Are there any intra-hepatic stones? & 73.6 & 93.5 & 96.2 & 85.1 & \textbf{98.9} \\
\arrayrulecolor{linegray}\hhline{------}
\rowcolor{rowgray}
\quad Are there any liver lacerations? & 89.7 & 84.8 & 88.4 & 71.8 & \textbf{91.8} \\
\arrayrulecolor{linegray}\hhline{------}
\rowcolor{rowgray}
\quad Are there any sites of active extravasation in the liver? & 76.2 & 88.1 & 71.2 & 71.8 & \textbf{93.9} \\
\arrayrulecolor{linegray}\hhline{------}
\rowcolor{rowgray}
\quad Are there any transjugular intrahepatic portosystemic shunts? & 88.7 & 96.2 & 87.6 & 93.2 & \textbf{98.7} \\
\arrayrulecolor{linegray}\hhline{------}
\rowcolor{rowgray}
\quad Is hepatic steatosis, fatty liver, focal fatty infiltration suspected based on the imaging findings? & 72.7 & 75.6 & 75.2 & 75.3 & \textbf{83.6} \\
\arrayrulecolor{linegray}\hhline{------}
\rowcolor{rowgray}
\quad Is polycystic liver disease present? & 61.4 & 77.1 & 74.5 & 77.6 & \textbf{94.0} \\
\arrayrulecolor{linegray}\hhline{------}
\rowcolor{rowgray}
\quad Is there any intrahepatic biliary ductal dilation? & 72.7 & 76.7 & 76.6 & 78.6 & \textbf{92.3} \\
\arrayrulecolor{linegray}\hhline{------}
\rowcolor{rowgray}
\quad Is there any portal venous gas? & 82.6 & 85.2 & 81.9 & \textbf{88.0} & 84.2 \\
\arrayrulecolor{linegray}\hhline{------}
\rowcolor{rowgray}
\quad Is there cirrhosis with decompensation? & 91.1 & 92.9 & 93.5 & 93.8 & \textbf{97.5} \\
\arrayrulecolor{linegray}\hhline{------}
\rowcolor{rowgray}
\quad Is there cirrhosis with portal hypertension? & 88.1 & 91.8 & 91.7 & 93.2 & \textbf{97.4} \\
\arrayrulecolor{linegray}\hhline{------}
\rowcolor{rowgray}
\quad Is there cirrhosis? & 86.9 & 90.6 & 90.3 & 90.9 & \textbf{96.0} \\
\arrayrulecolor{linegray}\hhline{------}
\rowcolor{rowgray}
\quad Is there hepatic lymphoma? & 68.9 & 55.2 & 75.3 & 71.2 & \textbf{83.1} \\
\arrayrulecolor{linegray}\hhline{------}
\rowcolor{rowgray}
\quad Is there hepatoblastoma? & NA & NA & NA & NA & NA \\
\arrayrulecolor{linegray}\hhline{------}
\rowcolor{rowgray}
\quad Is there hepatocellular carcinoma? & 90.9 & 92.3 & 92.5 & 93.2 & \textbf{97.8} \\
\arrayrulecolor{linegray}\hhline{------}
\rowcolor{rowgray}
\quad Is there hepatomegaly? & 72.1 & 83.0 & 85.0 & 78.5 & \textbf{87.3} \\
\arrayrulecolor{linegray}\hhline{------}
\rowcolor{rowgray}
\quad Is there portal venous occlusion? & 78.1 & 84.6 & 85.8 & 84.8 & \textbf{92.6} \\
\arrayrulecolor{linegray}\hhline{------}
\rowcolor{rowgray}
\quad Is there viral hepatitis? & 79.6 & 83.6 & 83.9 & 82.7 & \textbf{90.7} \\
\arrayrulecolor{linegray}\hhline{------}
\rowcolor{rowgray}
\quad is there any extrahepatic biliary ductal dilation? & 75.7 & 78.0 & 76.5 & 76.6 & \textbf{90.7} \\
\arrayrulecolor{black}\midrule

\textbf{Mass/Tumor findings} & & & & & \\

\quad Are there any Wilms tumors? & NA & NA & NA & NA & NA \\
\arrayrulecolor{linegray}\hhline{------}

\quad Are there any bladder transitional cell carcinomas? & 70.4 & 74.4 & 74.7 & 67.8 & \textbf{85.2} \\
\arrayrulecolor{linegray}\hhline{------}

\quad Are there any neuroendocrine tumors? & 64.4 & 73.4 & 73.8 & 67.7 & \textbf{84.2} \\
\arrayrulecolor{linegray}\hhline{------}

\quad Are there any rectal carcinomas? & 66.9 & 69.9 & 74.9 & 73.7 & \textbf{90.0} \\
\arrayrulecolor{linegray}\hhline{------}

\quad Are there any solid pseudopapillary tumors? & NA & NA & NA & NA & NA \\
\arrayrulecolor{linegray}\hhline{------}

\quad Are there any testicular masses? & 77.3 & 81.4 & 85.1 & 66.2 & \textbf{95.4} \\
\arrayrulecolor{linegray}\hhline{------}

\quad Are there any vaginal or vulvar cancers? & \textbf{97.3} & 94.8 & 92.7 & 49.9 & 97.2 \\
\arrayrulecolor{linegray}\hhline{------}

\quad Is there adenoma malignum? & 74.1 & 72.3 & 74.1 & 65.4 & \textbf{83.5} \\
\arrayrulecolor{linegray}\hhline{------}

\quad Is there cholangiocarcinoma? & 81.0 & 84.4 & 84.7 & 84.2 & \textbf{96.9} \\
\arrayrulecolor{linegray}\hhline{------}

\quad Is there fibrolamellar carcinoma? & 8.1 & 56.6 & 35.6 & \textbf{87.4} & 74.7 \\
\arrayrulecolor{linegray}\hhline{------}

\quad Is there mesenteric carcinoid tumor? & 57.4 & 68.5 & 64.5 & 67.0 & \textbf{88.3} \\
\arrayrulecolor{linegray}\hhline{------}

\quad Is there periductal/intraductal cholangiocarcinoma? & 63.0 & 72.3 & 87.2 & 85.1 & \textbf{98.7} \\
\arrayrulecolor{linegray}\hhline{------}

\quad Is there peritoneal carcinomatosis or omental caking? & 82.7 & 87.9 & 88.1 & 84.2 & \textbf{93.9} \\
\arrayrulecolor{linegray}\hhline{------}

\quad Is there prostate cancer? & 92.0 & 93.1 & 91.6 & 90.3 & \textbf{97.2} \\
\arrayrulecolor{black}\midrule
\rowcolor{rowgray}
\textbf{Other findings} & & & & & \\
\rowcolor{rowgray}
\quad Are there any angiomyolipomas? & 62.6 & 64.6 & 65.4 & 65.4 & \textbf{66.8} \\
\arrayrulecolor{linegray}\hhline{------}
\rowcolor{rowgray}
\quad Are there any bladder stones? & 81.3 & 83.3 & 85.0 & 78.4 & \textbf{85.7} \\
\arrayrulecolor{linegray}\hhline{------}
\rowcolor{rowgray}
\quad Are there any dysplastic nodules? & 52.4 & 48.7 & \textbf{71.1} & 61.9 & 58.8 \\
\arrayrulecolor{linegray}\hhline{------}
\rowcolor{rowgray}
\quad Are there any enteric diverticula? & 72.5 & 75.1 & 74.2 & 73.0 & \textbf{77.2} \\
\arrayrulecolor{linegray}\hhline{------}
\rowcolor{rowgray}
\quad Are there any femoral/ventral/Spigelian hernias? & 73.1 & 76.9 & 75.2 & 72.9 & \textbf{83.9} \\
\arrayrulecolor{linegray}\hhline{------}
\rowcolor{rowgray}
\quad Are there any hydroceles? & 78.9 & 79.5 & 83.5 & 81.7 & \textbf{85.8} \\
\arrayrulecolor{linegray}\hhline{------}
\rowcolor{rowgray}
\quad Are there any iliopsoas muscle abscesses? & 79.5 & 76.8 & 76.9 & 77.4 & \textbf{90.9} \\
\arrayrulecolor{linegray}\hhline{------}
\rowcolor{rowgray}
\quad Are there any inguinal hernias? & 79.7 & \textbf{81.1} & 79.1 & 77.9 & 80.5 \\
\arrayrulecolor{linegray}\hhline{------}
\rowcolor{rowgray}
\quad Are there any intraductal papillary mucinous neoplasms (IPMN)? & 72.6 & 75.3 & 75.6 & 66.8 & \textbf{77.5} \\
\arrayrulecolor{linegray}\hhline{------}
\rowcolor{rowgray}
\quad Are there any intraductal papillary mucinous neoplasms with worrisome features? & 51.5 & 66.7 & \textbf{70.4} & 64.1 & 65.1 \\
\arrayrulecolor{linegray}\hhline{------}
\rowcolor{rowgray}
\quad Are there any leiomyomas? & 75.6 & 78.3 & 79.6 & 77.5 & \textbf{90.3} \\
\arrayrulecolor{linegray}\hhline{------}
\rowcolor{rowgray}
\quad Are there any leiomyosarcomas? & 73.0 & 83.9 & 84.3 & 90.7 & \textbf{96.4} \\
\arrayrulecolor{linegray}\hhline{------}
\rowcolor{rowgray}
\quad Are there any oncocytomas? & NA & NA & NA & NA & NA \\
\arrayrulecolor{linegray}\hhline{------}
\rowcolor{rowgray}
\quad Are there any osteosclerotic lesions? & 71.0 & 73.9 & 71.5 & 70.4 & \textbf{79.7} \\
\arrayrulecolor{linegray}\hhline{------}
\rowcolor{rowgray}
\quad Are there any pheochromocytomas? & 82.3 & \textbf{93.5} & 80.9 & 73.1 & 87.8 \\
\arrayrulecolor{linegray}\hhline{------}
\rowcolor{rowgray}
\quad Are there any scrotal hematomas? & NA & NA & NA & NA & NA \\
\arrayrulecolor{linegray}\hhline{------}
\rowcolor{rowgray}
\quad Are there any subcapsular hemorrhages? & 80.6 & 82.1 & 82.1 & 79.5 & \textbf{87.8} \\
\arrayrulecolor{linegray}\hhline{------}
\rowcolor{rowgray}
\quad Are there any urachal diverticula? & 60.4 & 82.0 & \textbf{88.9} & 76.8 & 84.3 \\
\arrayrulecolor{linegray}\hhline{------}
\rowcolor{rowgray}
\quad Are there any ureteroceles? & 11.5 & 10.1 & 37.4 & 14.5 & \textbf{75.4} \\
\arrayrulecolor{linegray}\hhline{------}
\rowcolor{rowgray}
\quad Are there any varicoceles? & 61.4 & 65.1 & 60.4 & 71.7 & \textbf{81.4} \\
\arrayrulecolor{linegray}\hhline{------}
\rowcolor{rowgray}
\quad Are there changes status post hysterectomy? & 77.4 & 79.7 & 79.9 & 77.7 & \textbf{90.2} \\
\arrayrulecolor{linegray}\hhline{------}
\rowcolor{rowgray}
\quad Is anasarca suspected based on the imaging findings? & 95.3 & 96.2 & 95.4 & 95.5 & \textbf{96.4} \\
\arrayrulecolor{linegray}\hhline{------}
\rowcolor{rowgray}
\quad Is appendicitis suspected based on the imaging findings? & 82.5 & 87.6 & 86.0 & 82.3 & \textbf{92.4} \\
\arrayrulecolor{linegray}\hhline{------}
\rowcolor{rowgray}
\quad Is hiatal hernia present? & 73.3 & 76.5 & 75.1 & 72.2 & \textbf{81.0} \\
\arrayrulecolor{linegray}\hhline{------}
\rowcolor{rowgray}
\quad Is metastatic disease suspected based on the imaging findings? & 74.5 & 79.0 & 77.8 & 76.0 & \textbf{87.0} \\
\arrayrulecolor{linegray}\hhline{------}
\rowcolor{rowgray}
\quad Is osteopenia suspected based on the imaging findings? & 83.4 & \textbf{84.8} & 82.6 & 81.0 & 84.3 \\
\arrayrulecolor{linegray}\hhline{------}
\rowcolor{rowgray}
\quad Is there Fournier's gangrene / necrotizing fasciitis? & 94.5 & 96.6 & \textbf{99.8} & 94.9 & 99.0 \\
\arrayrulecolor{linegray}\hhline{------}
\rowcolor{rowgray}
\quad Is there adnexal torsion? & 99.7 & 99.8 & 88.5 & 86.5 & \textbf{99.9} \\
\arrayrulecolor{linegray}\hhline{------}
\rowcolor{rowgray}
\quad Is there angiosarcoma? & 87.8 & \textbf{98.0} & 98.0 & 92.3 & 39.7 \\
\arrayrulecolor{linegray}\hhline{------}
\rowcolor{rowgray}
\quad Is there any adenomyosis? & 60.4 & 62.3 & 61.4 & 55.4 & \textbf{71.3} \\
\arrayrulecolor{linegray}\hhline{------}
\rowcolor{rowgray}
\quad Is there any ascites? & 84.9 & 87.6 & 88.2 & 87.2 & \textbf{91.6} \\
\arrayrulecolor{linegray}\hhline{------}
\rowcolor{rowgray}
\quad Is there any consolidation? & 81.6 & 83.8 & 82.8 & 79.1 & \textbf{85.7} \\
\arrayrulecolor{linegray}\hhline{------}
\rowcolor{rowgray}
\quad Is there any evidence of enteric contrast leak suggestive of postoperative leakage/fistula after surgery? & 83.4 & 87.7 & 88.2 & 84.2 & \textbf{92.8} \\
\arrayrulecolor{linegray}\hhline{------}
\rowcolor{rowgray}
\quad Is there any pleural effusion? & 89.8 & 91.7 & 91.0 & 89.7 & \textbf{96.5} \\
\arrayrulecolor{linegray}\hhline{------}
\rowcolor{rowgray}
\quad Is there ascending cholangitis? & 67.4 & 79.1 & 89.3 & 79.1 & \textbf{98.1} \\
\arrayrulecolor{linegray}\hhline{------}
\rowcolor{rowgray}
\quad Is there bladder rupture? & 75.7 & 73.5 & 62.4 & 32.9 & \textbf{91.6} \\
\arrayrulecolor{linegray}\hhline{------}
\rowcolor{rowgray}
\quad Is there diverticulitis? & 73.9 & 78.5 & 77.2 & 76.3 & \textbf{92.8} \\
\arrayrulecolor{linegray}\hhline{------}
\rowcolor{rowgray}
\quad Is there epididymitis? & 95.7 & 84.2 & 81.8 & 78.2 & \textbf{99.0} \\
\arrayrulecolor{linegray}\hhline{------}
\rowcolor{rowgray}
\quad Is there epithelioid hemangioendothelioma? & NA & NA & NA & NA & NA \\
\arrayrulecolor{linegray}\hhline{------}
\rowcolor{rowgray}
\quad Is there evidence of sleeve gastrectomy or fundoplication? & 69.6 & 73.3 & 68.0 & 69.0 & \textbf{90.9} \\
\arrayrulecolor{linegray}\hhline{------}
\rowcolor{rowgray}
\quad Is there hemochromatosis? & NA & NA & NA & NA & NA \\
\arrayrulecolor{linegray}\hhline{------}
\rowcolor{rowgray}
\quad Is there hiatal hernia? & 73.2 & 76.6 & 75.1 & 72.2 & \textbf{81.0} \\
\arrayrulecolor{linegray}\hhline{------}
\rowcolor{rowgray}
\quad Is there ileus? & 88.1 & 90.6 & 90.4 & 87.7 & \textbf{93.0} \\
\arrayrulecolor{linegray}\hhline{------}
\rowcolor{rowgray}
\quad Is there intussusception? & 57.2 & 64.8 & 70.1 & 60.4 & \textbf{71.6} \\
\arrayrulecolor{linegray}\hhline{------}
\rowcolor{rowgray}
\quad Is there nephritis? & 74.5 & 76.6 & 78.1 & 76.4 & \textbf{91.0} \\
\arrayrulecolor{linegray}\hhline{------}
\rowcolor{rowgray}
\quad Is there primary sclerosing cholangitis? & 80.6 & 74.7 & 79.3 & 77.3 & \textbf{91.1} \\
\arrayrulecolor{linegray}\hhline{------}
\rowcolor{rowgray}
\quad Is there prostatitis? & 66.2 & 67.8 & 61.2 & 64.7 & \textbf{91.3} \\
\arrayrulecolor{linegray}\hhline{------}
\rowcolor{rowgray}
\quad Is there prostatomegaly or prostatic hyperplasia? & 83.4 & 85.1 & 84.3 & 82.8 & \textbf{90.2} \\
\arrayrulecolor{linegray}\hhline{------}
\rowcolor{rowgray}
\quad Is there pseudomembranous colitis? & 85.3 & 90.9 & 86.8 & 82.6 & \textbf{96.9} \\
\arrayrulecolor{linegray}\hhline{------}
\rowcolor{rowgray}
\quad Is there pseudomyxoma peritonei? & 44.1 & 60.8 & 69.4 & 69.6 & \textbf{74.7} \\
\arrayrulecolor{linegray}\hhline{------}
\rowcolor{rowgray}
\quad Is there retroperitoneal fibrosis? & 55.0 & 54.0 & 73.2 & 63.6 & \textbf{88.2} \\
\arrayrulecolor{linegray}\hhline{------}
\rowcolor{rowgray}
\quad Is there retroperitoneal hemorrhage or retroperitoneal hematomas? & 87.1 & 89.2 & 88.7 & 86.0 & \textbf{93.7} \\
\arrayrulecolor{linegray}\hhline{------}
\rowcolor{rowgray}
\quad Is there retroperitoneal liposarcoma? & 81.6 & \textbf{85.1} & 80.4 & 61.7 & 78.8 \\
\arrayrulecolor{linegray}\hhline{------}
\rowcolor{rowgray}
\quad Is there submucosal edema, enterocolitis, or gastritis? & 79.7 & 82.1 & 82.5 & 79.9 & \textbf{87.9} \\
\arrayrulecolor{linegray}\hhline{------}
\rowcolor{rowgray}
\quad Is there testicular (segmental) infarct? & NA & NA & NA & NA & NA \\
\arrayrulecolor{linegray}\hhline{------}
\rowcolor{rowgray}
\quad Is there testicular torsion? & NA & NA & NA & NA & NA \\
\arrayrulecolor{linegray}\hhline{------}
\rowcolor{rowgray}
\quad Is there volvulus? & 81.9 & 82.5 & 84.1 & 78.1 & \textbf{89.5} \\
\arrayrulecolor{linegray}\hhline{------}
\rowcolor{rowgray}
\quad Is there xanthogranulomatous pyelonephritis? & 82.9 & 86.9 & 74.4 & 69.2 & \textbf{99.5} \\
\arrayrulecolor{black}\midrule

\textbf{Pancreas-related findings} & & & & & \\

\quad Are there any ductal pancreatic carcinomas? & 80.6 & 84.8 & 85.9 & 82.0 & \textbf{96.1} \\
\arrayrulecolor{linegray}\hhline{------}

\quad Are there any pancreatic tumors? & 75.1 & 77.3 & 78.1 & 76.2 & \textbf{85.9} \\
\arrayrulecolor{linegray}\hhline{------}

\quad Is chronic pancreatitis suspected based on imaging findings? & 72.6 & 75.0 & 73.2 & 72.9 & \textbf{83.2} \\
\arrayrulecolor{linegray}\hhline{------}

\quad Is cystic fibrosis suspected based on the pancreatic imaging findings? & 57.6 & 53.3 & 36.7 & 39.1 & \textbf{96.4} \\
\arrayrulecolor{linegray}\hhline{------}

\quad Is there acute pancreatitis? & 80.5 & 84.3 & 87.1 & 85.6 & \textbf{95.4} \\
\arrayrulecolor{linegray}\hhline{------}

\quad Is there an annular pancreas? & 43.5 & 76.6 & 62.3 & \textbf{81.0} & 41.3 \\
\arrayrulecolor{linegray}\hhline{------}

\quad Is there any pancreatic pseudocysts? & 73.9 & 76.0 & 77.1 & 69.0 & \textbf{87.7} \\
\arrayrulecolor{linegray}\hhline{------}

\quad Is there autoimmune pancreatitis? & NA & NA & NA & NA & NA \\
\arrayrulecolor{linegray}\hhline{------}

\quad Is there groove pancreatitis? & 39.8 & 65.3 & 69.0 & 72.9 & \textbf{88.0} \\
\arrayrulecolor{linegray}\hhline{------}

\quad Is there main pancreatic duct dilatation? & 74.2 & 78.2 & 78.1 & 75.8 & \textbf{88.4} \\
\arrayrulecolor{linegray}\hhline{------}

\quad Is there necrotizing pancreatitis? & 88.3 & 94.4 & 92.6 & 92.0 & \textbf{98.2} \\
\arrayrulecolor{linegray}\hhline{------}

\quad Is there pancreatic atrophy? & 70.3 & 72.9 & 71.7 & 71.1 & \textbf{81.7} \\
\arrayrulecolor{black}\midrule
\rowcolor{rowgray}
\textbf{Pelvic/GYN findings} & & & & & \\
\rowcolor{rowgray}
\quad Are there any cervical masses? & 85.0 & 90.9 & 85.9 & 88.5 & \textbf{94.3} \\
\arrayrulecolor{linegray}\hhline{------}
\rowcolor{rowgray}
\quad Are there any endometrial polyps? & 70.0 & 54.7 & 53.6 & \textbf{91.3} & 65.5 \\
\arrayrulecolor{linegray}\hhline{------}
\rowcolor{rowgray}
\quad Are there any endometriomas? & 82.1 & \textbf{85.5} & 78.8 & 71.4 & 82.5 \\
\arrayrulecolor{linegray}\hhline{------}
\rowcolor{rowgray}
\quad Are there any fractures of the pelvic girdle or sacrum? & 76.7 & 78.7 & 78.8 & 74.4 & \textbf{82.0} \\
\arrayrulecolor{linegray}\hhline{------}
\rowcolor{rowgray}
\quad Are there any intra-uterine devices? & 87.4 & 90.9 & 86.9 & 90.7 & \textbf{98.8} \\
\arrayrulecolor{linegray}\hhline{------}
\rowcolor{rowgray}
\quad Are there any ovarian cancers? & 80.5 & 84.6 & 84.5 & 81.6 & \textbf{96.2} \\
\arrayrulecolor{linegray}\hhline{------}
\rowcolor{rowgray}
\quad Are there any ovarian teratomas? & 73.6 & 75.9 & 70.3 & 73.9 & \textbf{79.4} \\
\arrayrulecolor{linegray}\hhline{------}
\rowcolor{rowgray}
\quad Are there any ovarian tumors? & 78.6 & 84.8 & 83.6 & 79.5 & \textbf{87.4} \\
\arrayrulecolor{linegray}\hhline{------}
\rowcolor{rowgray}
\quad Are there any uterine malformations? & 58.5 & 57.6 & 66.4 & 61.7 & \textbf{75.1} \\
\arrayrulecolor{linegray}\hhline{------}
\rowcolor{rowgray}
\quad Is there any lymphadenopathy in retroperitoneum, peritoneum or pelvis? & 65.5 & 68.7 & 70.1 & 68.2 & \textbf{75.3} \\
\arrayrulecolor{linegray}\hhline{------}
\rowcolor{rowgray}
\quad Is there deep infiltrative endometriosis? & 89.5 & 88.0 & 65.6 & 69.4 & \textbf{97.8} \\
\arrayrulecolor{linegray}\hhline{------}
\rowcolor{rowgray}
\quad Is there endometrial thickening? & 73.8 & 80.7 & 78.8 & 74.5 & \textbf{87.4} \\
\arrayrulecolor{linegray}\hhline{------}
\rowcolor{rowgray}
\quad Is there pelvic inflammatory disease? & 85.6 & 82.5 & 88.0 & 96.8 & \textbf{97.7} \\
\arrayrulecolor{linegray}\hhline{------}
\rowcolor{rowgray}
\quad Is there polycystic ovarian syndrome? & 88.1 & 61.7 & 12.0 & \textbf{96.1} & 69.4 \\
\arrayrulecolor{black}\midrule

\textbf{Spleen-related findings} & & & & & \\

\quad Are there any accessory spleens? & 54.6 & 56.7 & \textbf{57.4} & 55.5 & 55.9 \\
\arrayrulecolor{linegray}\hhline{------}

\quad Are there any splenic abscesses? & 73.6 & 79.7 & 82.3 & 62.9 & \textbf{99.9} \\
\arrayrulecolor{linegray}\hhline{------}

\quad Are there any splenic hamartomas? & 43.5 & 38.9 & 60.2 & 53.8 & \textbf{66.2} \\
\arrayrulecolor{linegray}\hhline{------}

\quad Are there any splenic infarcts? & 75.4 & 81.5 & 81.5 & 80.1 & \textbf{91.9} \\
\arrayrulecolor{linegray}\hhline{------}

\quad Are there any splenic lacerations? & 40.6 & 66.4 & 70.7 & 66.8 & \textbf{87.0} \\
\arrayrulecolor{linegray}\hhline{------}

\quad Are there any splenic pseudocysts? & \textbf{68.2} & 65.3 & 64.8 & 45.6 & 59.9 \\
\arrayrulecolor{linegray}\hhline{------}

\quad Is splenic lymphoma present? & 49.4 & 47.5 & 51.8 & 50.7 & \textbf{95.6} \\
\arrayrulecolor{linegray}\hhline{------}

\quad Is the spleen surgically absent? Are there any post-splenectomy changes? & 56.3 & \textbf{62.0} & 59.9 & 45.6 & 55.5 \\
\arrayrulecolor{linegray}\hhline{------}

\quad Is there splenomegaly? & 82.4 & 86.1 & 89.8 & 90.3 & \textbf{94.7} \\
\arrayrulecolor{black}\midrule
\rowcolor{rowgray}
\textbf{Vascular findings} & & & & & \\
\rowcolor{rowgray}
\quad Are there any aortic dissections? & 79.2 & 86.3 & 86.1 & 80.2 & \textbf{92.0} \\
\arrayrulecolor{linegray}\hhline{------}
\rowcolor{rowgray}
\quad Are there any aortic intramural hematomas? & 89.4 & 78.8 & 79.9 & 76.8 & \textbf{97.1} \\
\arrayrulecolor{linegray}\hhline{------}
\rowcolor{rowgray}
\quad Are there any aortic stents? & 96.4 & 97.5 & 96.8 & 92.8 & \textbf{98.4} \\
\arrayrulecolor{linegray}\hhline{------}
\rowcolor{rowgray}
\quad Are there any aortic valve calcifications? & 87.3 & 89.9 & 86.5 & 87.7 & \textbf{91.3} \\
\arrayrulecolor{linegray}\hhline{------}
\rowcolor{rowgray}
\quad Are there any deep venous thromboses? & 69.3 & 74.1 & 74.9 & 71.2 & \textbf{83.7} \\
\arrayrulecolor{linegray}\hhline{------}
\rowcolor{rowgray}
\quad Are there any penetrating atherosclerotic aortic ulcers? & 82.4 & 85.1 & 85.5 & 78.8 & \textbf{90.4} \\
\arrayrulecolor{linegray}\hhline{------}
\rowcolor{rowgray}
\quad Are there coronary artery calcifications? & 83.9 & 86.5 & 84.6 & 84.2 & \textbf{89.6} \\
\arrayrulecolor{linegray}\hhline{------}
\rowcolor{rowgray}
\quad Is abdominal aortic aneurysm suspected based on the imaging findings? & 85.9 & 91.4 & 90.1 & 87.4 & \textbf{95.7} \\
\arrayrulecolor{linegray}\hhline{------}
\rowcolor{rowgray}
\quad Is aortic atherosclerosis suspected based on the imaging findings? & 77.8 & 81.3 & 78.9 & 78.5 & \textbf{84.8} \\
\arrayrulecolor{linegray}\hhline{------}
\rowcolor{rowgray}
\quad Is arterial or aortic thrombosis suspected based on the imaging findings? & 75.6 & 80.3 & 78.3 & 76.5 & \textbf{87.1} \\
\end{longtable}
\end{center}

\vspace{2mm}
\noindent
\footnotesize \textit{Abbreviations:} AUC, area under the curve; NA, not applicable.

\newpage
\section{Performance on full set of RATE-Evals tasks on UCSF Chest CT test set}
\label{ref:chest-finding}

\begin{center}
\footnotesize
\begin{longtable}{p{5cm} >{\centering\arraybackslash}p{2.0cm} >{\centering\arraybackslash}p{2.0cm} >{\centering\arraybackslash}p{2.0cm} >{\centering\arraybackslash}p{2.0cm} >{\centering\arraybackslash}p{2.0cm}} 
\caption{Model Performance (AUC x 100) for UCSF Chest CT} \label{tab:auc_ucsf_chest_ct} \\
\arrayrulecolor{black}
\toprule
& \multicolumn{5}{c}{\textbf{Model AUC (x 100)}} \\
\cmidrule(lr){2-6} 
\textbf{Finding} & \textbf{LingShu} & \textbf{MedGemma} & \textbf{MedImageInsight} & \textbf{Merlin} & \textbf{\name} \\
\arrayrulecolor{black}
\midrule
\endfirsthead

\multicolumn{6}{c}{%
\tablename\ \thetable{} -- Model Performance (AUC x 100) for UCSF Chest CT \textit{(continued)}} \\
\arrayrulecolor{black}
\toprule
& \multicolumn{5}{c}{\textbf{Model AUC (x 100)}} \\
\cmidrule(lr){2-6} 
\textbf{Finding} & \textbf{LingShu} & \textbf{MedGemma} & \textbf{MedImageInsight} & \textbf{Merlin} & \textbf{\name} \\
\arrayrulecolor{black}
\midrule
\endhead

\arrayrulecolor{black}
\midrule
\multicolumn{6}{r}{\textit{Continued on next page}} \\
\endfoot

\arrayrulecolor{black}
\bottomrule
\endlastfoot
\rowcolor{rowgray}
\textbf{Bone/Fracture findings} & & & & & \\
\rowcolor{rowgray}
\quad Is there any fracture of rib or clavicle? & 73.5 & 74.9 & 75.0 & 70.8 & \textbf{84.4} \\
\arrayrulecolor{linegray}\hhline{------}
\rowcolor{rowgray}
\quad Is there any fracture of spine? & 73.8 & 74.7 & 74.2 & 71.4 & \textbf{91.7} \\
\arrayrulecolor{linegray}\hhline{------}
\rowcolor{rowgray}
\quad Is there any osteolytic lesion? & 73.3 & 78.3 & 75.1 & 68.4 & \textbf{84.9} \\
\arrayrulecolor{black}\midrule

\textbf{Breast findings} & & & & & \\

\quad Is there any breast implant? & 91.2 & 94.4 & 93.7 & 85.9 & \textbf{99.1} \\
\arrayrulecolor{linegray}\hhline{------}

\quad Is there any breast mass? & 79.2 & 79.4 & 79.2 & 72.9 & \textbf{83.5} \\
\arrayrulecolor{black}\midrule
\rowcolor{rowgray}
\textbf{Cardiac findings} & & & & & \\
\rowcolor{rowgray}
\quad Are there any coronary soft plaques? & 66.9 & 86.8 & 78.0 & 75.4 & \textbf{90.2} \\
\arrayrulecolor{linegray}\hhline{------}
\rowcolor{rowgray}
\quad Is cardiac tamponade suspected? & 77.1 & 78.7 & 85.6 & 81.4 & \textbf{92.6} \\
\arrayrulecolor{linegray}\hhline{------}
\rowcolor{rowgray}
\quad Is cardiomegaly suspected? & 85.8 & 89.0 & 90.3 & 83.4 & \textbf{90.5} \\
\arrayrulecolor{linegray}\hhline{------}
\rowcolor{rowgray}
\quad Is dilated cardiomyopathy suspected? & \textbf{90.8} & 85.7 & 89.7 & 85.5 & 89.9 \\
\arrayrulecolor{linegray}\hhline{------}
\rowcolor{rowgray}
\quad Is hypertrophic cardiomyopathy suspected? & 93.9 & 77.4 & 70.4 & 79.9 & \textbf{97.8} \\
\arrayrulecolor{linegray}\hhline{------}
\rowcolor{rowgray}
\quad Is ischemic cardiomyopathy suspected? & 62.6 & 71.2 & 73.7 & 76.7 & \textbf{85.1} \\
\arrayrulecolor{linegray}\hhline{------}
\rowcolor{rowgray}
\quad Is there any coronary atherosclerosis? & 75.4 & 76.9 & 76.7 & 74.8 & \textbf{85.1} \\
\arrayrulecolor{linegray}\hhline{------}
\rowcolor{rowgray}
\quad Is there any heart valve calcification? & 75.7 & 78.0 & 77.5 & 76.7 & \textbf{88.4} \\
\arrayrulecolor{linegray}\hhline{------}
\rowcolor{rowgray}
\quad Is there any heart valve replacement? & 88.3 & 94.4 & 90.0 & 91.2 & \textbf{99.4} \\
\arrayrulecolor{black}\midrule

\textbf{Chest/Lung findings} & & & & & \\

\quad Is COVID-19 suspected? & 88.8 & 86.1 & 81.5 & 80.9 & \textbf{93.6} \\
\arrayrulecolor{linegray}\hhline{------}

\quad Is aspiration pneumonia suspected? & 81.6 & 84.0 & 85.5 & 79.4 & \textbf{90.8} \\
\arrayrulecolor{linegray}\hhline{------}

\quad Is pulmonary edema suspected? & 91.1 & 91.7 & 91.2 & 89.4 & \textbf{93.0} \\
\arrayrulecolor{linegray}\hhline{------}

\quad Is pulmonary embolism suspected? & 81.2 & \textbf{85.5} & 84.8 & 77.6 & 85.0 \\
\arrayrulecolor{linegray}\hhline{------}

\quad Is pulmonary hypertension suspected? & 82.3 & 84.9 & 85.5 & 78.6 & \textbf{87.8} \\
\arrayrulecolor{linegray}\hhline{------}

\quad Is radiation pneumonitis suspected? & 70.8 & 80.8 & 84.4 & 67.9 & \textbf{92.0} \\
\arrayrulecolor{linegray}\hhline{------}

\quad Is there any atypical cyst in the lung? & 55.7 & 70.6 & 58.3 & 58.4 & \textbf{86.6} \\
\arrayrulecolor{linegray}\hhline{------}

\quad Is there any calcification in the lung? & 60.1 & 62.4 & 62.1 & 60.3 & \textbf{80.3} \\
\arrayrulecolor{linegray}\hhline{------}

\quad Is there any cavitation in the lung? & 63.0 & 74.6 & 77.8 & 62.6 & \textbf{85.6} \\
\arrayrulecolor{linegray}\hhline{------}

\quad Is there any chest tube? & 94.6 & 96.8 & 96.9 & 92.2 & \textbf{98.8} \\
\arrayrulecolor{linegray}\hhline{------}

\quad Is there any consolidation in the lung? & 81.4 & 85.8 & 86.3 & 79.3 & \textbf{91.1} \\
\arrayrulecolor{linegray}\hhline{------}

\quad Is there any ground-glass nodules in the lung? & 64.3 & 66.9 & 66.6 & 62.3 & \textbf{76.3} \\
\arrayrulecolor{linegray}\hhline{------}

\quad Is there any mass in the lung? & 66.4 & 73.8 & 75.3 & 65.9 & \textbf{82.1} \\
\arrayrulecolor{linegray}\hhline{------}

\quad Is there any part-solid nodules in the lung? & 75.5 & 77.5 & 79.2 & 73.0 & \textbf{88.6} \\
\arrayrulecolor{linegray}\hhline{------}

\quad Is there any pneumoconiosis? & 90.2 & 92.3 & 88.0 & 81.2 & \textbf{98.0} \\
\arrayrulecolor{linegray}\hhline{------}

\quad Is there any pneumomediastinum? & 90.4 & 91.8 & 93.4 & 90.9 & \textbf{95.8} \\
\arrayrulecolor{linegray}\hhline{------}

\quad Is there any pneumothorax? & 91.3 & 93.2 & 94.8 & 88.8 & \textbf{95.6} \\
\arrayrulecolor{linegray}\hhline{------}

\quad Is there any post-lung resection scar? & 77.6 & 84.5 & 82.8 & 70.7 & \textbf{94.4} \\
\arrayrulecolor{linegray}\hhline{------}

\quad Is there any solid nodules in the lung? & 62.3 & 64.6 & 65.8 & 60.5 & \textbf{76.2} \\
\arrayrulecolor{black}\midrule
\rowcolor{rowgray}
\textbf{Device/Procedural findings} & & & & & \\
\rowcolor{rowgray}
\quad Is miliary tuberculosis suspected? & 86.8 & \textbf{91.7} & 89.2 & 69.6 & 83.8 \\
\arrayrulecolor{linegray}\hhline{------}
\rowcolor{rowgray}
\quad Is non-tuberculous mycobacterial infection suspected? & 83.3 & 86.4 & 84.6 & 81.7 & \textbf{93.3} \\
\arrayrulecolor{linegray}\hhline{------}
\rowcolor{rowgray}
\quad Is primary tuberculosis suspected? & 83.2 & 89.2 & \textbf{92.2} & 74.7 & 83.0 \\
\arrayrulecolor{linegray}\hhline{------}
\rowcolor{rowgray}
\quad Is there any central venous catheter? & 81.4 & 91.5 & 83.7 & 82.0 & \textbf{95.6} \\
\arrayrulecolor{linegray}\hhline{------}
\rowcolor{rowgray}
\quad Is there any endotracheal tube? & 97.4 & 98.5 & 98.0 & 97.7 & \textbf{99.1} \\
\arrayrulecolor{black}\midrule

\textbf{Gastrointestinal findings} & & & & & \\

\quad Is there any nasogastric tube? & 96.0 & 97.7 & 95.7 & 95.7 & \textbf{98.1} \\
\arrayrulecolor{black}\midrule
\rowcolor{rowgray}
\textbf{Liver-related findings} & & & & & \\
\rowcolor{rowgray}
\quad Is cirrhosis suspected? & 85.7 & 91.1 & 92.9 & 87.5 & \textbf{94.4} \\
\arrayrulecolor{linegray}\hhline{------}
\rowcolor{rowgray}
\quad Is diffuse fatty liver / steatosis suspected? & 80.2 & 81.7 & 85.0 & 82.9 & \textbf{89.8} \\
\arrayrulecolor{linegray}\hhline{------}
\rowcolor{rowgray}
\quad Is there any liver cyst? & 73.3 & 75.0 & 77.5 & 70.7 & \textbf{88.2} \\
\arrayrulecolor{linegray}\hhline{------}
\rowcolor{rowgray}
\quad Is there any liver mass? & 78.1 & 78.7 & 79.4 & 72.8 & \textbf{83.9} \\
\arrayrulecolor{black}\midrule

\textbf{Mass/Tumor findings} & & & & & \\

\quad Is there any mediastinal mass? & 67.8 & 73.2 & 74.5 & 62.7 & \textbf{86.4} \\
\arrayrulecolor{linegray}\hhline{------}

\quad Is there any tracheal mass? & 82.4 & 88.4 & 3.1 & 18.4 & \textbf{90.5} \\
\arrayrulecolor{black}\midrule
\rowcolor{rowgray}
\textbf{Other findings} & & & & & \\
\rowcolor{rowgray}
\quad Is aspergillosis suspected? & 82.8 & 86.9 & 86.2 & 75.3 & \textbf{92.6} \\
\arrayrulecolor{linegray}\hhline{------}
\rowcolor{rowgray}
\quad Is sarcoidosis suspected? & 75.8 & 78.3 & 81.1 & 73.1 & \textbf{94.8} \\
\arrayrulecolor{linegray}\hhline{------}
\rowcolor{rowgray}
\quad Is there any ECMO? & \textbf{99.0} & 99.0 & 98.6 & 97.9 & 98.7 \\
\arrayrulecolor{linegray}\hhline{------}
\rowcolor{rowgray}
\quad Is there any arthritis? & 78.9 & \textbf{85.5} & 81.6 & 75.4 & 80.5 \\
\arrayrulecolor{linegray}\hhline{------}
\rowcolor{rowgray}
\quad Is there any atelectasis? & 77.9 & 80.0 & 79.9 & 74.7 & \textbf{85.4} \\
\arrayrulecolor{linegray}\hhline{------}
\rowcolor{rowgray}
\quad Is there any bronchiectasis? & 79.9 & 82.7 & 82.7 & 75.8 & \textbf{88.4} \\
\arrayrulecolor{linegray}\hhline{------}
\rowcolor{rowgray}
\quad Is there any bulla, bleb, or pneumatocele? & 76.9 & 81.1 & 78.5 & 72.5 & \textbf{89.1} \\
\arrayrulecolor{linegray}\hhline{------}
\rowcolor{rowgray}
\quad Is there any carotid atherosclerosis? & 64.2 & 58.7 & 64.3 & 68.6 & \textbf{83.0} \\
\arrayrulecolor{linegray}\hhline{------}
\rowcolor{rowgray}
\quad Is there any centrilobular emphysema? & 81.0 & 85.7 & 83.8 & 79.1 & \textbf{92.7} \\
\arrayrulecolor{linegray}\hhline{------}
\rowcolor{rowgray}
\quad Is there any crazy paving pattern? & 89.8 & 90.4 & 89.5 & 87.2 & \textbf{91.8} \\
\arrayrulecolor{linegray}\hhline{------}
\rowcolor{rowgray}
\quad Is there any empyema? & 88.8 & 90.1 & 95.2 & 85.2 & \textbf{97.0} \\
\arrayrulecolor{linegray}\hhline{------}
\rowcolor{rowgray}
\quad Is there any goiter? & 64.3 & 65.1 & 65.8 & 62.0 & \textbf{82.3} \\
\arrayrulecolor{linegray}\hhline{------}
\rowcolor{rowgray}
\quad Is there any ground-glass opacity? & 70.6 & 74.0 & 74.3 & 67.8 & \textbf{83.6} \\
\arrayrulecolor{linegray}\hhline{------}
\rowcolor{rowgray}
\quad Is there any hemothorax? & 94.0 & 93.9 & 95.4 & 93.2 & \textbf{96.8} \\
\arrayrulecolor{linegray}\hhline{------}
\rowcolor{rowgray}
\quad Is there any hiatal hernia? & 80.5 & 84.6 & 84.7 & 77.9 & \textbf{93.0} \\
\arrayrulecolor{linegray}\hhline{------}
\rowcolor{rowgray}
\quad Is there any honeycombing? & 97.6 & \textbf{98.9} & 98.7 & 94.4 & 98.9 \\
\arrayrulecolor{linegray}\hhline{------}
\rowcolor{rowgray}
\quad Is there any interlobular septal thickening? & 87.0 & 88.3 & 86.7 & 83.2 & \textbf{89.4} \\
\arrayrulecolor{linegray}\hhline{------}
\rowcolor{rowgray}
\quad Is there any lymphadenopathy in axilla? & 72.1 & 73.6 & 72.3 & 67.8 & \textbf{84.6} \\
\arrayrulecolor{linegray}\hhline{------}
\rowcolor{rowgray}
\quad Is there any lymphadenopathy in mediastinum or hila? & 75.8 & 79.1 & 79.8 & 74.1 & \textbf{87.4} \\
\arrayrulecolor{linegray}\hhline{------}
\rowcolor{rowgray}
\quad Is there any lymphadenopathy in neck? & 69.0 & 72.1 & 73.2 & 65.7 & \textbf{77.8} \\
\arrayrulecolor{linegray}\hhline{------}
\rowcolor{rowgray}
\quad Is there any mosaic attenuation? & 79.4 & 81.7 & 81.0 & 74.6 & \textbf{88.0} \\
\arrayrulecolor{linegray}\hhline{------}
\rowcolor{rowgray}
\quad Is there any mucous plugging? & 74.5 & 76.2 & 75.6 & 70.9 & \textbf{84.1} \\
\arrayrulecolor{linegray}\hhline{------}
\rowcolor{rowgray}
\quad Is there any myxoma? & 6.4 & 8.6 & 1.4 & \textbf{32.5} & 17.2 \\
\arrayrulecolor{linegray}\hhline{------}
\rowcolor{rowgray}
\quad Is there any osteopenia? & 80.3 & \textbf{80.4} & 79.0 & 71.5 & 75.2 \\
\arrayrulecolor{linegray}\hhline{------}
\rowcolor{rowgray}
\quad Is there any osteosclerotic lesion? & 69.9 & 72.2 & 69.9 & 67.2 & \textbf{80.8} \\
\arrayrulecolor{linegray}\hhline{------}
\rowcolor{rowgray}
\quad Is there any pace-maker or defibrillator? & 95.4 & 97.9 & 96.5 & 95.6 & \textbf{99.0} \\
\arrayrulecolor{linegray}\hhline{------}
\rowcolor{rowgray}
\quad Is there any panbronchiolitis? & 51.1 & 45.8 & 67.8 & 71.3 & \textbf{93.8} \\
\arrayrulecolor{linegray}\hhline{------}
\rowcolor{rowgray}
\quad Is there any paraseptal emphysema? & 79.1 & 84.2 & 81.9 & 79.3 & \textbf{91.0} \\
\arrayrulecolor{linegray}\hhline{------}
\rowcolor{rowgray}
\quad Is there any peribronchial thickening? & 73.5 & 77.3 & 75.5 & 69.3 & \textbf{83.7} \\
\arrayrulecolor{linegray}\hhline{------}
\rowcolor{rowgray}
\quad Is there any pericardial effusion? & 75.2 & 77.4 & 78.3 & 73.9 & \textbf{87.7} \\
\arrayrulecolor{linegray}\hhline{------}
\rowcolor{rowgray}
\quad Is there any pleural effusion? & 92.5 & 94.2 & 94.4 & 89.2 & \textbf{97.6} \\
\arrayrulecolor{linegray}\hhline{------}
\rowcolor{rowgray}
\quad Is there any pleural plaques? & 79.1 & 78.9 & 79.7 & 81.2 & \textbf{90.7} \\
\arrayrulecolor{linegray}\hhline{------}
\rowcolor{rowgray}
\quad Is there any pleural thickening? & 82.5 & 85.3 & 84.7 & 75.7 & \textbf{89.3} \\
\arrayrulecolor{linegray}\hhline{------}
\rowcolor{rowgray}
\quad Is there any reticulation or reticular pattern? & 80.6 & 83.7 & 82.9 & 75.2 & \textbf{89.4} \\
\arrayrulecolor{linegray}\hhline{------}
\rowcolor{rowgray}
\quad Is there any sternotomy? & 88.4 & 94.2 & 90.4 & 84.4 & \textbf{96.4} \\
\arrayrulecolor{linegray}\hhline{------}
\rowcolor{rowgray}
\quad Is there any teratoma? & NA & NA & NA & NA & NA \\
\arrayrulecolor{linegray}\hhline{------}
\rowcolor{rowgray}
\quad Is there any thymoma? & 46.2 & 55.9 & 61.8 & 51.6 & \textbf{91.0} \\
\arrayrulecolor{linegray}\hhline{------}
\rowcolor{rowgray}
\quad Is there any thyroid nodule? & 68.8 & 69.3 & 69.9 & 65.3 & \textbf{75.0} \\
\arrayrulecolor{linegray}\hhline{------}
\rowcolor{rowgray}
\quad Is there any tree-in-bud or focal bronchiolitis? & 76.2 & 79.1 & 77.9 & 71.5 & \textbf{88.7} \\
\arrayrulecolor{black}\midrule

\textbf{Pancreas-related findings} & & & & & \\

\quad Is chronic pancreatitis suspected? & 71.6 & 71.0 & 75.3 & 71.7 & \textbf{87.8} \\
\arrayrulecolor{black}\midrule
\rowcolor{rowgray}
\textbf{Spleen-related findings} & & & & & \\
\rowcolor{rowgray}
\quad Is splenomegaly suspected? & 90.4 & 93.0 & 93.3 & 91.4 & \textbf{94.8} \\
\arrayrulecolor{black}\midrule

\textbf{Vascular findings} & & & & & \\

\quad Is aortic dissection suspected? & 72.4 & 77.0 & 80.5 & 70.1 & \textbf{85.2} \\
\arrayrulecolor{linegray}\hhline{------}

\quad Is there any aberrant subclavian artery? & 50.0 & 52.3 & 58.0 & 51.6 & \textbf{80.8} \\
\arrayrulecolor{linegray}\hhline{------}

\quad Is there any aortic aneurysm? & 78.4 & 83.6 & 81.4 & 75.5 & \textbf{91.7} \\
\arrayrulecolor{linegray}\hhline{------}

\quad Is there any aortic atherosclerosis? & 77.4 & 80.3 & 79.6 & 79.1 & \textbf{84.4} \\
\arrayrulecolor{linegray}\hhline{------}

\quad Is there any coronary artery stenting? & 69.3 & 70.6 & 71.6 & 71.0 & \textbf{88.1} \\
\arrayrulecolor{linegray}\hhline{------}

\quad Is there any post-coronary artery bypass graft? & 90.1 & 95.6 & 91.0 & 85.5 & \textbf{98.0} \\
\arrayrulecolor{linegray}\hhline{------}

\quad Is there any single coronary artery? & 29.0 & 29.6 & 41.4 & 29.1 & \textbf{42.1} \\
\end{longtable}
\end{center}

\vspace{2mm}
\noindent
\footnotesize \textit{Abbreviations:} AUC, area under the curve; NA, not applicable.

\newpage
\section{Performance on full set of RATE-Evals tasks on UCSF Head CT test set}
\label{ref:brain-finding}

\begin{center}
\footnotesize
\begin{longtable}{p{5cm} >{\centering\arraybackslash}p{2.0cm} >{\centering\arraybackslash}p{2.0cm} >{\centering\arraybackslash}p{2.0cm} >{\centering\arraybackslash}p{2.0cm} >{\centering\arraybackslash}p{2.0cm}} 
\caption{Model Performance (AUC x 100) for UCSF Head CT} \label{tab:auc_ucsf_brain_ct} \\
\arrayrulecolor{black}
\toprule
& \multicolumn{5}{c}{\textbf{Model AUC (x 100)}} \\
\cmidrule(lr){2-6} 
\textbf{Finding} & \textbf{LingShu} & \textbf{MedGemma} & \textbf{MedImageInsight} & \textbf{Merlin} & \textbf{\name} \\
\arrayrulecolor{black}
\midrule
\endfirsthead

\multicolumn{6}{c}{%
\tablename\ \thetable{} -- Model Performance (AUC x 100) for UCSF Brain CT \textit{(continued)}} \\
\arrayrulecolor{black}
\toprule
& \multicolumn{5}{c}{\textbf{Model AUC (x 100)}} \\
\cmidrule(lr){2-6} 
\textbf{Finding} & \textbf{LingShu} & \textbf{MedGemma} & \textbf{MedImageInsight} & \textbf{Merlin} & \textbf{\name} \\
\arrayrulecolor{black}
\midrule
\endhead

\arrayrulecolor{black}
\midrule
\multicolumn{6}{r}{\textit{Continued on next page}} \\
\endfoot

\arrayrulecolor{black}
\bottomrule
\endlastfoot
\rowcolor{rowgray}
\textbf{Bone/Fracture findings} & & & & & \\
\rowcolor{rowgray}
\quad Are there any fractures of skull vault (calvarial fracture)? & 74.0 & 78.3 & 78.2 & 66.3 & \textbf{78.9} \\
\arrayrulecolor{linegray}\hhline{------}
\rowcolor{rowgray}
\quad Are there any osteolytic lesions? & 52.2 & 71.4 & \textbf{74.4} & 63.0 & 72.0 \\
\arrayrulecolor{black}\midrule

\textbf{Brain findings} & & & & & \\

\quad Are there any brain tumors? & 62.8 & 65.9 & 66.2 & 56.4 & \textbf{86.2} \\
\arrayrulecolor{linegray}\hhline{------}

\quad Is cerebral edema present ? & 70.7 & 76.3 & 79.5 & 70.4 & \textbf{92.1} \\
\arrayrulecolor{linegray}\hhline{------}

\quad Is hydrocephalus present? & 79.8 & 83.0 & 89.8 & 81.7 & \textbf{95.2} \\
\arrayrulecolor{linegray}\hhline{------}

\quad Is intra-cranial hemorrhage (ICH) present? & 68.8 & 78.0 & 81.2 & 71.4 & \textbf{93.8} \\
\arrayrulecolor{linegray}\hhline{------}

\quad Is intra-cranial metastasis suspected based on the imaging findings? & 65.9 & 72.7 & 73.7 & 60.2 & \textbf{90.2} \\
\arrayrulecolor{linegray}\hhline{------}

\quad Is meningioma suspected based on the imaging findings? & 64.6 & 71.0 & 71.4 & 62.4 & \textbf{78.6} \\
\arrayrulecolor{linegray}\hhline{------}

\quad Is there any brain atrophy? & 82.6 & 79.5 & 80.8 & 72.9 & \textbf{85.5} \\
\arrayrulecolor{linegray}\hhline{------}

\quad Is there any deep brain stimulation (DBS) device present? & 84.2 & 97.2 & 96.9 & 94.3 & \textbf{99.9} \\
\arrayrulecolor{linegray}\hhline{------}

\quad Is there cerebral / cerebellar herniation? & 78.8 & 79.5 & 86.4 & 74.0 & \textbf{95.1} \\
\arrayrulecolor{black}\midrule
\rowcolor{rowgray}
\textbf{Device/Procedural findings} & & & & & \\
\rowcolor{rowgray}
\quad Are any post-craniectomy changes? & 69.1 & 90.4 & 89.5 & 77.5 & \textbf{94.7} \\
\arrayrulecolor{linegray}\hhline{------}
\rowcolor{rowgray}
\quad Are there any post-craniotomy changes? & 69.5 & 92.3 & 91.4 & 79.9 & \textbf{96.0} \\
\arrayrulecolor{linegray}\hhline{------}
\rowcolor{rowgray}
\quad Are there any post-radiotherapy changes? & 54.9 & 68.3 & 66.2 & 49.2 & \textbf{83.7} \\
\arrayrulecolor{linegray}\hhline{------}
\rowcolor{rowgray}
\quad Is there transtentorial herniation? & 79.9 & 82.5 & 89.4 & 75.1 & \textbf{96.2} \\
\arrayrulecolor{black}\midrule

\textbf{Mass/Tumor findings} & & & & & \\

\quad Is mass effect present? & 70.7 & 80.2 & 82.2 & 71.4 & \textbf{93.4} \\
\arrayrulecolor{black}\midrule
\rowcolor{rowgray}
\textbf{Other findings} & & & & & \\
\rowcolor{rowgray}
\quad Are there any osteosclerotic lesions? & 59.9 & 69.7 & \textbf{71.0} & 55.1 & 68.4 \\
\arrayrulecolor{linegray}\hhline{------}
\rowcolor{rowgray}
\quad Is acute infarct suspected based on the imaging findings? & 68.2 & 70.1 & 69.2 & 64.4 & \textbf{90.2} \\
\arrayrulecolor{linegray}\hhline{------}
\rowcolor{rowgray}
\quad Is chronic infarct suspected based on the imaging findings? & 73.4 & 72.0 & 72.5 & 65.9 & \textbf{83.8} \\
\arrayrulecolor{linegray}\hhline{------}
\rowcolor{rowgray}
\quad Is epidural hemorrhage (EDH) present? & 66.3 & 76.0 & 73.5 & 73.4 & \textbf{91.3} \\
\arrayrulecolor{linegray}\hhline{------}
\rowcolor{rowgray}
\quad Is external ventricular drainage (EVD) present? & 77.2 & 97.5 & 95.4 & 93.0 & \textbf{98.3} \\
\arrayrulecolor{linegray}\hhline{------}
\rowcolor{rowgray}
\quad Is hemorrhagic transformation of infarct present? & 74.7 & 77.0 & 82.4 & 63.2 & \textbf{94.3} \\
\arrayrulecolor{linegray}\hhline{------}
\rowcolor{rowgray}
\quad Is intra-parenchymal hemorrhage (IPH) present? & 68.1 & 78.9 & 82.4 & 69.9 & \textbf{95.2} \\
\arrayrulecolor{linegray}\hhline{------}
\rowcolor{rowgray}
\quad Is intra-ventricular hemorrhage (IVH) present? & 72.4 & 88.2 & 89.4 & 82.3 & \textbf{96.2} \\
\arrayrulecolor{linegray}\hhline{------}
\rowcolor{rowgray}
\quad Is subarachnoid hemorrhage (SAH) present? & 70.2 & 78.7 & 81.9 & 76.6 & \textbf{93.4} \\
\arrayrulecolor{linegray}\hhline{------}
\rowcolor{rowgray}
\quad Is subdural hemorrhage (SDH) present? & 62.0 & 71.5 & 75.1 & 63.0 & \textbf{92.6} \\
\arrayrulecolor{linegray}\hhline{------}
\rowcolor{rowgray}
\quad Is there any loss of the gray-white matter junction? & 60.2 & 66.5 & 67.0 & 58.2 & \textbf{85.1} \\
\arrayrulecolor{linegray}\hhline{------}
\rowcolor{rowgray}
\quad Is there midline shift? & 73.6 & 85.7 & 88.7 & 75.9 & \textbf{96.2} \\
\arrayrulecolor{linegray}\hhline{------}
\rowcolor{rowgray}
\quad Is there subfalcine herniation? & 73.5 & 71.3 & 81.1 & 67.6 & \textbf{95.6} \\
\end{longtable}
\end{center}

\vspace{2mm}
\noindent
\footnotesize \textit{Abbreviations:} AUC, area under the curve; NA, not applicable.

\newpage
\section{Performance on full set of RATE-Evals tasks on UCSF Breast MRI test set}
\label{ref:breast-finding}

\begin{center}
\footnotesize
\begin{longtable}{p{6cm} >{\centering\arraybackslash}p{2.3cm} >{\centering\arraybackslash}p{2.3cm} >{\centering\arraybackslash}p{2.3cm} >{\centering\arraybackslash}p{2.3cm}} 
\caption{Model Performance (AUC x 100) for UCSF Breast MRI} \label{tab:auc_ucsf_breast_mr} \\
\arrayrulecolor{black}
\toprule
& \multicolumn{4}{c}{\textbf{Model AUC (x 100)}} \\
\cmidrule(lr){2-5} 
\textbf{Finding} & \textbf{LingShu} & \textbf{MedGemma} & \textbf{MedImageInsight} & \textbf{\name} \\
\arrayrulecolor{black}
\midrule
\endfirsthead

\multicolumn{5}{c}{%
\tablename\ \thetable{} -- Model Performance (AUC x 100) for UCSF Breast MR \textit{(continued)}} \\
\arrayrulecolor{black}
\toprule
& \multicolumn{4}{c}{\textbf{Model AUC (x 100)}} \\
\cmidrule(lr){2-5} 
\textbf{Finding} & \textbf{LingShu} & \textbf{MedGemma} & \textbf{MedImageInsight} & \textbf{\name} \\
\arrayrulecolor{black}
\midrule
\endhead

\arrayrulecolor{black}
\midrule
\multicolumn{5}{r}{\textit{Continued on next page}} \\
\endfoot

\arrayrulecolor{black}
\bottomrule
\endlastfoot
\rowcolor{rowgray}
\textbf{Breast findings} & & & & \\
\rowcolor{rowgray}
\quad Are any foci of enhancement present in the breast? & 59.2 & 65.8 & 66.0 & \textbf{79.6} \\
\arrayrulecolor{linegray}\hhline{-----}
\rowcolor{rowgray}
\quad Are any masses present in the breast? & 56.6 & 65.4 & 64.9 & \textbf{80.9} \\
\arrayrulecolor{linegray}\hhline{-----}
\rowcolor{rowgray}
\quad Are any simple or complicated breast cysts present? & 58.2 & 73.5 & 76.1 & \textbf{77.8} \\
\arrayrulecolor{linegray}\hhline{-----}
\rowcolor{rowgray}
\quad Are there any masses in the breast contralateral to the known cancer? & 56.5 & 63.4 & 60.5 & \textbf{71.7} \\
\arrayrulecolor{linegray}\hhline{-----}
\rowcolor{rowgray}
\quad Is angiosarcoma in the breast suspected based on the imaging findings? & NA & NA & NA & NA \\
\arrayrulecolor{linegray}\hhline{-----}
\rowcolor{rowgray}
\quad Is breast abscess suspected based on the imaging findings? & NA & NA & NA & NA \\
\arrayrulecolor{linegray}\hhline{-----}
\rowcolor{rowgray}
\quad Is breast cancer recurrence suspected based on the imaging findings? & 58.1 & 60.6 & 61.9 & \textbf{72.6} \\
\arrayrulecolor{linegray}\hhline{-----}
\rowcolor{rowgray}
\quad Is carcinoma in the breast suspected based on the imaging findings? & 57.6 & 65.7 & 64.9 & \textbf{80.6} \\
\arrayrulecolor{linegray}\hhline{-----}
\rowcolor{rowgray}
\quad Is fat necrosis or oil cyst in the breast suspected based on the imaging findings? & 68.1 & 78.0 & 69.5 & \textbf{80.0} \\
\arrayrulecolor{linegray}\hhline{-----}
\rowcolor{rowgray}
\quad Is fibroadenoma in the breast suspected based on the imaging findings? & 51.5 & 61.5 & \textbf{72.0} & 67.9 \\
\arrayrulecolor{linegray}\hhline{-----}
\rowcolor{rowgray}
\quad Is hamartoma in the breast suspected based on the imaging findings? & 10.7 & 33.0 & 25.4 & \textbf{94.0} \\
\arrayrulecolor{linegray}\hhline{-----}
\rowcolor{rowgray}
\quad Is intraductal papilloma in the breast suspected based on the imaging findings? & 55.4 & 24.8 & 21.8 & \textbf{99.3} \\
\arrayrulecolor{linegray}\hhline{-----}
\rowcolor{rowgray}
\quad Is lipoma in the breast suspected based on the imaging findings? & 52.5 & \textbf{95.5} & 11.4 & 67.0 \\
\arrayrulecolor{linegray}\hhline{-----}
\rowcolor{rowgray}
\quad Is non-mass enhancement present in the breast? & 59.6 & 62.6 & 61.8 & \textbf{73.2} \\
\arrayrulecolor{linegray}\hhline{-----}
\rowcolor{rowgray}
\quad Is phyllodes tumor in the breast suspected based on the imaging findings? & 51.1 & 60.4 & 50.4 & \textbf{85.2} \\
\arrayrulecolor{linegray}\hhline{-----}
\rowcolor{rowgray}
\quad Is pseudoangiomatous stromal hyperplasia (PASH) in the breast suspected based on the imaging findings? & 72.6 & 36.8 & 59.3 & \textbf{80.7} \\
\arrayrulecolor{linegray}\hhline{-----}
\rowcolor{rowgray}
\quad Is there any internal mammary lymphadenopathy? & 55.5 & 64.4 & 63.5 & \textbf{77.4} \\
\arrayrulecolor{linegray}\hhline{-----}
\rowcolor{rowgray}
\quad Is tubular adenoma in the breast suspected based on the imaging findings? & NA & NA & NA & NA \\
\arrayrulecolor{black}\midrule

\textbf{Device/Procedural findings} & & & & \\

\quad Are any post-operative hematomas or seromas present? & 54.4 & 66.6 & 60.6 & \textbf{81.6} \\
\arrayrulecolor{linegray}\hhline{-----}

\quad Are any post-partial mastectomy (lumpectomy) changes present? & 60.0 & 75.6 & 71.3 & \textbf{88.4} \\
\arrayrulecolor{black}\midrule
\rowcolor{rowgray}
\textbf{Liver-related findings} & & & & \\
\rowcolor{rowgray}
\quad Are any liver cysts present? & 54.3 & 64.4 & 65.1 & \textbf{79.2} \\
\arrayrulecolor{linegray}\hhline{-----}
\rowcolor{rowgray}
\quad Are any liver masses present? & 57.8 & 58.6 & 63.4 & \textbf{81.3} \\
\arrayrulecolor{black}\midrule

\textbf{Mass/Tumor findings} & & & & \\

\quad Are any metastases present in the visible scan field? & 70.4 & 74.2 & 77.0 & \textbf{83.3} \\
\arrayrulecolor{linegray}\hhline{-----}

\quad Is the margin of the mass spiculated? & 60.9 & 70.1 & 70.7 & \textbf{88.0} \\
\arrayrulecolor{linegray}\hhline{-----}

\quad Is the shape of the mass irregular? & 60.2 & 67.5 & 67.8 & \textbf{84.8} \\
\arrayrulecolor{linegray}\hhline{-----}

\quad Is there any satellite mass around the main tumor? & 65.2 & 73.1 & 70.1 & \textbf{84.8} \\
\arrayrulecolor{linegray}\hhline{-----}

\quad Is there any skin thickening associated with the known cancer? & 58.7 & 68.9 & 64.1 & \textbf{88.8} \\
\arrayrulecolor{linegray}\hhline{-----}

\quad Is there evidence of tumor involvement of the pectoralis muscle? & 66.6 & 74.4 & 75.1 & \textbf{78.1} \\
\arrayrulecolor{linegray}\hhline{-----}

\quad Is there evidence of tumor involvement of the skin? & 49.3 & 63.2 & 63.4 & \textbf{85.5} \\
\arrayrulecolor{black}\midrule
\rowcolor{rowgray}
\textbf{Other findings} & & & & \\
\rowcolor{rowgray}
\quad Are any galactoceles present? & 53.2 & 27.6 & 29.0 & \textbf{77.9} \\
\arrayrulecolor{linegray}\hhline{-----}
\rowcolor{rowgray}
\quad Are any post radical mastectomy changes present? & 67.0 & 95.9 & 93.4 & \textbf{97.2} \\
\arrayrulecolor{linegray}\hhline{-----}
\rowcolor{rowgray}
\quad Are any post simple mastectomy changes present? & 66.6 & 93.8 & 91.0 & \textbf{94.0} \\
\arrayrulecolor{linegray}\hhline{-----}
\rowcolor{rowgray}
\quad Are there any biopsy clips? & 55.6 & 67.0 & 65.6 & \textbf{74.9} \\
\arrayrulecolor{linegray}\hhline{-----}
\rowcolor{rowgray}
\quad Is Paget's disease suspected based on the imaging findings? & NA & NA & NA & NA \\
\arrayrulecolor{linegray}\hhline{-----}
\rowcolor{rowgray}
\quad Is implant-associated anaplastic large-cell lymphoma (BIA-ALCL) suspected based on the imaging findings? & NA & NA & NA & NA \\
\arrayrulecolor{linegray}\hhline{-----}
\rowcolor{rowgray}
\quad Is mastitis suspected based on the imaging findings? & 22.7 & 93.3 & \textbf{97.1} & 85.7 \\
\arrayrulecolor{linegray}\hhline{-----}
\rowcolor{rowgray}
\quad Is there any axillary lymphadenopathy ipsilateral? & 66.6 & 73.9 & 73.6 & \textbf{83.2} \\
\arrayrulecolor{linegray}\hhline{-----}
\rowcolor{rowgray}
\quad Is there any nipple retraction or inversion? & 61.4 & 68.1 & 69.9 & \textbf{89.3} \\
\arrayrulecolor{linegray}\hhline{-----}
\rowcolor{rowgray}
\quad Is there evidence of capsular contracture around the implant? & NA & NA & NA & NA \\
\arrayrulecolor{linegray}\hhline{-----}
\rowcolor{rowgray}
\quad Is there evidence of extra-capsular implant rupture? & NA & NA & NA & NA \\
\arrayrulecolor{linegray}\hhline{-----}
\rowcolor{rowgray}
\quad Is there evidence of intra-capsular implant rupture? & 65.0 & 94.4 & 90.3 & \textbf{99.6} \\
\arrayrulecolor{linegray}\hhline{-----}
\rowcolor{rowgray}
\quad Is there fast initial enhancement and washout delayed enhancement? & 55.3 & 67.9 & 65.8 & \textbf{87.0} \\
\end{longtable}
\end{center}

\vspace{2mm}
\noindent
\footnotesize \textit{Abbreviations:} AUC, area under the curve; NA, not applicable.

\newpage
\section{Performance on full set of RATE-Evals tasks on Merlin Abdomen-Pelvis CT test set}

\begin{center}
\footnotesize
\begin{longtable}{p{5cm} >{\centering\arraybackslash}p{2.0cm} >{\centering\arraybackslash}p{2.0cm} >{\centering\arraybackslash}p{2.0cm} >{\centering\arraybackslash}p{2.0cm} >{\centering\arraybackslash}p{2.0cm}} 
\caption{Model Performance (AUC x 100) for Merlin} \label{tab:auc_merlin} \\
\arrayrulecolor{black}
\toprule
& \multicolumn{5}{c}{\textbf{Model AUC (x 100)}} \\
\cmidrule(lr){2-6} 
\textbf{Finding} & \textbf{Atlas} & \textbf{MedGemma} & \textbf{MedImageInsight} & \textbf{Merlin} & \textbf{\name} \\
\arrayrulecolor{black}
\midrule
\endfirsthead

\multicolumn{6}{c}{%
\tablename\ \thetable{} -- Model Performance (AUC x 100) for Merlin \textit{(continued)}} \\
\arrayrulecolor{black}
\toprule
& \multicolumn{5}{c}{\textbf{Model AUC (x 100)}} \\
\cmidrule(lr){2-6} 
\textbf{Finding} & \textbf{Atlas} & \textbf{MedGemma} & \textbf{MedImageInsight} & \textbf{Merlin} & \textbf{\name} \\
\arrayrulecolor{black}
\midrule
\endhead

\arrayrulecolor{black}
\midrule
\multicolumn{6}{r}{\textit{Continued on next page}} \\
\endfoot

\arrayrulecolor{black}
\bottomrule
\endlastfoot
\rowcolor{rowgray}
\textbf{Biliary findings} & & & & & \\
\rowcolor{rowgray}
\quad Are there any biliary cystadenomas? & 86.8 & 34.7 & 40.5 & 39.7 & \textbf{93.5} \\
\arrayrulecolor{linegray}\hhline{------}
\rowcolor{rowgray}
\quad Are there any biliary hamartomas? & 73.2 & 58.6 & 63.1 & 68.7 & \textbf{85.0} \\
\arrayrulecolor{linegray}\hhline{------}
\rowcolor{rowgray}
\quad Are there any biliary stents? & 94.3 & 89.0 & 83.7 & 98.5 & \textbf{98.6} \\
\arrayrulecolor{linegray}\hhline{------}
\rowcolor{rowgray}
\quad Are there any choledochal cysts? & 39.4 & 52.4 & \textbf{73.6} & 38.0 & 60.4 \\
\arrayrulecolor{linegray}\hhline{------}
\rowcolor{rowgray}
\quad Are there any gallbladder polyps? & 62.5 & \textbf{77.3} & 69.3 & 69.2 & 67.0 \\
\arrayrulecolor{linegray}\hhline{------}
\rowcolor{rowgray}
\quad Are there any gallbladder stones or cholelithiasis? & 73.2 & 65.8 & 66.9 & 74.2 & \textbf{87.5} \\
\arrayrulecolor{linegray}\hhline{------}
\rowcolor{rowgray}
\quad Is the gallbladder surgically absent? & 94.8 & 79.2 & 77.4 & 96.7 & \textbf{98.0} \\
\arrayrulecolor{linegray}\hhline{------}
\rowcolor{rowgray}
\quad Is there gallbladder adenomyomatosis? & 68.7 & 70.8 & 69.1 & 66.5 & \textbf{77.8} \\
\arrayrulecolor{linegray}\hhline{------}
\rowcolor{rowgray}
\quad Is there gallbladder cancer? & 98.4 & 53.7 & 63.1 & 91.6 & \textbf{98.6} \\
\arrayrulecolor{linegray}\hhline{------}
\rowcolor{rowgray}
\quad Is there gallbladder wall thickening? & 81.7 & 71.7 & 74.1 & 79.0 & \textbf{86.5} \\
\arrayrulecolor{linegray}\hhline{------}
\rowcolor{rowgray}
\quad Is there porcelain gallbladder? & 75.6 & 9.8 & 34.2 & 57.1 & \textbf{86.5} \\
\arrayrulecolor{black}\midrule

\textbf{Bone/Fracture findings} & & & & & \\

\quad Are there any femoral fractures? & 82.4 & 79.9 & 78.0 & 83.0 & \textbf{85.7} \\
\arrayrulecolor{linegray}\hhline{------}

\quad Are there any fractures (general)? & 75.7 & 74.1 & 74.6 & 74.6 & \textbf{78.0} \\
\arrayrulecolor{linegray}\hhline{------}

\quad Are there any osteolytic lesions? & 77.2 & 75.6 & 73.7 & 75.6 & \textbf{83.1} \\
\arrayrulecolor{linegray}\hhline{------}

\quad Are there any rib fractures? & 80.3 & 76.5 & 76.5 & 79.3 & \textbf{81.8} \\
\arrayrulecolor{linegray}\hhline{------}

\quad Are there any spinal fractures? & 76.3 & 73.9 & 76.5 & 73.3 & \textbf{80.8} \\
\arrayrulecolor{black}\midrule
\rowcolor{rowgray}
\textbf{Cardiac findings} & & & & & \\
\rowcolor{rowgray}
\quad Is there any coronary atherosclerosis? & 84.3 & 85.5 & 84.7 & 87.0 & \textbf{87.2} \\
\arrayrulecolor{linegray}\hhline{------}
\rowcolor{rowgray}
\quad Is there cardiomegaly? & 87.2 & 82.3 & 82.3 & \textbf{87.7} & 87.4 \\
\arrayrulecolor{black}\midrule

\textbf{Chest/Lung findings} & & & & & \\

\quad Are there any lung masses or nodules? & 65.9 & 63.0 & 62.6 & 63.5 & \textbf{66.6} \\
\arrayrulecolor{linegray}\hhline{------}

\quad Is there any atelectasis in the visible lung? & \textbf{77.9} & 73.9 & 74.4 & 76.1 & 77.5 \\
\arrayrulecolor{linegray}\hhline{------}

\quad Is there any free air or pneumoperitoneum? & 87.6 & 83.0 & 84.2 & 87.3 & \textbf{88.6} \\
\arrayrulecolor{linegray}\hhline{------}

\quad Is there pneumobilia? & 92.1 & 80.6 & 79.7 & 95.4 & \textbf{97.4} \\
\arrayrulecolor{black}\midrule
\rowcolor{rowgray}
\textbf{Cyst findings} & & & & & \\
\rowcolor{rowgray}
\quad Are there any Bartholin gland cysts? & 82.3 & 74.7 & 77.5 & 85.3 & \textbf{85.7} \\
\arrayrulecolor{linegray}\hhline{------}
\rowcolor{rowgray}
\quad Are there any Gartner duct cysts? & 87.7 & 66.8 & 75.0 & 82.1 & \textbf{92.0} \\
\arrayrulecolor{linegray}\hhline{------}
\rowcolor{rowgray}
\quad Are there any Nabothian cysts? & 87.3 & 83.8 & 81.6 & 84.9 & \textbf{87.7} \\
\arrayrulecolor{linegray}\hhline{------}
\rowcolor{rowgray}
\quad Are there any enteric duplication cysts? & NA & NA & NA & NA & NA \\
\arrayrulecolor{linegray}\hhline{------}
\rowcolor{rowgray}
\quad Are there any epithelial cysts? & 62.9 & 56.4 & 58.1 & 58.9 & \textbf{68.6} \\
\arrayrulecolor{linegray}\hhline{------}
\rowcolor{rowgray}
\quad Are there any mucinous cystic neoplasms? & 72.8 & 70.4 & 71.0 & 70.1 & \textbf{73.3} \\
\arrayrulecolor{linegray}\hhline{------}
\rowcolor{rowgray}
\quad Are there any serous cystic neoplasms? & 73.6 & 52.6 & 52.2 & \textbf{78.4} & 62.9 \\
\arrayrulecolor{linegray}\hhline{------}
\rowcolor{rowgray}
\quad Is there acute cholecystitis with rupture? & 87.3 & 80.7 & 80.2 & 73.8 & \textbf{92.9} \\
\arrayrulecolor{linegray}\hhline{------}
\rowcolor{rowgray}
\quad Is there acute cholecystitis? & 95.5 & 76.4 & 77.9 & 84.9 & \textbf{96.0} \\
\arrayrulecolor{linegray}\hhline{------}
\rowcolor{rowgray}
\quad Is there emphysematous cholecystitis? & 85.7 & 64.4 & 63.5 & 82.0 & \textbf{95.4} \\
\arrayrulecolor{black}\midrule

\textbf{Device/Procedural findings} & & & & & \\

\quad Are there any Foley catheters? & 93.3 & 90.0 & 89.0 & 94.4 & \textbf{96.3} \\
\arrayrulecolor{linegray}\hhline{------}

\quad Are there any central venous catheters? & 86.0 & 84.9 & 85.1 & 87.2 & \textbf{91.4} \\
\arrayrulecolor{linegray}\hhline{------}

\quad Are there any drainage catheters? & 92.7 & 89.8 & 88.5 & 93.2 & \textbf{95.4} \\
\arrayrulecolor{linegray}\hhline{------}

\quad Are there any extracorporeal membrane oxygenation (ECMO) devices? & 95.9 & \textbf{100.0} & 98.9 & 99.8 & 99.1 \\
\arrayrulecolor{linegray}\hhline{------}

\quad Are there any gastrostomy tubes? & 92.7 & 88.7 & 89.0 & 93.4 & \textbf{95.3} \\
\arrayrulecolor{linegray}\hhline{------}

\quad Are there any post-nephrectomy changes? & 81.1 & 76.2 & 69.0 & 77.1 & \textbf{91.5} \\
\arrayrulecolor{black}\midrule
\rowcolor{rowgray}
\textbf{Gastrointestinal findings} & & & & & \\
\rowcolor{rowgray}
\quad Are there any colonic carcinomas? & 76.1 & 65.4 & 75.3 & \textbf{82.3} & 78.3 \\
\arrayrulecolor{linegray}\hhline{------}
\rowcolor{rowgray}
\quad Are there any duodenal carcinomas? & \textbf{94.1} & 81.4 & 78.7 & 85.1 & 87.1 \\
\arrayrulecolor{linegray}\hhline{------}
\rowcolor{rowgray}
\quad Are there any esophageal carcinomas? & 92.2 & 61.8 & 69.2 & 94.2 & \textbf{99.1} \\
\arrayrulecolor{linegray}\hhline{------}
\rowcolor{rowgray}
\quad Are there any esophageal or gastric varices? & 92.6 & 85.4 & 88.8 & \textbf{95.2} & 94.8 \\
\arrayrulecolor{linegray}\hhline{------}
\rowcolor{rowgray}
\quad Are there any gastric carcinomas? & \textbf{88.9} & 77.5 & 79.5 & 88.2 & 88.0 \\
\arrayrulecolor{linegray}\hhline{------}
\rowcolor{rowgray}
\quad Are there any gastric foreign bodies (not related to medical devices or post-surgical changes)? & \textbf{72.7} & 38.2 & 49.6 & 66.7 & 38.3 \\
\arrayrulecolor{linegray}\hhline{------}
\rowcolor{rowgray}
\quad Are there any gastrointestinal stromal tumors (GISTs)? & 35.9 & 40.7 & 32.7 & 41.5 & \textbf{49.3} \\
\arrayrulecolor{linegray}\hhline{------}
\rowcolor{rowgray}
\quad Are there any nasogastric tubes? & 91.3 & 93.7 & 89.0 & 95.2 & \textbf{97.1} \\
\arrayrulecolor{linegray}\hhline{------}
\rowcolor{rowgray}
\quad Is bowel obstruction present? & 93.8 & 86.7 & 89.8 & 94.2 & \textbf{94.8} \\
\arrayrulecolor{linegray}\hhline{------}
\rowcolor{rowgray}
\quad Is there a surgical gastric conduit? & 87.2 & 72.8 & 74.2 & 88.1 & \textbf{91.9} \\
\arrayrulecolor{linegray}\hhline{------}
\rowcolor{rowgray}
\quad Is there achalasia or scleroderma suspected based on gastrointestinal imaging findings? & 75.0 & 47.0 & 63.6 & 86.1 & \textbf{90.8} \\
\arrayrulecolor{linegray}\hhline{------}
\rowcolor{rowgray}
\quad Is there bowel obstruction? & 93.7 & 86.5 & 89.6 & 94.2 & \textbf{94.7} \\
\arrayrulecolor{linegray}\hhline{------}
\rowcolor{rowgray}
\quad Is there gastric volvulus? & \textbf{99.6} & 93.0 & 90.3 & 95.6 & 99.3 \\
\arrayrulecolor{linegray}\hhline{------}
\rowcolor{rowgray}
\quad Is there gastrointestinal lymphoma? & 63.9 & 64.7 & 49.4 & 55.0 & \textbf{75.2} \\
\arrayrulecolor{linegray}\hhline{------}
\rowcolor{rowgray}
\quad Is there inflammatory bowel disease? & 80.9 & 78.9 & 79.6 & 79.1 & \textbf{83.8} \\
\arrayrulecolor{linegray}\hhline{------}
\rowcolor{rowgray}
\quad Is there large bowel obstruction? & 89.9 & 90.5 & 88.9 & 92.7 & \textbf{93.2} \\
\arrayrulecolor{linegray}\hhline{------}
\rowcolor{rowgray}
\quad Is there pneumatosis intestinalis? & \textbf{80.1} & 77.2 & 78.1 & 79.3 & 80.0 \\
\arrayrulecolor{linegray}\hhline{------}
\rowcolor{rowgray}
\quad Is there small bowel obstruction? & 94.8 & 85.7 & 90.0 & 94.7 & \textbf{95.4} \\
\arrayrulecolor{black}\midrule

\textbf{Kidney-related findings} & & & & & \\

\quad Are there any adrenal adenomas? & 71.0 & \textbf{73.5} & 73.0 & 71.4 & 70.6 \\
\arrayrulecolor{linegray}\hhline{------}

\quad Are there any adrenal hemorrhages? & 85.6 & 71.8 & 79.2 & \textbf{87.5} & 86.5 \\
\arrayrulecolor{linegray}\hhline{------}

\quad Are there any adrenal infarcts? & NA & NA & NA & NA & NA \\
\arrayrulecolor{linegray}\hhline{------}

\quad Are there any adrenal masses? & \textbf{68.6} & 67.8 & 68.4 & 66.6 & 67.9 \\
\arrayrulecolor{linegray}\hhline{------}

\quad Are there any adrenal myelolipomas? & \textbf{81.4} & 78.5 & 79.0 & 74.8 & 72.8 \\
\arrayrulecolor{linegray}\hhline{------}

\quad Are there any complex renal cysts? & 73.2 & 65.0 & 70.1 & 65.5 & \textbf{80.3} \\
\arrayrulecolor{linegray}\hhline{------}

\quad Are there any percutaneous nephrostomy tubes/catheters? & 95.7 & 89.8 & 87.9 & 92.9 & \textbf{98.9} \\
\arrayrulecolor{linegray}\hhline{------}

\quad Are there any post-renal transplantation changes? & 92.9 & 89.4 & 87.0 & 94.7 & \textbf{98.0} \\
\arrayrulecolor{linegray}\hhline{------}

\quad Are there any renal abscesses? & 94.0 & 48.1 & 28.8 & 81.8 & \textbf{97.8} \\
\arrayrulecolor{linegray}\hhline{------}

\quad Are there any renal cell carcinomas? & 72.8 & 80.7 & 84.8 & 77.4 & \textbf{86.5} \\
\arrayrulecolor{linegray}\hhline{------}

\quad Are there any renal hypodensities? & 66.3 & 66.3 & 66.3 & 66.2 & \textbf{79.2} \\
\arrayrulecolor{linegray}\hhline{------}

\quad Are there any renal infarcts? & 57.1 & \textbf{73.4} & 72.8 & 70.7 & 68.7 \\
\arrayrulecolor{linegray}\hhline{------}

\quad Are there any renal lacerations? & 91.7 & 89.7 & 81.7 & 95.2 & \textbf{98.9} \\
\arrayrulecolor{linegray}\hhline{------}

\quad Are there any renal stones / nephrolithiasis? & 65.4 & 61.5 & 60.5 & 62.0 & \textbf{70.4} \\
\arrayrulecolor{linegray}\hhline{------}

\quad Are there any simple renal cysts? & 70.1 & 68.6 & 69.0 & 68.1 & \textbf{81.1} \\
\arrayrulecolor{linegray}\hhline{------}

\quad Are there any ureteral double-J catheters/stents? & 93.6 & 96.6 & 94.2 & 95.9 & \textbf{98.0} \\
\arrayrulecolor{linegray}\hhline{------}

\quad Are there any ureteral stones? & 87.2 & 71.7 & 74.9 & 78.7 & \textbf{89.8} \\
\arrayrulecolor{linegray}\hhline{------}

\quad Is there adrenal hyperplasia? & 55.5 & 50.6 & 60.9 & \textbf{74.7} & 69.3 \\
\arrayrulecolor{linegray}\hhline{------}

\quad Is there horseshoe kidney? & 88.9 & 54.5 & 76.4 & 60.6 & \textbf{93.0} \\
\arrayrulecolor{linegray}\hhline{------}

\quad Is there hydronephrosis? & 93.0 & 74.9 & 76.0 & 79.9 & \textbf{94.7} \\
\arrayrulecolor{linegray}\hhline{------}

\quad Is there nephrocalcinosis? & \textbf{72.6} & 59.8 & 64.6 & 67.4 & 72.3 \\
\arrayrulecolor{linegray}\hhline{------}

\quad Is there polycystic kidney disease? & 95.7 & 89.6 & 92.9 & 95.7 & \textbf{99.3} \\
\arrayrulecolor{linegray}\hhline{------}

\quad Is there renal lymphomas? & NA & NA & NA & NA & NA \\
\arrayrulecolor{black}\midrule
\rowcolor{rowgray}
\textbf{Liver-related findings} & & & & & \\
\rowcolor{rowgray}
\quad Are there any hepatic abscesses? & 90.6 & 80.0 & 79.1 & 88.5 & \textbf{92.6} \\
\arrayrulecolor{linegray}\hhline{------}
\rowcolor{rowgray}
\quad Are there any hepatic adenomas? & 1.1 & 71.8 & 84.2 & \textbf{89.3} & 59.6 \\
\arrayrulecolor{linegray}\hhline{------}
\rowcolor{rowgray}
\quad Are there any hepatic changes post-resection? & 85.8 & 75.3 & 74.3 & 87.9 & \textbf{89.4} \\
\arrayrulecolor{linegray}\hhline{------}
\rowcolor{rowgray}
\quad Are there any hepatic changes post-transplantation? & 88.5 & 75.3 & 73.0 & 91.7 & \textbf{93.9} \\
\arrayrulecolor{linegray}\hhline{------}
\rowcolor{rowgray}
\quad Are there any hepatic cysts? & 75.4 & 64.1 & 65.2 & 66.1 & \textbf{88.5} \\
\arrayrulecolor{linegray}\hhline{------}
\rowcolor{rowgray}
\quad Are there any hepatic focal nodular hyperplasias? & 58.9 & 63.6 & 67.6 & \textbf{70.3} & 65.5 \\
\arrayrulecolor{linegray}\hhline{------}
\rowcolor{rowgray}
\quad Are there any hepatic hemangiomas? & 68.8 & 62.8 & 61.6 & 60.0 & \textbf{83.7} \\
\arrayrulecolor{linegray}\hhline{------}
\rowcolor{rowgray}
\quad Are there any hepatic infarcts? & 96.5 & 81.3 & 95.4 & \textbf{99.0} & 97.7 \\
\arrayrulecolor{linegray}\hhline{------}
\rowcolor{rowgray}
\quad Are there any hepatic masses? & 80.6 & 67.5 & 68.4 & 75.3 & \textbf{89.5} \\
\arrayrulecolor{linegray}\hhline{------}
\rowcolor{rowgray}
\quad Are there any hepatic metastases? & 92.7 & 81.2 & 81.9 & 90.0 & \textbf{96.6} \\
\arrayrulecolor{linegray}\hhline{------}
\rowcolor{rowgray}
\quad Are there any hepatic regenerative nodules? & 92.8 & 70.6 & 88.9 & 90.8 & \textbf{97.3} \\
\arrayrulecolor{linegray}\hhline{------}
\rowcolor{rowgray}
\quad Are there any intra-hepatic stones? & 67.7 & 65.2 & 67.9 & 72.5 & \textbf{76.2} \\
\arrayrulecolor{linegray}\hhline{------}
\rowcolor{rowgray}
\quad Are there any liver lacerations? & 82.9 & 82.9 & 82.6 & 79.0 & \textbf{84.8} \\
\arrayrulecolor{linegray}\hhline{------}
\rowcolor{rowgray}
\quad Are there any sites of active extravasation in the liver? & 89.1 & 51.4 & 24.5 & 92.5 & \textbf{96.7} \\
\arrayrulecolor{linegray}\hhline{------}
\rowcolor{rowgray}
\quad Are there any transjugular intrahepatic portosystemic shunts? & 96.8 & 91.7 & 91.6 & 95.7 & \textbf{99.3} \\
\arrayrulecolor{linegray}\hhline{------}
\rowcolor{rowgray}
\quad Is hepatic steatosis, fatty liver, focal fatty infiltration suspected based on the imaging findings? & 81.5 & 74.6 & 73.6 & 78.9 & \textbf{83.1} \\
\arrayrulecolor{linegray}\hhline{------}
\rowcolor{rowgray}
\quad Is polycystic liver disease present? & 86.9 & 90.0 & 97.5 & 87.9 & \textbf{98.7} \\
\arrayrulecolor{linegray}\hhline{------}
\rowcolor{rowgray}
\quad Is there any intrahepatic biliary ductal dilation? & 87.9 & 73.2 & 73.3 & 87.9 & \textbf{91.6} \\
\arrayrulecolor{linegray}\hhline{------}
\rowcolor{rowgray}
\quad Is there any portal venous gas? & 92.6 & \textbf{92.8} & 92.4 & 87.9 & 86.6 \\
\arrayrulecolor{linegray}\hhline{------}
\rowcolor{rowgray}
\quad Is there cirrhosis with decompensation? & 97.7 & 91.5 & 92.2 & 97.9 & \textbf{98.6} \\
\arrayrulecolor{linegray}\hhline{------}
\rowcolor{rowgray}
\quad Is there cirrhosis with portal hypertension? & 96.7 & 90.9 & 91.9 & 97.7 & \textbf{98.3} \\
\arrayrulecolor{linegray}\hhline{------}
\rowcolor{rowgray}
\quad Is there cirrhosis? & 94.5 & 87.6 & 88.8 & 94.6 & \textbf{96.1} \\
\arrayrulecolor{linegray}\hhline{------}
\rowcolor{rowgray}
\quad Is there hepatic lymphoma? & 62.5 & 67.9 & 74.0 & 63.8 & \textbf{75.4} \\
\arrayrulecolor{linegray}\hhline{------}
\rowcolor{rowgray}
\quad Is there hepatoblastoma? & NA & NA & NA & NA & NA \\
\arrayrulecolor{linegray}\hhline{------}
\rowcolor{rowgray}
\quad Is there hepatocellular carcinoma? & 96.7 & 78.5 & 84.7 & 94.4 & \textbf{98.4} \\
\arrayrulecolor{linegray}\hhline{------}
\rowcolor{rowgray}
\quad Is there hepatomegaly? & 83.9 & 79.7 & 83.5 & 80.7 & \textbf{84.8} \\
\arrayrulecolor{linegray}\hhline{------}
\rowcolor{rowgray}
\quad Is there portal venous occlusion? & 88.8 & 79.1 & 81.6 & \textbf{90.5} & 89.6 \\
\arrayrulecolor{linegray}\hhline{------}
\rowcolor{rowgray}
\quad Is there viral hepatitis? & 31.1 & \textbf{62.8} & 24.5 & 43.6 & 26.9 \\
\arrayrulecolor{linegray}\hhline{------}
\rowcolor{rowgray}
\quad is there any extrahepatic biliary ductal dilation? & 86.4 & 71.3 & 71.3 & 83.9 & \textbf{90.2} \\
\arrayrulecolor{black}\midrule

\textbf{Mass/Tumor findings} & & & & & \\

\quad Are there any Wilms tumors? & NA & NA & NA & NA & NA \\
\arrayrulecolor{linegray}\hhline{------}

\quad Are there any bladder transitional cell carcinomas? & 95.7 & 83.9 & 80.0 & 87.6 & \textbf{99.3} \\
\arrayrulecolor{linegray}\hhline{------}

\quad Are there any neuroendocrine tumors? & \textbf{77.3} & 54.6 & 55.5 & 62.3 & 72.3 \\
\arrayrulecolor{linegray}\hhline{------}

\quad Are there any rectal carcinomas? & 90.4 & 91.0 & 79.1 & 87.2 & \textbf{94.3} \\
\arrayrulecolor{linegray}\hhline{------}

\quad Are there any solid pseudopapillary tumors? & NA & NA & NA & NA & NA \\
\arrayrulecolor{linegray}\hhline{------}

\quad Are there any testicular masses? & NA & NA & NA & NA & NA \\
\arrayrulecolor{linegray}\hhline{------}

\quad Are there any vaginal or vulvar cancers? & NA & NA & NA & NA & NA \\
\arrayrulecolor{linegray}\hhline{------}

\quad Is there adenoma malignum? & 61.3 & 56.8 & \textbf{76.1} & 61.5 & 57.2 \\
\arrayrulecolor{linegray}\hhline{------}

\quad Is there cholangiocarcinoma? & 95.9 & 88.9 & 84.9 & 97.8 & \textbf{98.3} \\
\arrayrulecolor{linegray}\hhline{------}

\quad Is there fibrolamellar carcinoma? & NA & NA & NA & NA & NA \\
\arrayrulecolor{linegray}\hhline{------}

\quad Is there mesenteric carcinoid tumor? & 48.4 & \textbf{67.4} & 58.9 & 48.6 & 60.7 \\
\arrayrulecolor{linegray}\hhline{------}

\quad Is there periductal/intraductal cholangiocarcinoma? & 96.2 & 86.5 & 84.7 & 98.5 & \textbf{99.0} \\
\arrayrulecolor{linegray}\hhline{------}

\quad Is there peritoneal carcinomatosis or omental caking? & 92.9 & 88.5 & 87.9 & 92.3 & \textbf{95.2} \\
\arrayrulecolor{linegray}\hhline{------}

\quad Is there prostate cancer? & 95.0 & 62.7 & 67.5 & 94.9 & \textbf{95.9} \\
\arrayrulecolor{black}\midrule
\rowcolor{rowgray}
\textbf{Other findings} & & & & & \\
\rowcolor{rowgray}
\quad Are there any angiomyolipomas? & 65.5 & 64.4 & 66.4 & 64.9 & \textbf{66.9} \\
\arrayrulecolor{linegray}\hhline{------}
\rowcolor{rowgray}
\quad Are there any bladder stones? & \textbf{79.1} & 64.5 & 69.6 & 79.0 & 77.6 \\
\arrayrulecolor{linegray}\hhline{------}
\rowcolor{rowgray}
\quad Are there any dysplastic nodules? & NA & NA & NA & NA & NA \\
\arrayrulecolor{linegray}\hhline{------}
\rowcolor{rowgray}
\quad Are there any enteric diverticula? & 80.9 & 77.9 & 76.7 & 79.5 & \textbf{81.4} \\
\arrayrulecolor{linegray}\hhline{------}
\rowcolor{rowgray}
\quad Are there any femoral/ventral/Spigelian hernias? & 77.9 & 73.1 & 73.2 & 76.6 & \textbf{80.6} \\
\arrayrulecolor{linegray}\hhline{------}
\rowcolor{rowgray}
\quad Are there any hydroceles? & \textbf{85.0} & 84.9 & 79.7 & 79.4 & 81.2 \\
\arrayrulecolor{linegray}\hhline{------}
\rowcolor{rowgray}
\quad Are there any iliopsoas muscle abscesses? & 90.7 & 81.2 & 83.0 & 88.5 & \textbf{91.7} \\
\arrayrulecolor{linegray}\hhline{------}
\rowcolor{rowgray}
\quad Are there any inguinal hernias? & 81.1 & 80.2 & 79.5 & \textbf{82.3} & 80.7 \\
\arrayrulecolor{linegray}\hhline{------}
\rowcolor{rowgray}
\quad Are there any intraductal papillary mucinous neoplasms (IPMN)? & 74.4 & \textbf{74.8} & 74.2 & 73.8 & 74.7 \\
\arrayrulecolor{linegray}\hhline{------}
\rowcolor{rowgray}
\quad Are there any intraductal papillary mucinous neoplasms with worrisome features? & 58.7 & 69.9 & \textbf{76.4} & 51.1 & 70.4 \\
\arrayrulecolor{linegray}\hhline{------}
\rowcolor{rowgray}
\quad Are there any leiomyomas? & 83.6 & 73.6 & 74.2 & 82.6 & \textbf{88.5} \\
\arrayrulecolor{linegray}\hhline{------}
\rowcolor{rowgray}
\quad Are there any leiomyosarcomas? & NA & NA & NA & NA & NA \\
\arrayrulecolor{linegray}\hhline{------}
\rowcolor{rowgray}
\quad Are there any oncocytomas? & NA & NA & NA & NA & NA \\
\arrayrulecolor{linegray}\hhline{------}
\rowcolor{rowgray}
\quad Are there any osteosclerotic lesions? & 58.3 & 59.9 & 59.0 & 58.1 & \textbf{63.4} \\
\arrayrulecolor{linegray}\hhline{------}
\rowcolor{rowgray}
\quad Are there any pheochromocytomas? & NA & NA & NA & NA & NA \\
\arrayrulecolor{linegray}\hhline{------}
\rowcolor{rowgray}
\quad Are there any scrotal hematomas? & 97.9 & 82.3 & 75.2 & 96.7 & \textbf{99.0} \\
\arrayrulecolor{linegray}\hhline{------}
\rowcolor{rowgray}
\quad Are there any subcapsular hemorrhages? & 73.6 & 52.6 & 54.3 & 73.4 & \textbf{81.1} \\
\arrayrulecolor{linegray}\hhline{------}
\rowcolor{rowgray}
\quad Are there any urachal diverticula? & \textbf{81.7} & 69.5 & 71.4 & 74.7 & 74.0 \\
\arrayrulecolor{linegray}\hhline{------}
\rowcolor{rowgray}
\quad Are there any ureteroceles? & 50.4 & 66.8 & 45.8 & \textbf{67.8} & 58.6 \\
\arrayrulecolor{linegray}\hhline{------}
\rowcolor{rowgray}
\quad Are there any varicoceles? & 62.5 & 68.2 & 71.3 & 78.7 & \textbf{84.6} \\
\arrayrulecolor{linegray}\hhline{------}
\rowcolor{rowgray}
\quad Are there changes status post hysterectomy? & 88.1 & 78.5 & 77.9 & 85.8 & \textbf{92.2} \\
\arrayrulecolor{linegray}\hhline{------}
\rowcolor{rowgray}
\quad Is anasarca suspected based on the imaging findings? & \textbf{96.1} & 94.0 & 93.6 & 95.1 & 95.1 \\
\arrayrulecolor{linegray}\hhline{------}
\rowcolor{rowgray}
\quad Is appendicitis suspected based on the imaging findings? & 88.9 & 78.1 & 76.1 & 85.4 & \textbf{91.9} \\
\arrayrulecolor{linegray}\hhline{------}
\rowcolor{rowgray}
\quad Is hiatal hernia present? & 79.2 & 77.0 & 75.6 & 77.6 & \textbf{81.7} \\
\arrayrulecolor{linegray}\hhline{------}
\rowcolor{rowgray}
\quad Is metastatic disease suspected based on the imaging findings? & 86.4 & 79.5 & 79.2 & 84.7 & \textbf{89.8} \\
\arrayrulecolor{linegray}\hhline{------}
\rowcolor{rowgray}
\quad Is osteopenia suspected based on the imaging findings? & 87.8 & \textbf{90.3} & 88.7 & 87.5 & 88.0 \\
\arrayrulecolor{linegray}\hhline{------}
\rowcolor{rowgray}
\quad Is there Fournier's gangrene / necrotizing fasciitis? & NA & NA & NA & NA & NA \\
\arrayrulecolor{linegray}\hhline{------}
\rowcolor{rowgray}
\quad Is there adnexal torsion? & 90.2 & 80.5 & 88.2 & 89.1 & \textbf{93.8} \\
\arrayrulecolor{linegray}\hhline{------}
\rowcolor{rowgray}
\quad Is there angiosarcoma? & \textbf{62.0} & 37.3 & 50.2 & 60.2 & 46.9 \\
\arrayrulecolor{linegray}\hhline{------}
\rowcolor{rowgray}
\quad Is there any adenomyosis? & \textbf{85.9} & 73.9 & 62.9 & 76.1 & 77.9 \\
\arrayrulecolor{linegray}\hhline{------}
\rowcolor{rowgray}
\quad Is there any ascites? & 86.9 & 81.4 & 82.7 & 85.7 & \textbf{87.7} \\
\arrayrulecolor{linegray}\hhline{------}
\rowcolor{rowgray}
\quad Is there any consolidation? & 87.6 & 84.3 & 84.6 & 87.1 & \textbf{87.8} \\
\arrayrulecolor{linegray}\hhline{------}
\rowcolor{rowgray}
\quad Is there any evidence of enteric contrast leak suggestive of postoperative leakage/fistula after surgery? & 87.9 & 82.0 & 84.7 & 88.1 & \textbf{90.2} \\
\arrayrulecolor{linegray}\hhline{------}
\rowcolor{rowgray}
\quad Is there any pleural effusion? & 97.6 & 91.8 & 91.7 & 97.5 & \textbf{97.8} \\
\arrayrulecolor{linegray}\hhline{------}
\rowcolor{rowgray}
\quad Is there ascending cholangitis? & 84.9 & 53.4 & 56.1 & 85.8 & \textbf{89.1} \\
\arrayrulecolor{linegray}\hhline{------}
\rowcolor{rowgray}
\quad Is there bladder rupture? & 88.0 & 16.2 & 41.8 & 56.1 & \textbf{99.1} \\
\arrayrulecolor{linegray}\hhline{------}
\rowcolor{rowgray}
\quad Is there diverticulitis? & 92.8 & 76.9 & 77.6 & 85.2 & \textbf{92.9} \\
\arrayrulecolor{linegray}\hhline{------}
\rowcolor{rowgray}
\quad Is there epididymitis? & 93.9 & 89.8 & 91.6 & \textbf{99.2} & 97.4 \\
\arrayrulecolor{linegray}\hhline{------}
\rowcolor{rowgray}
\quad Is there epithelioid hemangioendothelioma? & NA & NA & NA & NA & NA \\
\arrayrulecolor{linegray}\hhline{------}
\rowcolor{rowgray}
\quad Is there evidence of sleeve gastrectomy or fundoplication? & 85.5 & 78.5 & 76.3 & 86.2 & \textbf{89.9} \\
\arrayrulecolor{linegray}\hhline{------}
\rowcolor{rowgray}
\quad Is there hemochromatosis? & NA & NA & NA & NA & NA \\
\arrayrulecolor{linegray}\hhline{------}
\rowcolor{rowgray}
\quad Is there hiatal hernia? & 79.3 & 77.0 & 75.5 & 77.5 & \textbf{81.7} \\
\arrayrulecolor{linegray}\hhline{------}
\rowcolor{rowgray}
\quad Is there ileus? & \textbf{88.7} & 85.4 & 86.5 & 87.7 & 88.2 \\
\arrayrulecolor{linegray}\hhline{------}
\rowcolor{rowgray}
\quad Is there intussusception? & 66.4 & 62.4 & 62.8 & \textbf{70.5} & 70.2 \\
\arrayrulecolor{linegray}\hhline{------}
\rowcolor{rowgray}
\quad Is there nephritis? & 85.8 & 76.7 & 76.9 & 83.0 & \textbf{87.4} \\
\arrayrulecolor{linegray}\hhline{------}
\rowcolor{rowgray}
\quad Is there primary sclerosing cholangitis? & 95.8 & 47.8 & 69.5 & 89.7 & \textbf{99.2} \\
\arrayrulecolor{linegray}\hhline{------}
\rowcolor{rowgray}
\quad Is there prostatitis? & NA & NA & NA & NA & NA \\
\arrayrulecolor{linegray}\hhline{------}
\rowcolor{rowgray}
\quad Is there prostatomegaly or prostatic hyperplasia? & 92.8 & 88.6 & 88.7 & 92.3 & \textbf{93.9} \\
\arrayrulecolor{linegray}\hhline{------}
\rowcolor{rowgray}
\quad Is there pseudomembranous colitis? & 79.0 & 80.2 & 80.8 & 86.8 & \textbf{87.6} \\
\arrayrulecolor{linegray}\hhline{------}
\rowcolor{rowgray}
\quad Is there pseudomyxoma peritonei? & 84.6 & 76.7 & 82.1 & 80.5 & \textbf{87.0} \\
\arrayrulecolor{linegray}\hhline{------}
\rowcolor{rowgray}
\quad Is there retroperitoneal fibrosis? & 80.0 & 43.0 & 58.6 & 85.3 & \textbf{89.9} \\
\arrayrulecolor{linegray}\hhline{------}
\rowcolor{rowgray}
\quad Is there retroperitoneal hemorrhage or retroperitoneal hematomas? & 85.8 & 82.5 & 82.2 & 82.4 & \textbf{87.8} \\
\arrayrulecolor{linegray}\hhline{------}
\rowcolor{rowgray}
\quad Is there retroperitoneal liposarcoma? & 64.5 & \textbf{88.1} & 77.2 & 76.7 & 62.4 \\
\arrayrulecolor{linegray}\hhline{------}
\rowcolor{rowgray}
\quad Is there submucosal edema, enterocolitis, or gastritis? & 75.3 & 71.6 & 73.4 & 73.5 & \textbf{77.1} \\
\arrayrulecolor{linegray}\hhline{------}
\rowcolor{rowgray}
\quad Is there testicular (segmental) infarct? & NA & NA & NA & NA & NA \\
\arrayrulecolor{linegray}\hhline{------}
\rowcolor{rowgray}
\quad Is there testicular torsion? & NA & NA & NA & NA & NA \\
\arrayrulecolor{linegray}\hhline{------}
\rowcolor{rowgray}
\quad Is there volvulus? & 89.9 & 80.4 & 84.0 & 86.4 & \textbf{90.2} \\
\arrayrulecolor{linegray}\hhline{------}
\rowcolor{rowgray}
\quad Is there xanthogranulomatous pyelonephritis? & 25.8 & 1.4 & \textbf{98.9} & 92.7 & 81.9 \\
\arrayrulecolor{black}\midrule

\textbf{Pancreas-related findings} & & & & & \\

\quad Are there any ductal pancreatic carcinomas? & 96.8 & 72.9 & 82.3 & 95.9 & \textbf{97.5} \\
\arrayrulecolor{linegray}\hhline{------}

\quad Are there any pancreatic tumors? & 75.6 & 72.7 & 73.7 & 76.6 & \textbf{78.1} \\
\arrayrulecolor{linegray}\hhline{------}

\quad Is chronic pancreatitis suspected based on imaging findings? & 74.6 & 70.5 & 69.0 & 73.4 & \textbf{78.0} \\
\arrayrulecolor{linegray}\hhline{------}

\quad Is cystic fibrosis suspected based on the pancreatic imaging findings? & 96.6 & 80.0 & 81.1 & 94.2 & \textbf{97.4} \\
\arrayrulecolor{linegray}\hhline{------}

\quad Is there acute pancreatitis? & 91.3 & 75.5 & 76.0 & 87.1 & \textbf{93.5} \\
\arrayrulecolor{linegray}\hhline{------}

\quad Is there an annular pancreas? & 3.9 & 27.6 & \textbf{29.9} & 27.2 & 7.0 \\
\arrayrulecolor{linegray}\hhline{------}

\quad Is there any pancreatic pseudocysts? & 89.6 & 72.4 & 73.3 & 85.8 & \textbf{92.0} \\
\arrayrulecolor{linegray}\hhline{------}

\quad Is there autoimmune pancreatitis? & NA & NA & NA & NA & NA \\
\arrayrulecolor{linegray}\hhline{------}

\quad Is there groove pancreatitis? & 77.7 & 54.0 & 63.1 & 73.8 & \textbf{95.1} \\
\arrayrulecolor{linegray}\hhline{------}

\quad Is there main pancreatic duct dilatation? & 81.4 & 71.9 & 72.3 & 81.0 & \textbf{85.6} \\
\arrayrulecolor{linegray}\hhline{------}

\quad Is there necrotizing pancreatitis? & 91.4 & 83.7 & 82.5 & 92.1 & \textbf{92.4} \\
\arrayrulecolor{linegray}\hhline{------}

\quad Is there pancreatic atrophy? & 79.2 & 78.4 & 78.1 & \textbf{80.3} & 80.2 \\
\arrayrulecolor{black}\midrule
\rowcolor{rowgray}
\textbf{Pelvic/GYN findings} & & & & & \\
\rowcolor{rowgray}
\quad Are there any cervical masses? & 88.0 & 61.2 & 66.6 & 79.7 & \textbf{89.2} \\
\arrayrulecolor{linegray}\hhline{------}
\rowcolor{rowgray}
\quad Are there any endometrial polyps? & 56.2 & 61.7 & 65.3 & 69.3 & \textbf{79.8} \\
\arrayrulecolor{linegray}\hhline{------}
\rowcolor{rowgray}
\quad Are there any endometriomas? & 83.0 & 77.9 & 77.1 & 83.4 & \textbf{97.3} \\
\arrayrulecolor{linegray}\hhline{------}
\rowcolor{rowgray}
\quad Are there any fractures of the pelvic girdle or sacrum? & 80.5 & 77.6 & 76.8 & 77.3 & \textbf{82.0} \\
\arrayrulecolor{linegray}\hhline{------}
\rowcolor{rowgray}
\quad Are there any intra-uterine devices? & 95.3 & 91.7 & 90.0 & 96.4 & \textbf{99.5} \\
\arrayrulecolor{linegray}\hhline{------}
\rowcolor{rowgray}
\quad Are there any ovarian cancers? & 88.7 & 77.0 & 79.5 & 88.5 & \textbf{90.3} \\
\arrayrulecolor{linegray}\hhline{------}
\rowcolor{rowgray}
\quad Are there any ovarian teratomas? & 76.9 & 60.7 & 63.3 & \textbf{80.7} & 77.3 \\
\arrayrulecolor{linegray}\hhline{------}
\rowcolor{rowgray}
\quad Are there any ovarian tumors? & \textbf{89.1} & 77.7 & 81.4 & 87.5 & 88.1 \\
\arrayrulecolor{linegray}\hhline{------}
\rowcolor{rowgray}
\quad Are there any uterine malformations? & 70.8 & 62.8 & 63.3 & 75.7 & \textbf{76.2} \\
\arrayrulecolor{linegray}\hhline{------}
\rowcolor{rowgray}
\quad Is there any lymphadenopathy in retroperitoneum, peritoneum or pelvis? & 72.7 & 67.4 & 69.2 & 70.6 & \textbf{75.4} \\
\arrayrulecolor{linegray}\hhline{------}
\rowcolor{rowgray}
\quad Is there deep infiltrative endometriosis? & 76.6 & 22.5 & 89.9 & 79.3 & \textbf{97.4} \\
\arrayrulecolor{linegray}\hhline{------}
\rowcolor{rowgray}
\quad Is there endometrial thickening? & 86.9 & 75.7 & 72.5 & 84.5 & \textbf{88.9} \\
\arrayrulecolor{linegray}\hhline{------}
\rowcolor{rowgray}
\quad Is there pelvic inflammatory disease? & 95.1 & 83.2 & 83.8 & \textbf{96.6} & 96.5 \\
\arrayrulecolor{linegray}\hhline{------}
\rowcolor{rowgray}
\quad Is there polycystic ovarian syndrome? & NA & NA & NA & NA & NA \\
\arrayrulecolor{black}\midrule

\textbf{Spleen-related findings} & & & & & \\

\quad Are there any accessory spleens? & \textbf{58.6} & 58.0 & 56.8 & 57.4 & 58.4 \\
\arrayrulecolor{linegray}\hhline{------}

\quad Are there any splenic abscesses? & 80.4 & 91.2 & \textbf{94.5} & 64.0 & 82.8 \\
\arrayrulecolor{linegray}\hhline{------}

\quad Are there any splenic hamartomas? & 12.0 & 17.3 & \textbf{58.4} & 10.4 & 23.8 \\
\arrayrulecolor{linegray}\hhline{------}

\quad Are there any splenic infarcts? & 91.3 & 86.5 & 87.8 & 88.9 & \textbf{95.2} \\
\arrayrulecolor{linegray}\hhline{------}

\quad Are there any splenic lacerations? & 90.1 & 87.5 & 88.7 & \textbf{93.2} & 89.3 \\
\arrayrulecolor{linegray}\hhline{------}

\quad Are there any splenic pseudocysts? & NA & NA & NA & NA & NA \\
\arrayrulecolor{linegray}\hhline{------}

\quad Is splenic lymphoma present? & 68.1 & 72.0 & 78.6 & 84.3 & \textbf{91.2} \\
\arrayrulecolor{linegray}\hhline{------}

\quad Is the spleen surgically absent? Are there any post-splenectomy changes? & NA & NA & NA & NA & NA \\
\arrayrulecolor{linegray}\hhline{------}

\quad Is there splenomegaly? & 91.6 & 84.3 & 87.3 & \textbf{95.4} & 93.6 \\
\arrayrulecolor{black}\midrule
\rowcolor{rowgray}
\textbf{Vascular findings} & & & & & \\
\rowcolor{rowgray}
\quad Are there any aortic dissections? & 88.9 & 66.9 & 75.9 & 81.6 & \textbf{93.1} \\
\arrayrulecolor{linegray}\hhline{------}
\rowcolor{rowgray}
\quad Are there any aortic intramural hematomas? & 71.6 & 21.5 & 41.5 & 81.5 & \textbf{85.9} \\
\arrayrulecolor{linegray}\hhline{------}
\rowcolor{rowgray}
\quad Are there any aortic stents? & 86.9 & \textbf{95.1} & 94.3 & 92.0 & 94.2 \\
\arrayrulecolor{linegray}\hhline{------}
\rowcolor{rowgray}
\quad Are there any aortic valve calcifications? & 87.3 & 89.0 & 88.3 & \textbf{90.6} & 90.4 \\
\arrayrulecolor{linegray}\hhline{------}
\rowcolor{rowgray}
\quad Are there any deep venous thromboses? & 72.2 & 71.2 & \textbf{73.6} & 68.3 & 73.3 \\
\arrayrulecolor{linegray}\hhline{------}
\rowcolor{rowgray}
\quad Are there any penetrating atherosclerotic aortic ulcers? & 67.8 & 73.8 & 92.3 & 94.5 & \textbf{95.5} \\
\arrayrulecolor{linegray}\hhline{------}
\rowcolor{rowgray}
\quad Are there coronary artery calcifications? & 85.1 & 87.0 & 85.3 & 87.4 & \textbf{87.7} \\
\arrayrulecolor{linegray}\hhline{------}
\rowcolor{rowgray}
\quad Is abdominal aortic aneurysm suspected based on the imaging findings? & 83.8 & 87.7 & 85.5 & 91.9 & \textbf{95.1} \\
\arrayrulecolor{linegray}\hhline{------}
\rowcolor{rowgray}
\quad Is aortic atherosclerosis suspected based on the imaging findings? & 84.8 & 85.8 & 84.3 & \textbf{87.9} & 87.8 \\
\arrayrulecolor{linegray}\hhline{------}
\rowcolor{rowgray}
\quad Is arterial or aortic thrombosis suspected based on the imaging findings? & \textbf{74.6} & 69.2 & 70.2 & 73.4 & 72.5 \\
\end{longtable}
\end{center}

\vspace{2mm}
\noindent
\footnotesize \textit{Abbreviations:} AUC, area under the curve; NA, not applicable.

\end{document}